\documentclass[letterpaper, 10 pt, conference]{ieeeconf}  % Comment this line out if you need a4paper

\IEEEoverridecommandlockouts                              % This command is only needed if 
\usepackage{subcaption} 
\usepackage{cite}
\usepackage{amsmath,amssymb,amsfonts}
\usepackage{algorithmic}
\usepackage{graphicx}
\usepackage{textcomp}
\usepackage{xcolor}
\usepackage{booktabs} 
\usepackage{multirow}
\usepackage{pbox}
\usepackage{overpic}
\usepackage{hyperref}
\def\BibTeX{{\rm B\kern-.05em{\sc i\kern-.025em b}\kern-.08em
		T\kern-.1667em\lower.7ex\hbox{E}\kern-.125emX}}

\title{\LARGE \bf
Deep Structured Reactive Planning
}

\author{Jerry Liu$^{1}$, Wenyuan Zeng $^{1,2}$, Raquel Urtasun $^{1,2}$, Ersin Yumer$^{1}$% <-this % stops a space
\thanks{$^{1}$ Uber ATG. Correspondence to: \tt\small jerryjliu98@gmail.com, wenyuan@cs.toronto.edu, urtasun@cs.toronto.edu, meyumer@gmail.com}%
\thanks{$^{2}$ University of Toronto}%
}

\begin{document}

\newcommand{\bw}{{\mathbf{w}}}
\newcommand{\by}{{\mathbf{y}}}
\newcommand{\cY}{{\mathcal{Y}}}
\newcommand{\cX}{{\mathcal{X}}}
\newcommand{\Exp}{{\mathbb{E}}}

\newcommand{\raquel}[1]{\textcolor{red}{Raquel: #1}}
\newcommand{\wenyuan}[1]{\textcolor{blue}{Wenyuan: #1}}
\newcommand{\hl}[1]{\textcolor{blue}{#1}}
\newcommand{\jerry}[1]{\textcolor{purple}{Jerry: #1}}
\newcommand{\ersin}[1]{\textcolor{cyan}{Ersin: #1}}

\maketitle
\thispagestyle{empty}
\pagestyle{empty}

%%%%%%%%%%%%%%%%%%%%%%%%%%%%%%%%%%%%%%%%%%%%%%%%%%%%%%%%%%%%%%%%%%%%%%%%%%%%%%%%
%\begin{abstract}
%
%This electronic document is a �live� template. The various components of your paper [title, text, heads, etc.] are already defined on the style sheet, as illustrated by the portions given in this document.
%
%\end{abstract}

% !TEX root = ./root.tex
\begin{abstract}

An intelligent agent operating in the real-world must  balance achieving its goal with maintaining the safety and comfort of not only itself, but also other participants within the surrounding scene.   
This requires jointly reasoning about the behavior of other actors  while deciding its own  actions  as these two process are inherently intertwined -- a vehicle will yield to us if we decide to proceed first at the intersection but will proceed first if we decide to yield.
However, this is not captured in most self-driving  pipelines, where planning follows prediction. 
In this paper  we propose a novel \textit{data-driven, reactive} planning objective which allows a self-driving vehicle to jointly reason about its own plans as well as how other actors will react to them.  
We formulate the problem as an  energy-based deep structured model that is learned from observational data and  encodes both the planning and prediction problems.
Through simulations based on both real-world driving  and synthetically generated dense traffic, we demonstrate that our reactive model outperforms  a non-reactive variant in successfully completing highly complex maneuvers (lane merges/turns in traffic) faster, without trading off collision rate. 
\end{abstract}

% Both the reactive and standard non-reactive objectives are based on a powerful deep energy-based model allowing us to explicitly model unary and pairwise interactions between actors. We demonstrate multiple key results: 1) that with validated hyperparameters, the reactive model outperforms the non-reactive model in simulated interactive scenarios, 2) that we can flexibly interpolate between reactive/non-reactive behavior to tradeoff goal completion/collision if necessary, 3) that our joint structured model outperforms state-of-the-art on standard prediction datasets. In future work, we plan on exploring how to better learn how to better adapt reactivity to the given scenario, potentially in a reinforcement learning setting.  

%%%%%%%%%%%%%%%%%%%%%%%%%%%%%%%%%%%%%%%%%%%%%%%%%%%%%%%%%%%%%%%%%%%%%%%%%%%%%%%%

% !TEX root = ./root.tex

\begin{figure*}
	\includegraphics[width=\linewidth]{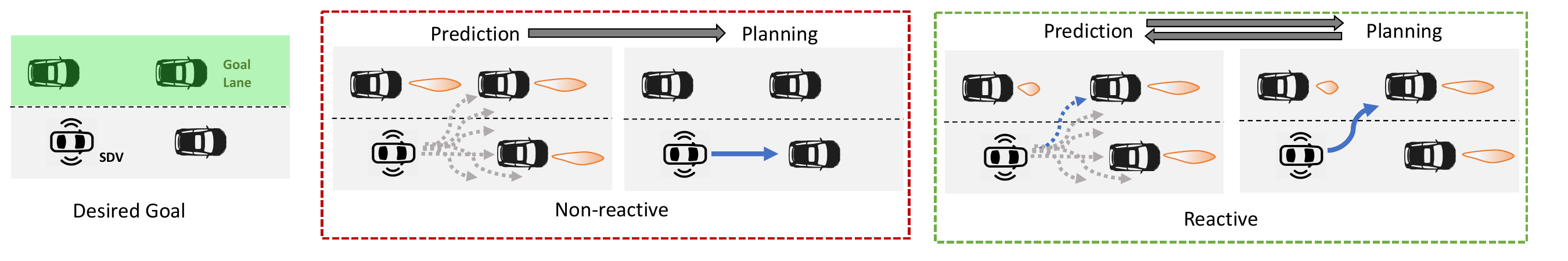}
	\caption{Comparison of a non-reactive planner vs. a reactive one in a
   lane merge scenario. Potential ego future trajectories are in
\textcolor{gray}{gray} and planned one is in \textcolor{blue}{blue}. The
non-reactive planner does not reason about how the actor will react to the
ego-agent's candidate trajectories and thus thinks it's impossible to lane-change without colliding with other actors, while the reactive planner reasons the neighboring actor will slow down, allowing it to complete the lane merge. }
	\label{fig:comp_planner}
\end{figure*}

\section{Introduction}

Self-driving vehicles (SDVs) face many challenging situations when dealing with complex dynamic environments. 
Consider a scenario where an SDV  is trying to merge left into a lane that is currently blocked by traffic. The SDV cannot reasonably merge by simply waiting - it could be waiting for quite a while and inconvenience the cars behind it. 
%Consider a scenario where a self-driving vehicle  is trying to merge off a ramp into a lane that is currently blocked by traffic. The SDV cannot reasonable merge by simply waiting - it could be stuck on that ramp for hours if the traffic in the merge lane is congested. 
On the other hand, it cannot aggressively merge into the lane  disregarding the lane congestion, as this will likely lead  to a collision. 
A human driver in this situation would think that if they gently nudge, other vehicles will have enough time to react without major inconvenience or safety risk, resulting in a  successful and safe merge.
While this is just an example,  similar situations  happen  often for example during rush hour,  in downtown areas, or highway ramp merging. 
The key idea here is that the human driver cannot be entirely \textit{passive} with respect to the dynamic multi-actor environment; they must exercise some degree of control by reasoning about how other actors will \textit{react} to their actions. 
Of course, the driver cannot use this control \textit{selfishly}; they must act in a responsible manner to  maximize their own utility while minimizing the risk/inconvenience to others.

%The execution stack of existing autonomous vehicle (AV) systems \cite{buehler_darpa}
%%\wenyuan{probably cite some DARPA SDV competition paper} 
%is an especially relevant example of
%modeling an intelligent actor navigating in a real-world multi-actor environment. 
This complex reasoning is, however, seldom used in self-driving approaches. 
Instead, the autonomy stack of an SDV is composed of a set of modules executed one after another. The AV first detects other actors in the scene
(\textit{perception}) and  predicts their future trajectories (\textit{prediction}). Given the output of perception and prediction, it plans a trajectory towards its intended goal that will be executed by the control module.
%\raquel{is devise the right word here?} a trajectory towards its intended goal that will be executed by the control module
%(\textit{planning} and \textit{control}). 
%In these systems, planning comes after prediction. \raquel{most systems? same comment about whether there is no work on reactive agents. I thought folks like Anca Dragan have work on this}
%\jerry{Definitely want to find a better way to word this, but didn't want to talk about related research work yet and mostly focus on existing self-driving stacks (which I'd still assume to follow this pipeline more or less.}
This implies that behavior forecasts of other actors are not affected by the AV's own
plan; the SDV is a passive actor assuming a stochastic world that it cannot change. 
As a consequence it might struggle when planning in high-traffic scenarios. 
In this paper we refer to prediction unconditioned on planning as \textit{non-reactive}.
%\raquel{isnt it non-reactive?}

Recently, there has been a line of   work that
identify similar issues and tries to incorporate how the ego-agent affects other
actors into the planning process; for instance, via game-theoretic
planning \cite{sadigh_effectshuman, sadigh_infogathering,
fisac_hierarchicalgame} and reinforcement learning \cite{saxena_densemfrl,
bouton_carl}. 
%Yet these works rely on hand-picked prediction models or planning rewards, which may not fully model real-world actor behavior, or make assumptions about their existence, which does not capture the full difficulty of learning predictions and rewards. 
Yet  these works rely on assumptions about a hand-picked prediction model or manually-tuned planning reward, which may not fully model real-world actor dynamics or human-like behaviors. 
%\raquel{human-like behaviors? expert behavior is an odd term}
%which may not generalize 
%to a diverse range of different real-world traffic scenarios. \raquel{this is not super clear. You mean that they only tackle one type of scenario?}
Thus there is a need for a more general approach to the problem. 

Towards this goal, we propose a novel \textbf{\textit{joint}} prediction and planning framework that can perform \textbf{reactive} planning. 
Our approach is based on cost minimization for planning where  we  predict the actor reactions to the potential ego-agent plans for costing the ego-car trajectories.  
We formulate the problem as a  \textbf{deep structured model} that defines a set of \textbf{learnable costs} across the future trajectories of all actors; these costs in turn induce a joint probability distribution over these actor future trajectories. 
A key advantage is that our model can be used jointly for prediction (with derived probabilities) and planning (with the costs). 
Another key advantage is that our structured formulation allows us to explicitly model interactions between actors and ensure a higher degree of safety in our planning.

We evaluate our reactive model as well as a non-reactive variant in a variety of highly interactive, complex closed-loop simulation scenarios, consisting of lane merges and turns in the presence of other actors. 
Our simulation settings involve both real-world traffic  as well as synthetic dense traffic settings. 
Importantly, we demonstrate  that using a reactive objective can more effectively and efficiently complete these complex maneuvers without trading off safety. 
Moreover, we validate the choice of our learned joint structured model by demonstrating that it is competitive or outperforms prior works in open-loop prediction tasks.

\section{Related Work}
%In this case, it might actually make sense to put the related work before the conclusion. 
\paragraph{Prediction}
%\raquel{talk about unrolling the state as first versions of prediction} \raquel{then talk about the use of map as rasterizations, then as lane graphs. }
The prediction task, also refer to as motion forecasting, aims to predict future states of each agent given the past. 
%This is a module crucial to robotics applications such as autonomous driving. 
Early
methods have used physics-based models to unroll the past actor
states \cite{welch_kalmanfilter, lefevre_pred_survey, cosgun_fullauto}. 
This field has exploded in recent years thanks to the advances in deep learning. 
One area of work in this space is to perform prediction (often jointly with detection) through rich unstructured sensor and map data as context \cite{luo_faf, casas_intentnet, zeng_nmp, liang_pnpnet, liang_lanegcn,li_e2epnptransformer}, starting with LiDAR context \cite{luo_faf} to map rasterizations \cite{casas_intentnet, djuric_shortmotion, cui_pred_dcn}, to lane graphs \cite{liang_lanegcn, gao_vectornet}.      
Modeling the future motion with a  multi-modal
distribution is of key importance given the  inherent future uncertainty \cite{casas_intentnet,
rhinehart_r2p2, rhinehart_precog, tang_mfp, hong_ror, zeng_dsdnet,
chai_multipath} and sequential nature of trajectories \cite{rhinehart_precog, tang_mfp} 
%\raquel{what do you mean by time dependence here?}. 
Recent works also model
interactions between actors \cite{casas_spagnn, rhinehart_precog, tang_mfp,
lee_desire, casas_ilvm,li_e2epnptransformer}, mostly through graph neural networks. In our work, we tackle the multi-modal and interactive
prediction with a joint structured model, through which we can efficiently
estimate probabilities. 
% In addition we also devise a holistic planning objective integrating both prediction and planning. 

%Prior work has generally used neural nets which take as input past trajectories and context information and output some representation of future trajectories. The output representation could be deterministic \cite{luo_faf,zeng_nmp} \raquel{note that intentnet predics a distribution over behaviors, so its not really deterministic}, or it can be a distribution \cite{casas_intentnet, rhinehart_r2p2, rhinehart_precog, tang_mfp, hong_ror}. A theme of recent work has also been to capture interactions between actors \cite{casas_spagnn, rhinehart_precog, tang_mfp, lee_desire, luke}, multi-modality \cite{chai_multipath, rhinehart_precog, tang_mfp, hong_ror, cui_pred_dcn} and time dependence \cite{rhinehart_precog, tang_mfp}. Some works have also focused on integrating detection with prediction such that uncertainties within visual information can be propagated to the predictor \cite{luo_faf, zeng_nmp}. In this work, we reason about trajectories using a joint structured model that allows us to encode priors about output dependencies, rather than directly regressing predictions through a neural net. \raquel{instead of just defining a set of works, tell a story of the prediction field that will allow the reviewer to understand, say the way somethign was develop, not just that it was developed}

\paragraph{Motion planning} Given observations of the environment and
predictions of the future, the purpose of motion planning is to find a safe and comfortable
trajectory towards a specified goal. 
Sample-based planning is a popular paradigm due to its low latency, where first a large set of trajectory candidates are sampled and evaluated based on a  pre-defined cost function,  and then  the minimal cost trajectory is chosen to be executed.
% The general framework of motion planning is finding a trajectory by minimizing some cost objective. 
Traditionally, such a cost function is hand-crafted to reflect our
prior knowledge \cite{fan_baidump, ziegler_bertha, montemerlo_junior,
buehler_darpa, bandyopadhyay_intentmp}. More recently, learning-based cost functions
also show promising results. Those costs can be learned through either Imitation
Learning \cite{sadat_plt} or Inverse Reinforcement Learning
\cite{ziebart_maxent_irl}.
In most of these systems, predictions are made independently of planning. While there has been recent work on accounting for actor reactivity in the planning process \cite{sun_courteous, sadigh_effectshuman, sadigh_infogathering, fisac_hierarchicalgame}, such works still rely on hand designed rewards or prediction models which may have difficulty accounting for all real-world scenarios in complex driving situations. 
%Furthermore, it is unclear if such approaches generalize to different scenario modalities.  

%\wenyuan{The last sentence is not very clear, why a pre-existing prediction
%model is a drawback? can you give me some hints what do you wanna criticize
%here? @jerry}
% These cost functions can also be learned through data using imitation learning / inverse reinforcement learning \cite{sadat_plt, ziebart_maxent_irl}.
%These costs can also be learned through approaches such as imitation learning \cite{sadat_plt, zeng_nmp}. 
% Recently, there has been work on accounting for actor reactivity in the planning process as well \cite{sun_courteous, sadigh_effectshuman, sadigh_infogathering, fisac_hierarchicalgame}, though such works rely on designed rewards or a pre-existing prediction model. 
% \raquel{you need a better related work on this}
% \jerry{I might need help on this. Wenyuan?}. 

\paragraph{Neural end-to-end motion planning} 
% Despite significant progresses in prediction and planning areas respectively, there
% are still fundamental issues been unaddressed. 
The traditional compartmentalization of prediction and planning results in the following
issues: First, hooking up both modules may result in a large system that can be
prohibitively slow for online settings. Second, classical planning usually
assumes predictions to be very accurate or errors
to be normally distributed,  which is not  realistic in practice. Third, the sequential
order of prediction and planning makes it difficult to model the interactions
between the ego-agent and other agents while making decisions.

To address these issues, prior works have started exploring
end-to-end planning approaches integrating perception, prediction and planning into
a holistic model. Such methods can enjoy fast
inference speed, while capture prediction uncertainties and model
prediction-planning interactions either implicitly or explicitly. One popular
way is to map sensor inputs directly to control commands via neural nets.
\cite{pomerleau_alvinn, bojarski_e2edrive, codevilla_e2e_condimit, muller_drivepolicy, bansal_chauffeurnet}. 
However, such methods lack interpretability and is hard to  verify safety. 
Recent works have proposed neural motion planners that produce interpretable intermediate representations. This are in the form of  non-parametric cost maps \cite{zeng_nmp}, occupancy maps
\cite{sadat_p3} or affordances \cite{sauer_affordance}.
% Due to some robustness issues associated with such a formulation \cite{codevilla_limit_bc}, other end-to-end works have focused on outputting a non-parametric cost map \cite{zeng_nmp, sadat_p3}. 
The most related work to ours, DSDNet \cite{zeng_dsdnet}, outputs a
structured model representation, similar to our setting - yet DSDNet still follows the traditional pipeline of separating prediction from planning, and thus cannot do reactive planning.
%\wenyuan{this sentence needs to be revised @jerry, right now it sounds a bit odd
%to me as typically people try to distinguish their methods with related work}
% Yet these works, there still generally exists an ordering where planning follows prediction, which follows perception.

%Another line of end-to-end planners focus their contributions on modeling
%reactiveness of other agents to ego-agent. This is in contrast with traditional
%system where predictions are made independent of planning.

% \paragraph{Joint prediction and motion planning}
% Motion planning is typically considered as a distinct component from prediction. \cite{rhinehart_deepimit} was an initial work that utilized a single-agent prediction model for planning. 
The two closest related works on modeling multi-agent predictions during end-to-end planning are PRECOG \cite{rhinehart_precog}, and PiP \cite{song_pip}. 
PiP is a prediction model that generates joint actor predictions conditioned on
the known future ego-agent trajectory, 
assuming planning is solved.
% however, the integration of such a model into a motion planner is left to future work.  
However, in the real-world, finding the future ego-trajectory
(planning) is in  itself a challenging open problem, since the ego-trajectory
depends on other actors which creates a complicated feedback-loop between planning
and prediction.
% \raquel{talk about why the known fugute ego-agent trajectory is not relistic, and why the integration into planning is non trivial}
The PRECOG planning objective accounts for joint reactivity under a flow-based \cite{rezende_flow} framework, yet it requires sequential decoding for planning and
prediction making it hard to satisfy low-latency online requirements. Moreover, the planning objective does not ensure collision avoidance and can suffer from mode collapse in the SDV trajectory space. 
%\raquel{why do you think tahat their prediction is not expressive?}
% however, we demonstrate that our reactive framework yields better planning results in simulation and better prediction metrics in CARLA and Nuscenes. 
% \raquel{also talk aoubt the conteptual differences with PRECOG, not just that it works worst}  
%PiP conditions on the ego-agent trajectory but is more focused on conditional prediction than the final planning objective.
%\wenyuan{sorry what do you mean here? It's a bit vague to
%me. If possible, we may want to say more here to distinguish the differences, as these two are very relevant} 

%(TODO: cite the load of papers on modeling both prediction as well as planning.)
%Prediction: IntentNet, PRECOG, MFP, RoR, SpaGNN, Waymo AnchorNet, etc. Maybe add some imitation learning papers too. 
%Planning: PLT (maybe more?)
%
%Integrated approaches: NMP, DSDNet. DSDNet actually has a similar formulation to our approach, however; prediction and planning are still factored into separate components, whereas we reason about how the joint energy model can be utilized to model reactive actor behavior. 

\paragraph{Structured Models}
Researchers have applied neural nets to learn parameters in undirected graphical models, also known as Markov Random Fields (MRF's). One of the key challenges in training MRF's in the discrete setting is the computation of the partition function, where the number of states increases exponentially with the number of nodes in the worst case scenario. Message-passing algorithms such as Loopy Belief Propagation (LBP) have been found to approximate the partition function \cite{yedidia_gbp, mceliece_turbo} well in practice. 
Other learning based methods have included dual minimization \cite{chen_deepstruct}, directly optimizing the Bethe Free Energy \cite{wiseman_abfem}, finding variational approximations \cite{kuleshov_nvi} or mean-field approximations \cite{schwing_fc_dsn}. 
%\raquel{In this last sentence there is a bit of a mix between learning and inference}
%Other deep-learning based methods have included dual minimization \cite{chen_deepstruct}, directly optimizing the Bethe Free Energy \cite{wiseman_abfem}, finding variational approximations \cite{kuleshov_nvi} or mean-field approximations \cite{schwing_fc_dsn} 
%In our paper, although we would like our costs to represent the joint likelihood of all actors, we found that simplifying the training to better fit our inference objective yielded better results. 
%\raquel{this last sentence is very odd, not clear what you mean}
%Although we focus on the discrete setting, methods for learning continuous MRF's also exist. 
%Rather than using a maximum likelihood formulation, 
Some approaches take an energy-minimization approach, unrolling the inference objective through differentiable optimization steps \cite{belanger_spen, wang_crf_polynomial} that can also be used to learn the model parameters \cite{belanger_spene2e,  wang_proximaldsm}.
\section{ Joint   Reactive Prediction and Planning}

\begin{figure}
	\includegraphics[width=\linewidth]{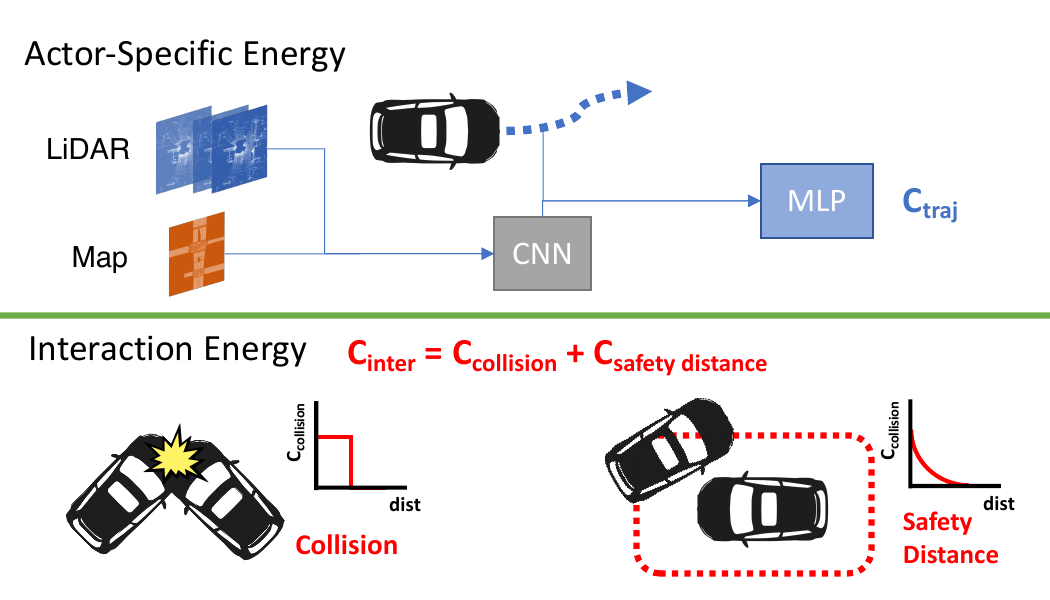}
	\caption{Overview of the actor-specific and interaction energy terms in our joint structured model.}
	\label{fig:struct_overview}
\end{figure}

%In this section we describe our joint structured framework for \textit{reactive} prediction/planning. We define reactive in this setting to mean the ego-agent being able to predict actor behavior given the ego-agent plan. Correspondingly, we define \textit{non-reactive} to mean that the ego-agent's predictions of other actors are not conditioned on the ego-agent plan. 
%\wenyuan{Alternatively, we can define in this way (less colloquially I hope) For example: Suppose other actor's ground truth behavior is $Y_o^*$ and ego plan
%is $y_e$. In reality, $Y_o^*$ is a random variable dependent of $y_e$. However, it's up to the model's assumption whether the model will follow this
%fact. In the following, we define reactive as a model's prediction $Y_o$ depends on $y_e$, and non-reactive if the model assumes they are independent. }
%\jerry{Nice, I like this.} 

Suppose the SDV is driving in a scenario where there are $N$ actors. Let $\mathcal{Y} = (\by_0, \by_1, \cdots, \by_N)$ be the set of random 
variables representing future trajectories of both the SDV, $\by_0$, and all other traffic participants,  $\cY_r = (\by_1,  \cdots \by_N)$. 
%Note that  the  trajectories of the other traffic participants  $\mathcal{Y}_o$ are dependent on the  SDV executed plan $\by_e$. %Traditional \textit{non-reactive} pipelines where planning follows prediction do not. 
%We define a \textit{reactive planner}  as one where the model's actor
%predictions $\cY_a$ depend on $\by_0$ \wenyuan{and planner in turns consider
%that reactive prediction?}, and a \textit{non-reactive planner} as one where the
%model assumes they are independent \wenyuan{this is not true as planning is
%still depends on prediction?}. 
We define a \textit{reactive} planner as one that considers actor predictions $\cY_r$ conditioned on $\by_0$ in the planning objective, and a \textit{non-reactive} planner as one that assumes a prediction model which is independent of $\by_0$. 
In this section, we first outline the framework of our joint structured model
which simultaneously models the costs and probability distribution of the future (Sec.
\ref{sec:struct_model}). 
We then introduce our reactive objective which
enables us to safely plan under such a distribution considering the
reactive behavior of other agents (Sec. \ref{sec:plan_objective}) and discuss how to evaluate it (Sec. \ref{sec:inference}), including with a goal-based extension (Sec. \ref{sec:goal_energy}). We finally describe our training procedure (Sec. \ref{sec:training}). We highlight additional model properties, such as interpolation between our non-reactive/reactive objectives, in our supplementary. 
%We demonstrate interpolation between the reactive/non-reactive objectives (Sec. \ref{sec:interpolation}), and describe our training procedure (Sec \ref{sec:training}).
%In this section \raquel{talk about what we show in this section and in which order}

\subsection{Structured Model for Joint Perception and Prediction}  \label{sec:struct_model}
We define a probabilistic deep structured model to represent the distribution
over the future trajectories of the actors conditioned on the environment context $\cX$
as follows 
%\begin{small}
	\begin{align}
	p(\cY | \cX; \bw) = \frac{1}{Z} \exp(-C(\cY, \cX; \bw)) \label{eq:joint_struct}
	\end{align}
%\end{small}
where $Z$ is the  partition function, $C(\cY, \cX; \bw)$ defines the joint  energy of all future trajectories $\cY$ and  $\bw$ represents all the parameters of the model. 
In this setting, the context $\cX$ includes each actor's past trajectories, LiDAR
sweeps and HD maps, represented by a birds-eye view (BEV) voxelized tensor
representation \cite{casas_intentnet, luo_faf}. 
%\raquel{if you dont give details point to where this came from}
%\hl{(environment information)} 
% Note that $\cX$ represents the joint \textit{past context} among all actors. 
% Finally, in the dataset setting where HD maps available, we similarly rasterize the HD maps to the same resolution as our LiDAR tensor and similar include it with $\cX$. Note that the dataset-dependent input details are provided in supplementary. 
%{\bf Trajectory parameterization:} 
Actor trajectories $\cY$ can naturally be represented in continuous space. However, performing inference on continuous structured models is extremely  challenging. We thus instead follow \cite{zeng_dsdnet,phan2020covernet} and discretize each actor's action
 space into $K$ possible trajectories (each continuous) using a realistic
 trajectory sampler inspired from \cite{zeng_dsdnet}, which takes past
 positions as input and samples a set of lines, circular curves, and euler
 spirals as future trajectories. 
% \raquel{NMP didnt use this for the other actors, so we should only cite DSDNet or?}
Thus, each $\by_i$ is a discrete random variable that can take up one of $K$ options, where each option is a full continuous trajectory  -- such a discretized distribution allows us to efficiently compute predictions (see Sec. \ref{sec:inference}). Additional details on input representation and trajectory sampling are provided in the supplementary material. 
%\raquel{explain that a value here is a full continuous trajectory. Important, as they might think that you discretize over time instead of over set of trajectories}

% \hl{Thus, each $y_i$ is formulated as a
% discrete random variables that can take value from $1$ to $K$. We end up with a
% discrete probability distribution which enables us to make efficient inference
% as we'll show shortly.}\wenyuan{todo, cite the equation} We refer the reader to the supplementary for more details.

%\begin{small}
%\begin{align}
%C(Y, X; w) = &C_{y_e}(y_e, X; w) + \sum_i{C_{y_e, y^i_{o}}(y_e, y^i_{o}, X; w)} + \\ 
%&\sum_i{C_{y^i_{o}}(y^i_{o}, X; w)} +  \sum_{i,j}{C_{y^i_{o}, y^j_{o}}(y^i_{o}, y^j_{o}, X; w)}
%\end{align}
%\end{small}
% \wenyuan{This trainsition is pretty sharp. I think we should add at least one
% transition sentence here...}
%We define a decomposable energy 
We decompose the joint energy $C(\cY, \cX; \bw)$ in terms of an \textbf{actor-specific}  energy that encodes the cost of a given trajectory for each actor, while the \textbf{ interaction term} captures the plausibility of trajectories across two actors: 
%\raquel{instead of unary and pairwise talk about object specific and interactions}
%\begin{small}
	\begin{align}
	%C(\cY, \cX; \bw) = &C_{\text{unary}}(\by_e, \cX; \bw) + \sum_i{C_{\text{unary}}(\by^i_{o}, \cX; \bw)} + \\ & \sum_i{C_{\text{inter}}(\by_e, \by^i_o)} + \sum_{i,j}{C_{\text{inter}}(\by^i_{o}, \by^j_{o})}	 
	 C(\cY, \cX; \bw) = & \sum_{i=0}^N C_{\text{traj}}(\by_i, \cX; \bw) +  \sum_{i,j} C_{\text{inter}}(\by_i, \by_j)   
	\end{align}
%\end{small}
%\raquel{maybe remind the reader that $\by_0$ is the ego-car}
%\wenyuan{I would prefer exchange the second and the third term as the first and third terms are unary?}
%\wenyuan{Also, what do you think if replacing the subscripts of C with unary/pairwise. As right now the subscript is identical to the variables inside
%parathensis so I'm afraid it doesn't bring information or ease the reading flow?}
%We will demonstrate that a discretization of trajectory space per actor gives us more flexibility in terms of both inference (Section \ref{sec:inference}) and training (Section \ref{sec:training}). 
We exploit a learnable neural network to compute the {\bf actor-specific energy}, 
$C_{\text{traj}}(\by_i, \cX; \bw)$, parameterized with weights $\bw$. A convolutional network takes as input the  context feature $\cX$ as a rasterized BEV 2D tensor grid centered around the ego-agent, and produces an intermediate spatial feature map $\mathbf{F} \in \mathbb{R}^{h \times w \times c}$, where $h,w$ represent the dimensions of the feature map (downsampled from the input grid), and $c$ represents the number of channels. 
%\raquel{describe a bit the architecture. Otherwise its impossible to understand for the reader. Or say a bit more and point to supplementary here}
These features are then combined with the candidate trajectories $\by_i$ and processed through an MLP, outputting a $(N+1) \times K$ matrix of trajectory scores, one per actor trajectory sample. 
%The model architecture for computing the
%actor-specific energy is divided into two components. The first is a
%\textbf{backbone network} which takes in available context input (sensor input
%like LiDAR, HD map data) and computes intermediate spatial feature maps within a
%region of interest around the ego-agent. The second is a \textbf{learnable unary
%module}, parametrized by deep neural nets with weights $\textbf{w}$, computing
%the trajectory costs $C_{\text{traj}}$ for each actor given their past
%trajectories and backbone features. \wenyuan{I'm not sure people can understand
%this part without a figure or more detailed explanations. What about just say a
%unary module, don't mention the backbone. Just say the 'unary is computed from
%a context feature (processed by CNN on $\cX$) and a trajectory feature
%(processed by MLP on $y_i$). We leave details to supp' } 
%\raquel{explain the architecture used for this and the backbone network}
%\raquel{explain this cost better}
%\raquel{merge this paragraph into the previous paragraph by simply stating how you compute the trajectory cost}
Our  \textbf{interaction energy}  is a combination of collision and safety distance violation costs. 
We define the collision energy to be $\gamma$ if a pair of future trajectories collide and 0 if not. 
Following \cite{sadat_plt}, we define the safety distance violation  to be a squared penalty within some safety distance of each actor's bounding box, scaled by the speed of the SDV. In our setting, we define safety distance to be 4 meters from other vehicles. 
%\raquel{say what you use for vehicles, peds and bicyclist}
Fig. \ref{fig:struct_overview} gives a graphic representation of the two energy terms. 
Full model details are in the supplementary, including the specific dataset-dependent input representation and model architecture. 

\subsection{Reactive Inference Objective} \label{sec:plan_objective}
The structured model defines both a set of costs and probabilities over possible futures. 
We develop a planning policy on top of this framework which decides what the
ego-agent should do in the next few seconds (i.e., planning horizon). 
Our reactive
planning objective is based on  an optimization formulation which finds the trajectory
that minimizes a set of planning costs -- these costs consider both the candidate SDV trajectory as well as other actor predictions conditioned on the SDV trajectory. 
%\raquel{we need to explain that this cost is not the same as in the previous C}
In contrast to existing literature, we re-emphasize that both prediction and planning components of our objective are derived from the same set of learnable costs in our structured model, removing the need to develop extraneous components outside this framework; we demonstrate that such a formulation inherently considers both the reactivity and safety of other actors. 
%Different from existing literatures, our
%planning objective explicitly consider the reactiveness of other actor, and can
%seamlessly work with a learnable planner} \wenyuan{I'm still kind of unsatisfied
%with this sentence. The issue is that we want to highlight th reactive planning,
%however I believe this itself is not brand new (there must be some feedback
%control literature), all we can say is in the DL era this is new, but I haven't
%find a elegant way to deliver this. @jerry can you also help think about this?
% Within this joint structured framework, we can design our planning objective - which entails finding a trajectory for the ego-agent that minimizes cost. 
We define our planning objective as 
\begin{align}
\by_0^* = \text{argmin}_{\by_0} f(\cY, \cX;\bw) \label{eq:cond_0}
\end{align}
where $\by_0$ is the ego-agent future trajectory and $f$ is the planning cost function defined over our structured model. 

In our reactive setting, we define the planning costs to be an expectation of the joint energies, over the distribution of actor predictions conditioned on the current candidate SDV trajectory: 
\begin{align}
%\text{argmin}_{\by_e} f_{\text{reactive}} =  \text{argmin}_{\by_e} \Exp_{\cY_o \sim p(\cY_o | \by_e, \cX; \bw)}[C(\by_e, \cY_o, \cX; \bw)]  \label{eq:cond_1}
f(\cY, \cX; \bw) =   \Exp_{\cY_r \sim p(\cY_r | \by_0, \cX; \bw)}[C(\cY, \cX; \bw)]  \label{eq:cond_1}
\end{align}
Note that $\cY_r \sim p(\cY_r | \by_0, \cX; \bw)$ describes the future distribution of other actors, conditioned on the current candidate trajectory $\by_0$ and is derived from the underlying joint distribution in Eq. (\ref{eq:joint_struct}). Meanwhile, the $C(\cY, \cX; \bw)$ term represents the joint energies of a given future configuration of joint actor trajectories. We can expand the planning objective by decomposing the joint energies into the actor-specific and interaction terms as follows: 

%\raquel{what is this? the definition of f? confusing}
\begin{small}
\begin{align}
C_{\text{traj}}(\by_0, \cX; \bw) &+ \Exp_{\cY_r \sim p(\cY_r | \by_0, \cX; \bw)}[\sum_{i=1}^{N}C_{\text{inter}}(\by_0, \by_i) + \\
			  & \sum_{i=1}^{N}C_{\text{traj}}(\by_i, \cX; \bw) + \sum_{i=1,j=1}^{N,N}C_{\text{inter}}(\by_i, \by_j)] \nonumber
\end{align}
\end{small}
%\raquel{rewrite this part now that we have remove the previous stuff}
The set of costs  includes the SDV-specific cost, outside the expectation. It also includes the SDV/actor interaction costs, the actor-specific cost, and actor/actor interaction costs within the expectation. Note that the SDV-specific cost $C_{\text{traj}}(\by_0, \cX; \bw)$ uses a different set of parameters from those of other actors $\by_i$ to better exploit the ego-centric sensor data and model SDV-specific behavior. Moreover, the set of actor-specific and interaction costs within the expectation leads to an inherent balancing property of additional responsibility to additional control: by explicitly modeling the reactive prediction distribution of other actors in the prediction model, we must also take into account their utilities as well. 
In the following, we further exclude the last energy term $ \sum_{i,j}C_{\text{inter}}(\by_i, \by_j)$ due to computational reasons. See supplementary material for more details. 
%The actor-specific and pairwise interaction costs of other actors are now relevant to determining the expectation term and cannot be simplified out. 

%\raquel{we need to explain either here or in the learning that we have different costs for the ego car. Probably in the learning subsection}

\subsection{Inference for Conditional Planning Objective} \label{sec:inference}

Due to our discrete setting and the nature of actor-specific and interaction costs, for any given $\by_0$, we can directly evaluate the expectation from Eq. (\ref{eq:cond_1}) without the need for Monte-Carlo sampling.
%\raquel{what do you mean by exactly? this is a hard problem to solve}
%Moreover, we extend the equation by adding $\lambda$'s to represent weighting factors on actor-specific and interaction costs among \textit{all} actors:  \raquel{this is folded into the cost, we dont need to have more paramerers}
%\begin{small}
%\begin{align}
%\tilde{L} &= \lambda_a C(y_e, X) + \\
%			  & \sum_{Y_o}{p(Y_o | y_e, X)}[\lambda_b \sum_{i}C_{y_e, y_{oi}} å+ \lambda_c \sum_{i}C_{y_{oi}} + \lambda_d \sum_{i,j}C_{y_{oi},y_{oj}}] \\
%&=  \lambda_a C(y_e, X) + \lambda_b \sum_{i,y_{oi}}p_{y_{oi}|y_e}C_{y_e, y_{oi}} + \\ 
%& \lambda_c \sum_{i,y_{oi}}p_{y_{oi}|y_e}C_{y_{oi}} + \lambda_d \sum_{i,j,y_{oi},y_{oj}}p_{y_{oi},y_{oj}|y_e}C_{y_{oi},y_{oj}} \label{eq:inf}
%\end{align}
%\end{small}
%\begin{small}
%	\begin{align}
%	f &= \lambda_a C_{\text{traj}}^{\by_0} + \sum_{\cY_r}{p_{\cY_r | \by_0}}[\lambda_b \sum_{i=1}^{N}C_{\text{inter}}^{\by_0, \by_i} + \lambda_c \sum_{i=1}^{N}C_{\text{traj}}^{\by_i}] 
%	\end{align}
%\end{small}
%\begin{small}
We thus have
	\begin{align}
	f &=  C_{\text{traj}}^{\by_0} + \sum_{\cY_r}{p_{\cY_r | \by_0}}[ \sum_{i=1}^{N}C_{\text{inter}}^{\by_0, \by_i} +  \sum_{i=1}^{N}C_{\text{traj}}^{\by_i}] 
	\end{align}
%\end{small}
%\wenyuan{I would advocate putting the paragraph of $\lambda$ to be the first one, otherwise reader
%will be confusing about $\lambda$ at this point. and then talk about inference complexity as people
% won't directly think of it until you mention, so it's less urgent compared to explaining $\lambda$}
where $p_{\by_i|\by_0}$ is short-hand for $p(\by_i | \by_0, \cX; \bw)$, and $C_{\text{traj}}^{\by_i}$ for $C_{\text{traj}}(\by_i, \cX; \bw)$ (same for pairwise).
Since the joint probabilities factorize over the actor-specific and pairwise interaction energies, they simplify into the marginal and pairwise marginal probabilities between all actors. 
	\begin{align}
	f &=   C_{\text{traj}}^{\by_0} +  \sum_{i,\by_i}p_{\by_i|\by_0}C_{\text{inter}}^{\by_0, \by_i} +  \sum_{i,\by_i}p_{\by_i|\by_0}C_{\text{traj}}^{\by_i}
	%f &=  \lambda_a C_{\text{traj}}^{\by_0} + \lambda_b \sum_{i,\by_i}p_{\by_i|\by_0}C_{\text{inter}}^{\by_0, \by_i} + \lambda_c \sum_{i,\by_i}p_{\by_i|\by_0}C_{\text{traj}}^{\by_i}
	\end{align}
%The $\lambda$'s represent weighting factors on unary and pairwise costs among \textit{all} actors. 
%These $\lambda$ weights can be adjusted during inference to give more weight to certain terms than others. 
%For instance, setting a higher $\lambda_b, \lambda_c$ puts more of an emphasis on the safety and comfort of other actors - a higher $\lambda_c$ would more strongly penalize deviations from the ``expected'' actor trajectory due to the ego-agent's actions, even if there is no pairwise collision. Conversely, lower values of $\lambda_b, \lambda_c$ would lead to more aggressive behavior. 
where 
$p_{\by_i|\by_0}$ represents the marginal probability of the actor trajectory conditioned on the candidate ego-agent trajectory. 
These  marginal probabilities which are tensors of size $N \times K \times K$, can all be efficiently approximated by exploiting Loopy Belief Propagation (LBP)  \cite{yedidia_gbp}. 
This in turn allows efficient batch evaluation of the planning objective: for every sample of every actor ($N \times K$ samples), evaluate the conditional marginal probability times the corresponding energy term.  
Note that LBP  can also be interpreted as a special form of recurrent network, and thus is amenable to end-to-end training. 
%across all $K$ and $K^2$ samples within a CRF can all be efficiently approximated at once through one round of Loopy Belief Propagation (LBP) message passing \cite{yedidia_gbp}. 
%Unfortunately, $p_{\by_o^i,\by_o^j|\by_e}$, the pairwise marginals of two actors conditioned on the ego-agent trajectory, would require $K$ rounds of message passing to obtain the actor-actor pairwise marginals for every value of $\by_e$; 
%hence, in practice we exclude this term. 
%hence, in practice we either exclude this term or simplify it to $p_{y_{oi},y_{oj}|y_e} = p_{y_{oi}|y_e}p_{y_{oj}|y_e}$, while acknowledging the safety hit in disregarding collisions among other actors.  
Then, since the ego-agent itself has $K$ trajectories to choose from, solving the minimization problem in (\ref{eq:cond_0}) involves simply picking the trajectory with the minimum planning cost.  
%\raquel{point to equation before eq. 3. Give it a number}

\begin{table}[]
	\centering
	%		\scalebox{0.64}{
%	\scalebox{0.85}{
		\scalebox{0.83}{
		\begin{tabular}{c|ccccc}
%		\begin{tabular}{@{}|@{}|p{2cm}p{2cm}p{2cm}p{2cm}p{2cm}|}
			\toprule
			Model & Succ (\%) $\uparrow$ & TTC (s)  $\downarrow$ & Goal (m) $\downarrow$ & CR (\%) $\downarrow$ & Brake $\downarrow$   \\ \midrule
			PRECOG (C) & 12.0 & 16.3 & 13.5 & 18.0 & 39.2 \\
		 	Non-Reactive (C) & 46.0 & 15.8 & 4.2 & 5.0 & \textbf{34.4} \\
			 \textbf{Reactive } (C)& \textbf{70.0} & \textbf{13.9} & \textbf{2.4} & 5.0 & 37.8  \\ \midrule
			%		   						 & PiP & - & - & - & - & - \\ \midrule
			
			PRECOG (S) & 21.0 & 9.4 & 16.8 & 20.5 & - \\
			 Non-Reactive (S) & 70.0 & 7.5 & 5.3 &  3.5 & - \\
		 	\textbf{Reactive} (S) & \textbf{82.0} & \textbf{6.8} & \textbf{4.3} & 3.5 & - \\ 

			%								 & PiP & - & - & - & - & - \\
			\bottomrule
	\end{tabular}}
	\caption{Results obtained from simulations in Simba/CARLA. C = CARLA, S = Simba.  }
	\label{tab:overall_sim}
\end{table}

\subsection{Goal Energy} \label{sec:goal_energy}
%\raquel{its odd to have this outside the model. Point to PRECOG or the paper that was doing this, and explain why you might want to keep it out to have flexibiity in what this encodes}
Similar to \cite{rhinehart_precog}, \cite{rhinehart_deepimit}, we make the observation that our current formulation, which encodes both actor behavior and desirable SDV behavior in the energies of our structured model, can be extended to  goal-directed planning to flexibly achieve arbitrary goals during inference.
In addition to the learned ego-agent cost $C_{\text{traj}}^{\by_0}$, we can specify a goal state $\mathcal{G}$ in each scenario and encourage the
ego-agent to reach the goal state via a goal energy  $C_{\text{goal}}^{\by_0}$. The goal state can take on different forms depending on the scenario:
in the case of a turn, $\mathcal{G}$ is a target position.
%or a sequence of positions (for more complex roundabout turns in CARLA, see Sec. \ref{sec:experiments}). 
%\wenyuan{Do you mean you actually have these two set of target in
%experiments? or just it can be any one of them? If it's the later case, I
%suggest just mention which one you use. If it's the former one, why do you use
%two different forms} 
%In the case of a lane change, $\mathcal{G}$ is a set of two points capturing the direction and position of the lane in continuous coordinates. \raquel{why two points?}
%In particular we define the goal energy term $C_{\text{goal}}^{\by_0}$ as follows: if $\mathcal{G}$ is a single point, the energy is the $\ell_2$ distance of the final waypoint; if $\mathcal{G}$ represents a lane, the energy represents the average projected distance to the lane goal vector.  \raquel{shouldnt this be piecewise linear?}
In the case of a lane change, $\mathcal{G}$ is a polyline representing the centerline of the lane in continuous coordinates. In particular we define the goal energy term $C_{\text{goal}}^{\by_0}$ as follows: if $\mathcal{G}$ is a single point, the energy is the $\ell_2$ distance of the final waypoint; if $\mathcal{G}$ represents a lane, the energy represents the average projected distance to the lane polyline.
We sum the goal energy cost to the conditional planning objective during inference.
%\wenyuan{Just a random thought. maybe we can just integrate this goal part to
%your model section? not sure.}

%\subsection{Interpolating between Reactive and Non-Reactive Objectives} \label{sec:interpolation}
%We observe an additional level of flexibility in this model: being able to interpolate between our reactive/non-reactive objectives, which have thus far been presented as distinct. The key lies within our conditional prediction model for a given actor $\by_o^i$: $p(\by_o^i | \by_e, \cX; \bw)$, currently conditioned on a single ego-agent plan. We can modify the conditioning to be on a set: $S^{\by_e}$ with $k, 1 \leq k \leq K$ elements, which are the top-$k$ candidate trajectories closest to $\by_e$ by L2 distance. Then, we define $p(\by_o^i | S^{\by_e}, \cX; \bw) = \frac{1}{Z} \sum_{\by^j_e \in S^{\by_e}} {p(\by_o^i, \by^j_e, \cX; \bw)}$, where $Z$ is a normalizing constant. Intuitively, conditioning actor predictions on this set implies that actors do not know the exact plan that the AV has, but may have a rough idea about the general intent. When $|S^{\by_e}|$ is 1, we obtain our reactive model. When $|S^{\by_e}|$ is $K$, it is straightforward to see that $Z=1$, and hence we obtain our actor marginals $p(\by_o^i | \cX; \bw)$ used in the non-reactive model. 

%\wenyuan{do we have a model architecture section somewhere?}

\newcommand{\figcap}[3]{
	\begin{overpic}[width=\imw, trim={#2},clip]{#1}
		\put(0,2){\sffamily \scriptsize \colorbox{gray}{\color{white} #3}}
	\end{overpic}
}

\begin{figure*}[h]
	\centering
	\def\imw{0.24\textwidth}
	\setlength{\tabcolsep}{0.2pt}
	\begin{tabular}{cccc}
	\figcap{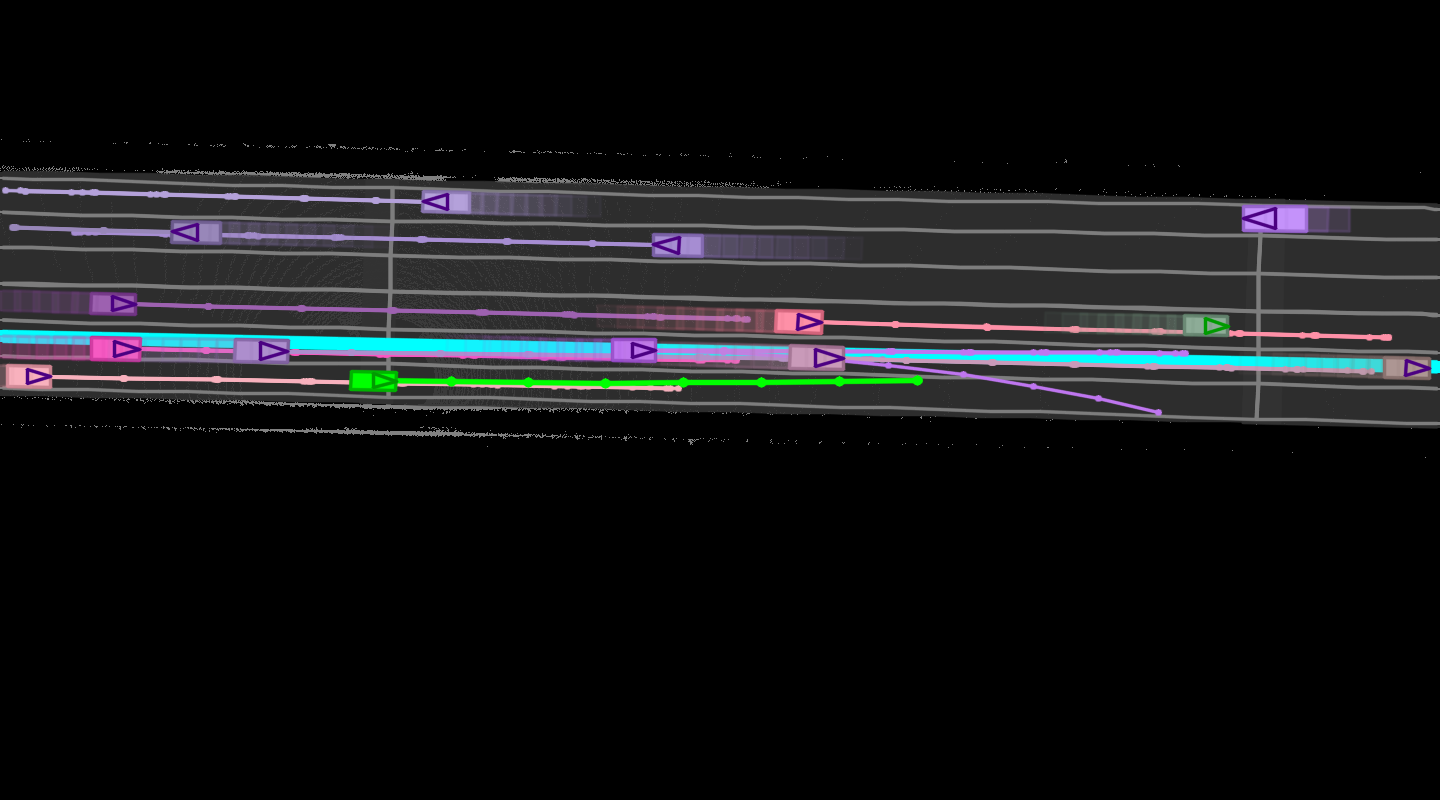}{300 275 500 225}{Reactive, t=0s} &
	\figcap{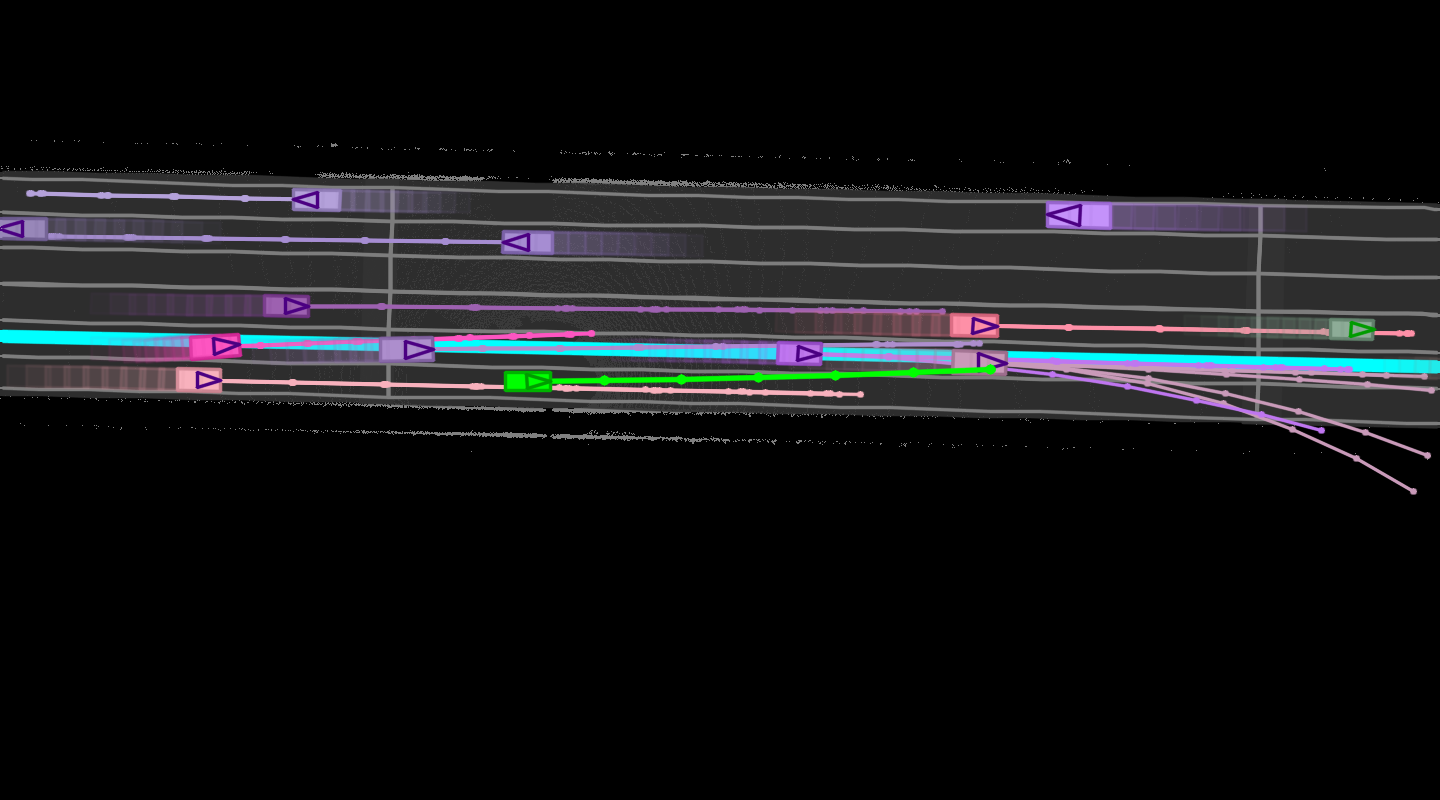}{300 275 500 225}{Reactive, t=1s} &
	\figcap{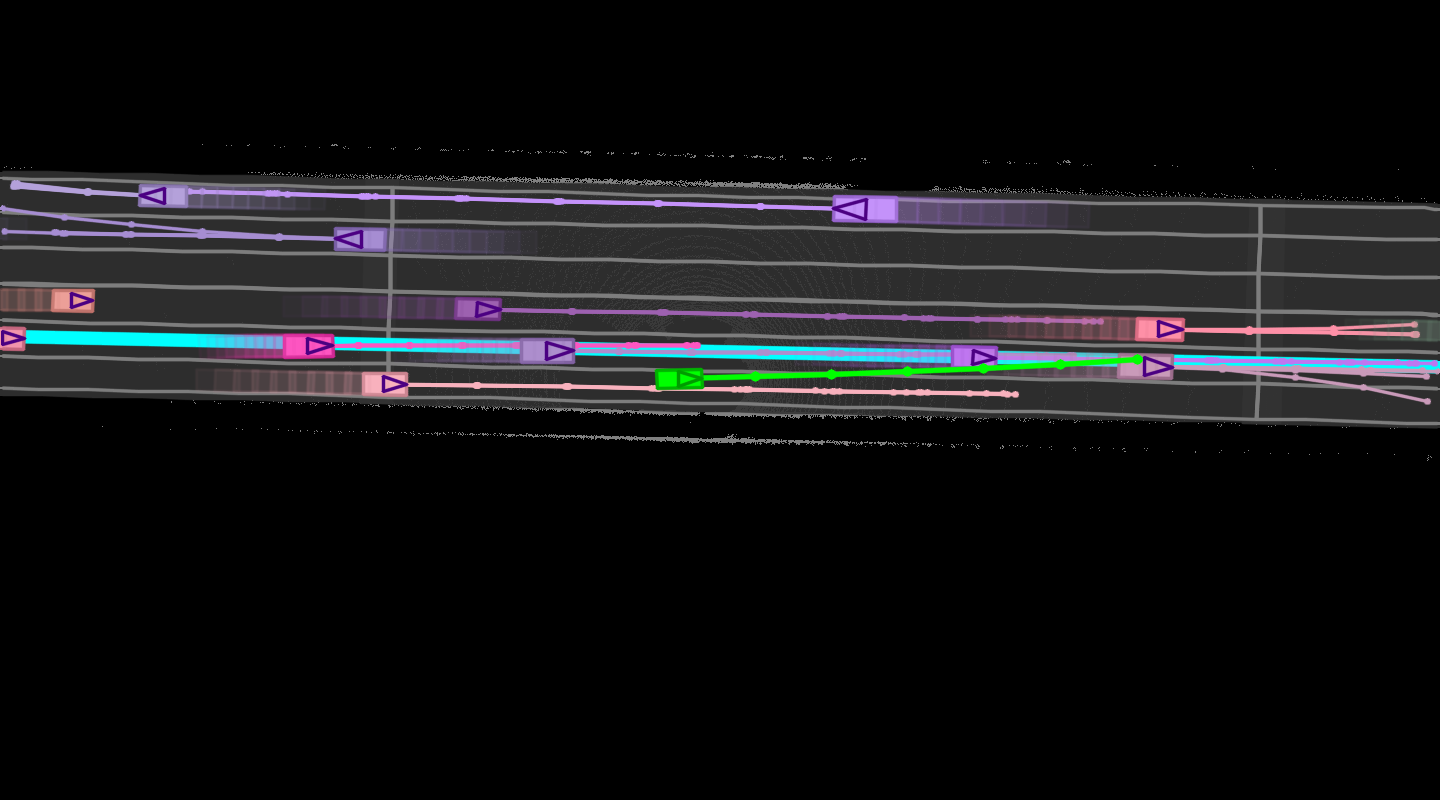}{300 275 500 225}{Reactive, t=2s} &
	\figcap{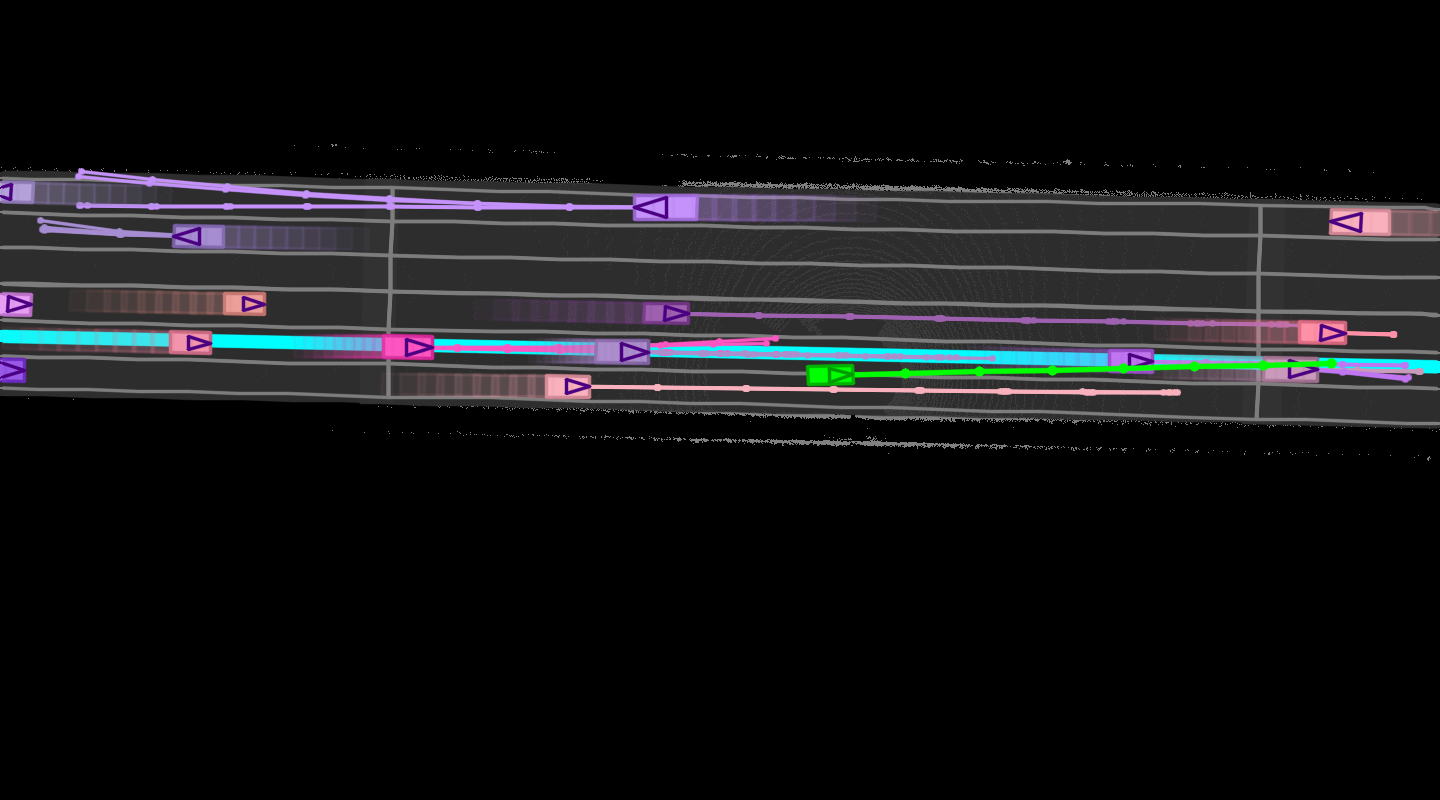}{300 275 500 225}{Reactive, t=3s} \\
	\figcap{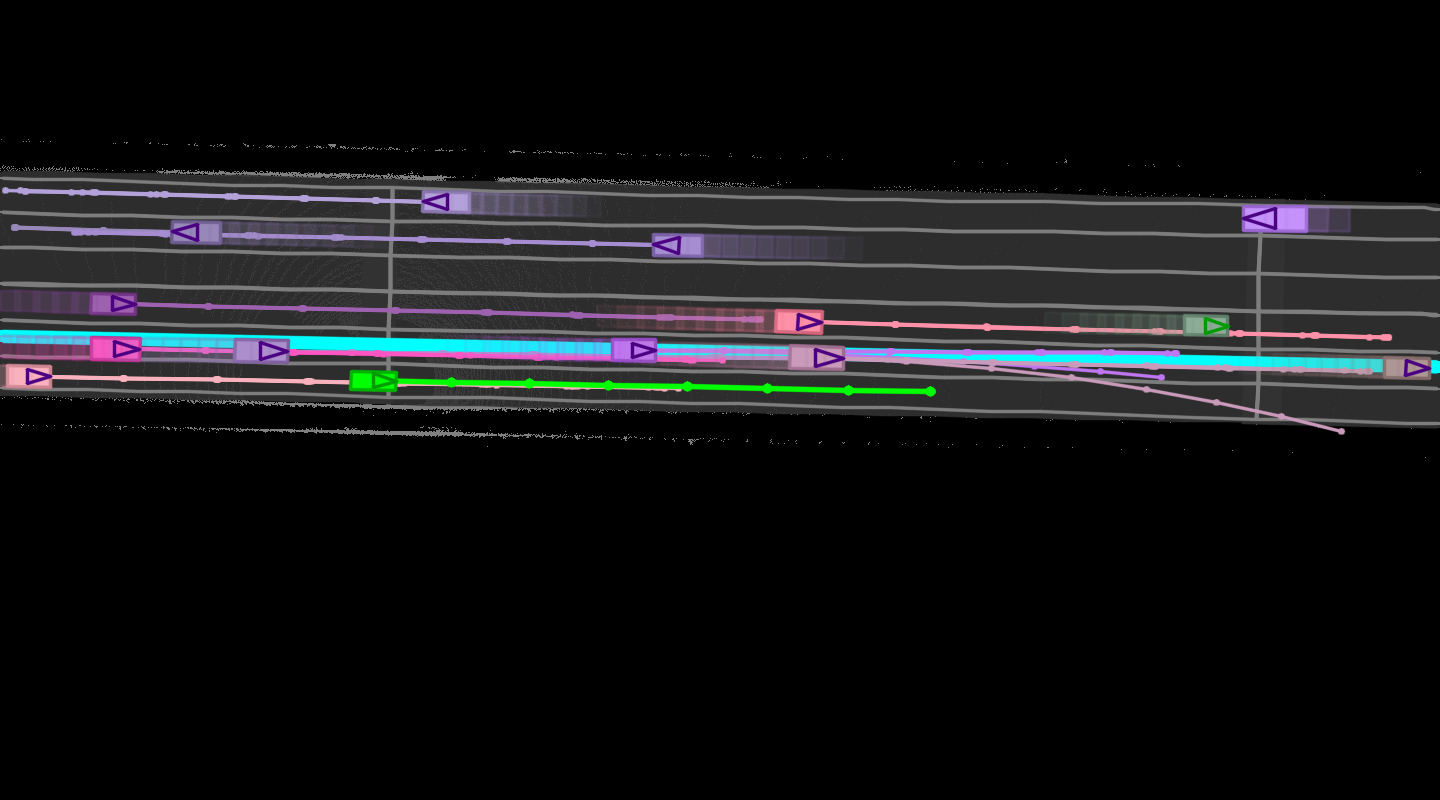}{300 275 500 225}{Non-Reactive, t=0s} &
	\figcap{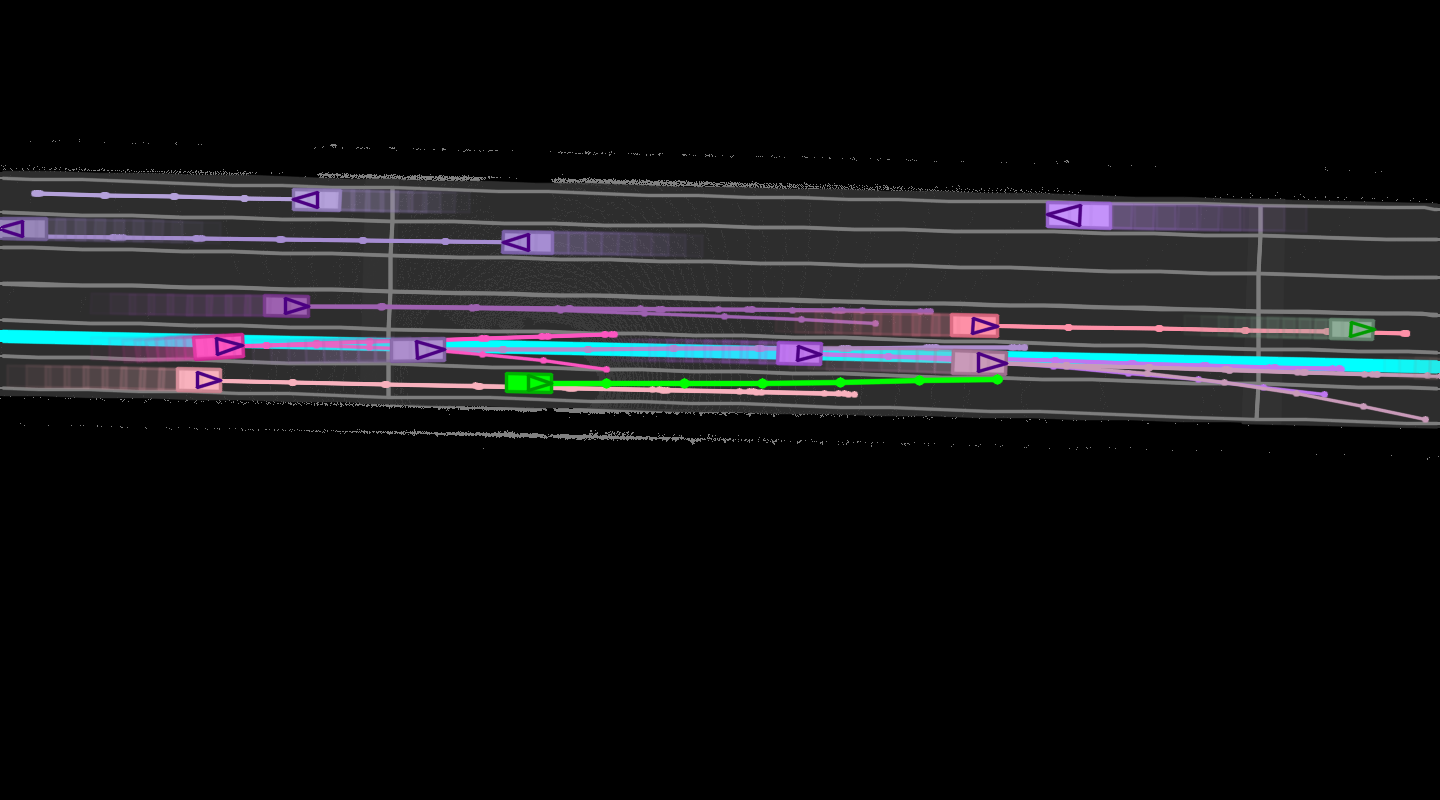}{300 275 500 225}{Non-Reactive, t=1s} &
	\figcap{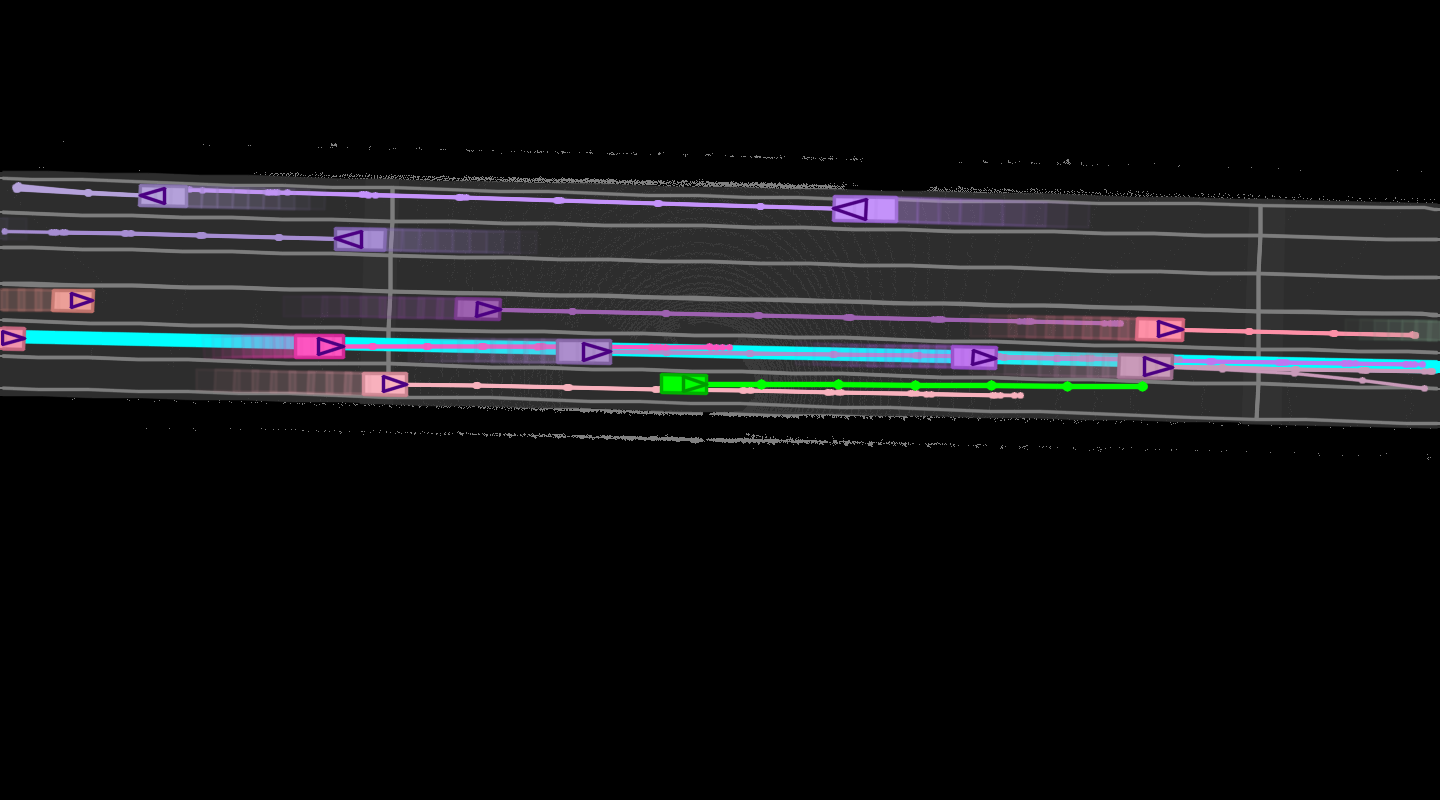}{300 275 500 225}{Non-Reactive, t=2s} &
	\figcap{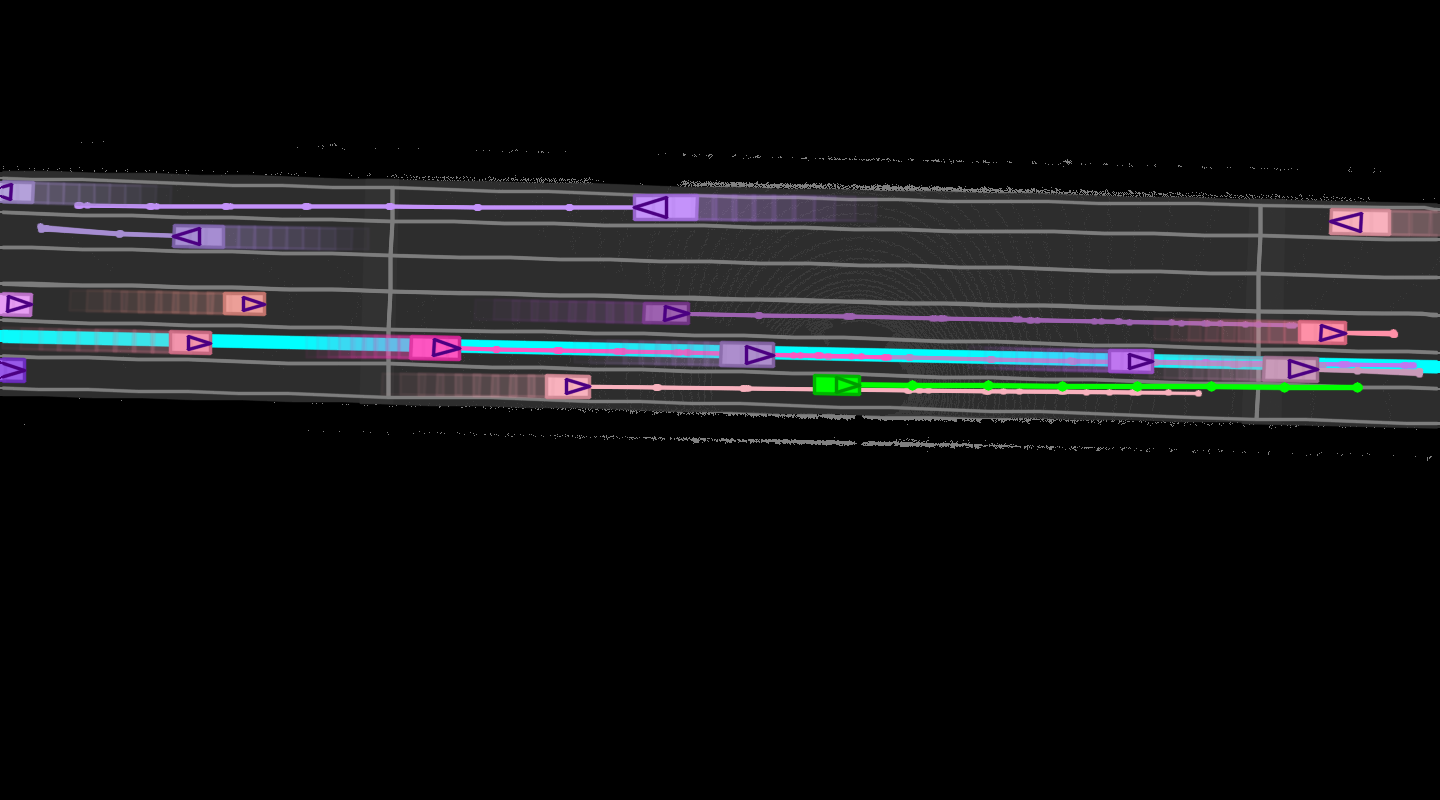}{300 275 500 225}{Non-Reactive, t=3s} \\
	\end{tabular}
	\caption{Visualization of a Simba lane merge for non-reactive (bottom) and reactive (top) models at 3 different time steps: 1s (left), 2s (middle), 3s (right).  AV is in green, other actors are in blue/purple, and goal lane is in cyan. The reactive model is able to decisively complete the lane merge, while the non-reactive model is not.   
		%	\caption{Visualization of a CARLA lane merge for non-reactive (bottom) and reactive (top) models at 3 different time steps: 1s (left), 2s (middle), 3s (right).  AV is green, planned trajectory orange. Other actors in blue/purple, with estimated heading in red. Goal lane in cyan.  
		%\wenyuan{I think it might be
		%better to overlay the lane boundary on the lidar viz, it's a bit hard to
		%understand where is the road and what would be a reasonable behavior} 
	}
	\label{fig:supp_qual_carla1}
\end{figure*}

\subsection{Learning} \label{sec:training}
%(TODO: fix, make more concise, less mathematical)

We train our joint structured model given observed ground-truth  trajectories for the ego-car and all other agents in the scene.  
We want to learn the model energies such that they induce both optimal plans for the ego-agent and accurate probabilities for the actor behaviors. 
Since the model energies induce a probability distribution used in our prediction model, this implies that minimizing the cross-entropy between our predictive distribution and the ground-truth trajectories will also learn a good set of costs for planning.
To this end, we minimize  the following cross-entropy loss function: 
%\raquel{give first intuitions of what this is going to be, then put the match}
\begin{small}
	\begin{align}
	\mathcal{L} &= \sum_{i} \mathcal{L}_{i} + \sum_{i,j}\mathcal{L}_{i,j} \\
	\mathcal{L}_{i} &= \frac{1}{K} \sum_{\by_i \notin \Delta(\by_i^*)} { p_{\text{g.t.}}(\by_i) \log p(\by_i, \cX; \bw) }  \\
	\mathcal{L}_{\by_i, \by_j} &= \frac{1}{K^2}  \sum_{\by_{i} \notin \Delta(\by_i^*), \by_{j} \notin \Delta(\by_j^*)} {p_{\text{g.t.}}(\by_i, \by_j) \log p(\by_i, \by_j, \cX; \bw) }
	\end{align}
\end{small}
where $p_{\by_i}$ and $p_{\by_i, \by_j}$ represent the marginal and pairwise marginal probabilities for every actor including the ego-agent,  and $p_{\text{g.t.}}$ represents the indicator function that is zero everywhere unless $\by_i, \by_j$ are equal to the ground-truth $\by_i^*, \by_j^*$. Recall that the marginal probabilities, $p_{\by_i}$ and $p_{\by_i, \by_j}$ for every actor including the ego-agent are computed through Loopy Belief Propagation, a
differentiable iterative message-passing procedure. Note that our method has a subtle but important distinction from raw cross-entropy loss: $\Delta(\by_i^*)$ is defined as the set of $k$ non-ground-truth trajectories for actor $i$ closest to $\by_i^*$ by $\ell_2$ distance, and we only compute the cross-entropy loss for trajectories outside of this set. 
%\raquel{previous sentence is confusing. Can you rephrase?}
We adopt this formulation since  any trajectory within $\Delta$ can reasonably be considered as a ground-truth substitute, and hence we do not wish to penalize the probabilities of these trajectories.

\begin{table*}[] 
	\centering
	\scalebox{1.0}{
	\begin{tabular}{@{}lllllllll@{}} 
		\toprule
		Carla & DESIRE \cite{lee_desire} & SocialGAN \cite{gupta_socialgan} & R2P2 \cite{rhinehart_r2p2} & MultiPath \cite{chai_multipath} & ESP \cite{rhinehart_precog} & MFP \cite{tang_mfp} &  DSDNet \cite{zeng_dsdnet} & Ours \\ \midrule
		Town1 & 2.422 & 1.141  &  0.770   &   0.680 &  0.447   &   0.279  &  \textbf{0.195} & 0.210 \\
		Town2 &  1.697 &   0.979  & 0.632 &   0.690  &  0.435 &   0.290  & 0.213 & \textbf{0.205} \\ \bottomrule
	\end{tabular}
	}

	\caption{CARLA prediction performance (minMSD, K=12). }
%		\raquel{I have split table into two so that we can see} 
%		\raquel{use same number of decimals for multipath too}} 
	\label{tab:exp_openloop_carla}
\end{table*}

\section{Experiments} \label{sec:experiments}

We  demonstrate the effectiveness of our reactive planning objective
in two closed-loop driving simulation settings: \textit{real-world traffic
scenarios} with our in-house simulator (Simba), and \textit{synthetically
generated dense traffic} with the open-source CARLA simulator \cite{dosovitskiy_carla}. We setup a large
number of  complex and highly interactive  scenarios from lane changes and (unprotected) turns - in order to tease apart the differences between reactive and non-reactive models. 
%\wenyuan{does it mean two logs? which seems small for a
%validation set. If it's hard to make the val set bigger, maybe we should somehow
%rephrase the sentence here.} 
%We use two scenarios
%each from Simba/CARLA as a validation set to identify optimal
%We use 50 scenarios each (25  turn, 25 merge) from Simba/CARLA as a validation set, and 100 CARLA scenarios / 250 Simba scenarios as our test set. 
%to identify optimal
%values of $\lambda_b, \lambda_c$ (ego-agent/actor weight and unary actor weight)
%for both non-reactive/reactive models. 

To better showcase the importance of reactivity, we created a non-reactive variant of our model by defining the planning costs as follows: $	f_{\text{nonreactive}} =  \Exp_{\cY_r \sim p(\cY_r | \cX; \bw)}[C(\cY, \cX; \bw)]$, which uses a prediction model unconditioned on the SDV trajectory. This non-reactive assumption leads to a simplification of the joint cost $\Exp_{\cY_r \sim p(\cY_r | \cX; \bw)}[C(\cY, \cX; \bw)] = C_{\text{traj}}(\by_0, \cX; \bw) + \Exp_{\cY_r \sim p(\cY_r | \cX; \bw)}[\sum_{i=1}^N{C_{\text{inter}}(\by_0, \by_i)]}$, where the considered terms are just the SDV-specific trajectory cost and the SDV/actor interaction cost. 

%\raquel{talk about the non-reactive baseline here}
%We created a non-reactive variant of our model by modeling \raquel{add equation} and say one sentence about if ther eis any changes in learning or inferece

Our results show  a key insight: 
%at our optimal $(\lambda_b, \lambda_c)$, 
our pure reactive model
alone achieves a higher success rate compared to the non-reactive model without
trading off collision rate, implying it is able to effectively consider the reactive behavior of other 
actors and formulate a goal-reaching plan without being unreasonably aggressive. Moreover, we justify the choice of a deep structured model by demonstrating that when our model is used for actor trajectory prediction,  it is competitive with the state-of-the-art in both CARLA and Nuscenes \cite{caesar_nuscenes}.  

\subsection{Experimental Setup} 
\subsubsection{Training Datasets}
Since our closed-loop evaluations are in Simba and CARLA, our models are trained on the respective datasets in the corresponding domains. Our Simba model is trained on a large-scale, real-world dataset collected through our self-driving vehicles in numerous North American cities, which we call \textbf{UrbanCity}. The dataset consists of over 6,500 snippets of approximately 25 seconds from over 1000 different trips,  with 10Hz LiDAR and HD map data, which are  included as  input context into the model $\cX$ in addition to the past trajectories per actor.
%\raquel{for our data is raw sensor data? I thought you were passing trajectories as well. Explain}
Meanwhile, the CARLA simulated dataset is a publicly available dataset  \cite{rhinehart_precog}, containing 60k training sequences. The input to the model for CARLA consists of rasterized LiDAR features and 2s of past trajectories to predict 4s future trajectories.
%Note that we also perform evaluations on the test set splits of UrbanCity and CARLA for prediction tasks - these metrics and results are discussed in \ref{sec:exp_openloop}. 

\subsubsection{Simulation Setup}
%We evaluate the reactive and non-reactive planning objectives of our model on two different simulation environments: an in-house simulated environment, called \textbf{Simba}, as well as the \textbf{CARLA} simulator \cite{dosovitskiy_carla}.
%\wenyuan{this sentence is redundant, you've said this in the very beginning.} 
Simba runs at 10Hz and leverages a realistic LiDAR simulator
\cite{manivasagam_lidarsim} to generate LiDAR sweeps around the
ego-agent at each timestep. HD map polygons are available per scenario. We first setup 12 different  interactive ``template" scenarios: we select these templates by analyzing logs in the validation set of UrbanCity and selecting a start time where there is a high degree of potential interactions with other actors. We set a goal state for the ego-agent, which for instance can be a turn or lane merge, and initialize actor positions according to their start time positions in the log. We then proceed to generate 25 distinct scenarios per template scenario by perturbing the initial position and velocity of each actor, for a total of 50 val/250 test scenarios.  During simulation each actor behaves according to a heuristic car-following model that performs basic hazard detection.

%\wenyuan{what's the
%difference between template scenario and scenario? maybe we should define them
%more concretely, right now it's a bit vague. you just say templates are from
%real-world datasets, but I guess it's not enough for the reader to understand
%what it is concretely}, initialized from a real-world, ground-truth driving sequence in UrbanCity. 
%From these template scenarios, we generate 240 total scenarios by perturbing the initial states of each vehicle actor (position, velocity).
%\wenyuan{what would be the lane / route of each of these actors? I assume the
%route come from the template? I may suggest define templates by explaining what
%information is included in a template, and then the rest of your setup would be
%clear I guess.}
% such that their reactive behavior is consistent with the reactivity assumed by our model. 
%\wenyuan{I'm not sure if this explanation for actor behavior is clear enough. Is it ACC or some variants of
%ACC?}
%\jerry{Yeah it is ACC, we can put more details in the supp.}

%Similarly,  from which we query the closest simulated
%LiDAR sweep as input. 
In CARLA we leverage the synthetic LiDAR sensor as input.  
Rather than initializing scenarios through ground-truth
data, we manually create 6 ``synthetic" template scenarios containing dense traffic, and spawn
actors at specified positions with random perturbations.
%\wenyuan{what is a initial position? do you mean you
%manually specify the actors' initial position?} 
We extend the \texttt{BasicAgent} class given in CARLA 0.9.5 as an actor model per agent, which performs basic route following to a goal and hazard detection.  We generate 50 val/100 test scenarios by perturbing the initial position / vehicle type / hazard detection range of each actor.  

Scenarios in all settings are run for a set timer. The scenario completes if 
1) the ego-agent has reached the goal, 2) the timer has expired, or 3) the ego-agent has collided. 
%1) the ego-agent has completed the task, \wenyuan{what does complete task means? achieving the goal?} 

%\subsubsection{Goal State} 
%We specify a goal state $\mathcal{G}$ in each scenario and encourage the
%ego-agent to reach the goal state. The goal state can take on different forms:
%in the case of a turn, $\mathcal{G}$ is either a target position or a route
%trace. \wenyuan{Do you mean you actually have these two set of target in
%experiments? or just it can be any one of them? If it's the later case, I
%suggest just mention which one you use. If it's the former one, why do you use
%two different forms} In the case of a lane change, $\mathcal{G}$ is a set of two points capturing the direction and position of the lane in continuous coordinates. We define a goal unary cost term for a given plan $y_e$ given $\mathcal{G}$, $C(y_e, X, \mathcal{G})$: if $\mathcal{G}$ is a single point, the unary is the L2 distance of the final waypoint; if $\mathcal{G}$ represents a lane, the unary represents the average projected distance to the lane goal vector. We append the goal unary cost to the conditional planning objective.
%\wenyuan{Just a random thought. maybe we can just integrate this goal part to
%your model section? not sure.}

\subsubsection{Closed-Loop Metrics}
The output metrics include: 1) Success Rate (whether the ego-agent successfully
completed the lane change or turn), 2) time to completion (TTC), 3) collision rate, 4)
number of actor brake events. This information is provided in CARLA but not in Simba. 

\subsection{Reactive/Non-Reactive Simulation Results} 
%\wenyuan{I feel like I'm a bit lost in terms of what would be the take home
%message you want to present. Table1 and 2 basically saying reactive is more good
%at reaching goal but more aggressive while non reactive is more conservative.
%But this seems not a very novel contribution or very important property of your
%model. To me, the Table III is more interesting in the sense although there is
%tradeoff between reactive and non-reactive, your model can be used in an
%interpolated way and find out the sweet point in the spectrum. My understanding
%is that this would be a more valuable take home message and probably we should
%highlight this such as put section C to the first.}
%Crucially, we discover that when averaged across higher values of $\lambda_b$ and $\lambda_c$ ($\lambda_b \geq 20, \lambda_c \geq 8$), the pure reactive model outperforms the non-reactive model on success rate, time to completion, goal distance, and even collision rate (Tab. \ref{tab:overall_sim}). 
%Crucially, we discover that at our operating points of $\lambda_b$ and $\lambda_c$, 
The pure reactive model outperforms the non-reactive model on
success rate, time to completion, goal distance, with no difference in collision
rate (Tab. \ref{tab:overall_sim}), on both Simba and CARLA. 
%\wenyuan{can we also show variance in Table I ?}
% \jerry{will work on it} 
%This is surprising because one would expect the reactive model to be more aggressive at equivalent hyperparameters, and indeed this is true for lower values of $\lambda_b, \lambda_c$ in CARLA (see Fig. \ref{fig:vary_hyp}, Tab. \ref{tab:interpolate}). 
%The results imply that when the ego-agent places sufficiently high weight on the
%pairwise costs and actor unary costs, it can still more efficiently navigate
%scenarios while maintaining safety. 
%\wenyuan{I think we don't need to talk about 
%  'sufficiently high weights'. we can just say this implies our reactive model,
%  when considering the reactiveness of other actors, can more efficiently
%  navigate to the goal in a highly-interactive environment, without aggressive
%  behaviros, e.g. higher collision rate.}
This implies that by considering the reactivity of other actors in its planning objective, the reactive model can more efficiently navigate to the goal in a highly-interactive environment, without performing overly aggressive behaviors that would result in a higher collision rate. 
Moreover, we also note that both the reactive/non-reactive models within our
joint structured framework outperform a strong joint prediction and planning
model,  PRECOG \cite{rhinehart_precog} -- we present our PRECOG implementation and visualizations in supplementary material. 
%additional discussion on PRECOG results is deferred to supplementary. \raquel{why is this defer? what is defer specifically?}

\begin{table}[] 
	\centering
	\scalebox{0.8}{
		\begin{tabular}{@{}lllllll@{}}
			\toprule
			Nuscenes & KDE \cite{lee_desire} & DESIRE \cite{gupta_socialgan} & SocialGAN \cite{rhinehart_r2p2} & R2P2 \cite{chai_multipath} & ESP \cite{rhinehart_precog} & Ours \\ \midrule
			5 agents &  52.071 &   6.575  & 3.871 &   3.311  &   2.892 &  \textbf{ 2.610} \\ \bottomrule
	\end{tabular}}

	\caption{Nuscenes prediction performance (5 nearest, minMSD, K=12).} 
%		\raquel{see that. caption is correct}} 
	\label{tab:exp_openloop_nusc}
\end{table}

\subsection{Qualitative Results}
To complement the quantitative metrics, we provide scenario visualizations. In Fig. \ref{fig:supp_qual_carla1}, we present a lane merge scenario  in Simba to better highlight the difference between the reactive and non-reactive models in a highly complex, interactive scenario. We provide simulation snapshots at $t=0,1,2,3$ seconds. 
Note that the reactive model is able to take decisive action and complete the lane merge; the neighboring actor slows down with adequate buffer to let the ego-agent pass. Meanwhile, the non-reactive agent does not complete a lane merge but drifts slowly to the left side of the lane over time. We provide several more comparative visualizations of various scenarios in both Simba and CARLA in our supplementary document and video.  

\subsection{Prediction Metrics} 
To validate the general performance of our joint structured framework, we
compute actor predictions using our model, by using Loopy Belief Propagation to
compute unconditioned actor marginals $p(\by_{i})$, 
%\raquel{I change to new notation} 
and  compare against state-of-the-art on standard prediction benchmarks in Fig. \ref{tab:exp_openloop_carla}, \ref{tab:exp_openloop_nusc}: the CARLA PRECOG dataset and
the Nuscenes dataset \cite{caesar_nuscenes} (note that a separate model was trained for
Nuscenes). We report minMSD \cite{rhinehart_precog}, the minimum mean squared
distance between a sample of predicted/planned trajectories and ground-truth as metric. As shown, our method is competitive with or
outperforms prior methods in minMSD. Similar to the findings of DSDNet
\cite{zeng_dsdnet}, this implies that an energy-based model relying on discrete
trajectory samples per actor is able to effectively make accurate trajectory predictions for each actor. 
%\wenyuan{the MSD metric itself doesn't necessarily imply
%interactions? maybe just saying able to make accurate / sota prediction}

%\raquel{you forgot to link to the tables where the results are shown}

%\wenyuan{shall we also report dsd result on carla, especially it's
%a paper from the same lab and this is a single blind... I think it's fine to
%still say on-par / outperform as it's very close and on Nusc you outperform.}

\subsection{Training Loss Functions}
\begin{table}[]
	\centering
	\scalebox{0.95}{
		\begin{tabular}{ccccc}
			\toprule
			 Loss & Pred FDE (3s) & Actor CR & Plan FDE (3s) & Ego CR \\ \midrule
			Cross-entropy   & 1.47 & 0.55\% & 2.01 & 0.24\% \\
			Chen et al. \cite{chen_deepstruct} & 1.30 & 0.67\% & 2.10 & 0.22\% \\ 
			Ours & \textbf{1.24} & \textbf{0.44\%} & \textbf{1.73} & \textbf{0.18\%}  \\
			\bottomrule
	\end{tabular}}
	\caption{ Ablation Study comparing training losses on UrbanCity. For all metrics, lower is better.}
	 \label{tab:train_ablate}
\end{table}
We also perform an ablation study on the UrbanCity validation set to analyze our proposed training loss function compared against vanilla cross-entropy loss (no ignore set), as well as the approach in Chen et al. \cite{chen_deepstruct}. We demonstrate in Fig. \ref{tab:train_ablate} that our approach achieves the lowest Final Displacement Error (FDE), for the SDV and other actors, as well as the lowest collision rate between actor collisions and collisions of the ego-agent with ground-truth actors. 

\section{Conclusion} \label{sec:conclusion}
%We presented a novel joint structured model unifying prediction and planning, as well as a principled objective allowing the ego-agent to model reactive behaviors of other actors during its own planning.
%We demonstrated the validity of our structured model through state-of-the-art results on standard prediction datasets; moreover we demonstrate that such an objective gives the ego-agent a wider array of options to achieve goal states compared to non-reactive objectives. In future work, we plan on exploring how to better learn how to better adapt reactivity to the given scenario, potentially in a reinforcement learning setting.  

We have presented a novel \textit{reactive} planning objective allowing the ego-agent to jointly reason about its own plans as well as how other actors will react to them. We formulated the problem with a  deep energy-based model which enables us to  explicitly model trajectory goodness as well as interaction cost between actors. Our experiments showed that our reactive model outperforms the non-reactive model in various highly interactive simulation scenarios without trading off collision rate. Moreover, we outperform or are competitive with state-of-the-art in prediction metrics.

%We demonstrate multiple key results: 1) that with validated hyperparameters, the reactive model outperforms the non-reactive model in simulated interactive scenarios, and 2) that our joint structured model outperforms state-of-the-art on standard prediction datasets. In future work, we plan on exploring how to better learn how to better adapt reactivity to the given scenario, potentially in a reinforcement learning setting.  

%We demonstrated the validity of our structured model through state-of-the-art results on standard prediction datasets; moreover we demonstrate that such an objective gives the ego-agent a wider array of options to achieve goal states compared to non-reactive objectives. 
%\ersin{add a few sentences for future work}

%%%%%%%%%%%%%%%%%%%%%%%%%%%%%%%%%%%%%%%%%%%%%%%%%%%%%%%%%%%%%%%%%%%%%%%%%%%%%%%%

%%%%%%%%%%%%%%%%%%%%%%%%%%%%%%%%%%%%%%%%%%%%%%%%%%%%%%%%%%%%%%%%%%%%%%%%%%%%%%%%

%%%%%%%%%%%%%%%%%%%%%%%%%%%%%%%%%%%%%%%%%%%%%%%%%%%%%%%%%%%%%%%%%%%%%%%%%%%%%%%%

%%%%%%%%%%%%%%%%%%%%%%%%%%%%%%%%%%%%%%%%%%%%%%%%%%%%%%%%%%%%%%%%%%%%%%%%%%%%%%%%

\bibliographystyle{splncs04}
\bibliography{egbib}

\clearpage

\appendix
\section{Model Details} 

In this section, we provide more precise model details regarding our joint structured model. Specifically, we first detail the dataset-dependent input representations used by our model (Sec. \ref{sec:inp_details}). We then present the architecture details of our network, which predicts actor-specific and interaction energies between actors (Sec. \ref{sec:arch_details}, Sec. \ref{sec:inter_energy}). This also includes details regarding our discrete trajectory sampler (Sec. \ref{sec:traj_sampler}).

\subsection{Input Representation} \label{sec:inp_details}
As mentioned in the main paper, we assume that the trajectory history of the other actors, including their bounding box width/height and heading, are known to the ego-agent at the given timestep. Hence, we directly feed the trajectories of all actors to our model, transformed to the current ego-agent coordinates. 

In addition, we add dataset-dependent context to the model. In UrbanCity and Nuscenes, we use both LiDAR sweeps and HD maps as context. Meanwhile, we directly use the input representation provided by the CARLA dataset \cite{rhinehart_precog}, which contains a rasterized LiDAR representation but no map data. 

\subsubsection{UrbanCity}
We use a past window of 1 second at 10Hz as input context, and use an input
region of $[-72, 72] \times [-72, 72] \times[-2, 4]$ meters centered around the
ego-agent. From this input region, we collect the past 10 LiDAR sweeps (1s),
voxelize them with a $0.2 \times 0.2 \times 0.2$ meter resolution, and combine
the time/z-dimensions, creating a $720 \times 720 \times 300$ input tensor. We
additionally rasterize given HD map info with the same resolution. The HD maps
include different lane polylines, and polygons representing roads,
intersections, and crossings. We rasterize these information into different
channels respectively.

\subsubsection{Nuscenes} 
The details are similar to UrbanCity. The main difference is that the input region is sized $[-49.6, 49.6] \times [-49.6, 49.6] \times [-3, 5]$ meters, and the voxel resolution is $0.2 \times 0.2 \times 0.25$ meters, creating a $496 \times 496 \times 320$ input LiDAR tensor. 

\subsubsection{CARLA} 
We use the input representation directly provided by the CARLA PRECOG dataset \cite{rhinehart_precog}, which consist of $200 \times 200 \times 4$ features derived from rasterized LiDAR. Each channel contains a histogram of points within each cell at a given z-threshold (or for all heights). Additional details can be found in 
\urlstyle{tt} \url{https://github.com/nrhine1/precog_carla_dataset/blob/master/README.md#overhead_features-format}.

\subsection{Network Architecture Details}  \label{sec:arch_details}  
Here, we present additional details of the actor-specific and interaction energies of our model. As mentioned in the main paper, the actor-specific energies are parameterized with neural nets, consisting of a few core components. The first component is a \textbf{backbone network} that takes in the input representations to compute intermediate spatial feature maps. Then, given the past trajectory for each actor, we sample $K$ trajectories for each actor using a \textbf{realistic trajectory sampler}, given in Sec. \ref{sec:traj_sampler}. These future actor trajectory samples, as well as the past actor trajectories and backbone feature map, are in turn passed to our \textbf{unary module} to predict the actor-specific energies for each actor trajectory. Finally, our \textbf{interaction energy} is determined by computing collision and safety distance violations between actor trajectories. 
\subsubsection{Backbone Network}
Given the input representation, we pass it through a backbone network to compute intermediate feature maps that are spatially downsampled from the input resolution. This backbone network is inspired by the detection networks in \cite{yang_pixor, zeng_nmp, zeng_dsdnet}. This network consists of 5 sub-blocks, each block containing $[2,2,3,6,5]$ Conv2d layers with $[32,64,128,256,256]$ output channels respectively, with a 2x downsampling max-pooling layer in front of the 2nd -- 4th blocks. The input is fed through the first 4 blocks, generating intermediate feature maps of different spatial resolutions. These intermediate feature maps are then pooled/upsampled to the same resolution (4x downsample from input), and then fed to the fifth block to generate the final feature map (at 4x downsample from the input). Each convolution is followed by a BatchNorm2d and ReLU layer. 
%\wenyuan{do you have different backbone arch on Carla? or it's the same}
% \jerry{it's the same.}

\subsubsection{Unary Module}
We broadcast the past trajectory per actor with their $K$ future trajectory samples to create a $N \times K$ matrix of concatenated trajectories. The purpose of the unary network is then to predict the actor-specific energy for each actor trajectory sample. 

We first define a Region of Interest (ROI) centered on each actor's current
position and oriented according to each actor's heading. In UrbanCity/Nuscenes,
we define the ROI to be $12.8 \times 12.8$ meters. In CARLA we define it to be much bigger at $100 \times 100$ meters due to the absence of map data in the training set.
%\wenyuan{I
%think we shall mention why carla is special. maybe say we found on dataset with
%map, larger ROI doesn't help where as on carla(without map) larger ROI help.}). 
We then use this rotated ROI to extract a corresponding feature per actor from the backbone feature map. We then obtain a 1D representation of this ROI feature per actor by feeding it through an MLP consisting of 6 Conv2d layers with $[512, 512, 1024, 1024, 512, 512]$ output filters with 2x downsampling in the 1st, 3rd, 5th layers, and an adaptive max-pool collapsing the spatial dimension at the end. 

We additionally extract \textit{positional embeddings} for each actor trajectory
sample, at each timestep, by indexing the backbone feature map at the
corresponding sample position at the given timestep, with bilinear interpolation. This extracts a $N \times K \times T \times 512$ dimensional tensor, where $T$ represents the total horizon of the trajectory (both past and future timesteps), and $512$ is the channel dimension of the backbone feature map. We collapse the tensor into a 1D representation as well: $N \times K \times (T * 512)$. 

Finally, we directly encode the trajectory information per timestep into a trajectory feature, consisting of $[\Delta_x, \Delta_y, \cos(\Delta_{\theta}), \sin(\theta), \Delta_d]$, where $\Delta_x, \Delta_y$ represent the displacement in x,y directions from the previous timestep, $\Delta_d$ represents displacement magnitude, and $\theta$ represents current heading. The trajectory feature, positional embeddings, and 1-D ROI feature are all concatenated along the channel dimension to form a final unary feature. It is fed to a final MLP consisting of 5 fc layers of $1024, 1024, 512, 256, 1$ output channels respectively, and the output is a $N \times K$ matrix of unary energies.  

\subsection{Interaction Energies} \label{sec:inter_energy}
Our interaction energy is a pairwise energy between two actor trajectory samples that contains two non-learnable components: collision cost and safety distance cost. The collision detection is efficiently implemented using a GPU kernel computing the IOU among every pairwise sample from two actors, given their timestep, positions, bounding box width/height, and heading. The output is a $N \times N \times K \times K$ matrix, where a given entry is 1 if the trajectory sample of actor $i$ and trajectory sample of actor $j$ collide at any point in the future, and 0 if not. The collision energy is only computed over future samples, not past samples. 

Similarly, the safety distance violation is computed over all future actor
samples. For a given pairwise sample between actor $i$ and actor $j$ at a given
timestep, the distance from the center point of actor $i$ is computed to the
polygon of actor $j$ (minimal point-to-polygon distance). If the distance is within a given safety threshold, then the energy is the squared distance of violation within the threshold. Note that unlike collision energy, this matrix is not symmetric. This choice was made to use a GPU kernel that efficiently computes point distances to polygons in parallel. 

%Finally, the learnable component of the network comes from using the concatenated unary features. Each unary feature per actor sample is fed to an MLP with 5 fc layers of $1024, 1024, 512, 256, 40$ output dimensions, creating a 40-dimensional embedding. Doing a pairwise multiplication among all samples, multiplied by a learnable scaling factor, creates a final $N \times N \times K \times K$ matrix of learnable pairwise energies. 

\subsection{Trajectory Sampler Details} \label{sec:traj_sampler} 
We follow the discrete trajectory sampler used in \cite{zeng_nmp, zeng_dsdnet}. The sampler first estimates the initial speed/heading of each actor given the provided past trajectory. From these values, the sampler samples from three trajectory modes: a straight line, circular trajectory, or spiral trajectory with $[0.3, 0.2, 0.5]$ probability. Within each mode, the control parameters such as radius, acceleration are uniformly sampled within a range to generate a sampled trajectory. We use 50 trajectories per actor (including the ego-agent) for CARLA, and 100 trajectories per actor on UrbanCity and Nuscenes. Additional details can be found in \cite{zeng_nmp, zeng_dsdnet}.

\section{Model Properties}

In this section, we discuss some additional properties of our model. First, we discuss the reasons and tradeoffs for excluding the actor-actor interaction term in the reactive objective (Sec. \ref{sec:exclude_actor_actor}). Next, recall that in addition to our reactive objective, we also implemented a non-reactive planning objective as a baseline: $f_{\text{nonreactive}} =  \Exp_{\cY_r \sim p(\cY_r | \cX; \bw)}[C(\cY, \cX; \bw)]$. We demonstrate that we can \textit{flexibly interpolate} between the non-reactive and reactive objectives within our deep structured model by varying the size of the conditioning set in the prediction model of the reactive objective (Sec. \ref{sec:interpolate}). Experimental results showcasing this behavior are demonstrated in Sec. \ref{sec:exp_interpolate}).  Moreover, we demonstrate that the non-reactive objective under our joint structured model is related to maximizing the marginal likelihood of the ego-agent, marginalizing out other actors (Sec. \ref{sec:nonreact_marg}). 

\subsection{Excluding the Actor-Actor Interaction Term} \label{sec:exclude_actor_actor}
We mention in Sec. III-B of the main paper that we exclude the actor-actor interaction term as a cost from our reactive planning objective. The primary reason for this is due to computational reasons. To illustrate this, we distribute the full expectation in the objective over the costs: 
{\small
	\begin{align}
	f_{\text{reactive}}  &= C_{\text{traj}}^{\by_0}+ \Exp_{p_{\cY_r | \by_0}}[\\
	&\sum_{i=1}^{N}C_{\text{inter}}^{\by_0, \by_i} + \sum_{i=1}^{N}C_{\text{traj}}^{\by_i} + \sum_{i=1,j=1}^{N,N}C_{\text{inter}}^{\by_i, \by_j}] \\
	&=  C_{\text{traj}}^{\by_0} +  \sum_{i,\by_i}p_{\by_i|\by_0}C_{\text{inter}}^{\by_0, \by_i} + \sum_{i,\by_i}p_{\by_i|\by_0}C_{\text{traj}}^{\by_i}  \\
	&+ \sum_{i,j,\by_i,\by_j}p_{\by_i, \by_j|\by_0}C_{\text{inter}}^{\by_i, \by_j} 
	\end{align}
}  

First, we note that the last summation term implies computing a full $N \times N \times K \times K$ matrix (containing both the interaction cost between any pair of samples from two actors, as well as the probabilities) for every value of $\by_0$. For our values of $N, K$, one of these matrices will generally fit in memory on a Nvidia 1080Ti GPU, but additionally batching by the number of ego-agent samples (which is $N$) will not. Moreover, we note that the Loopy Belief Propagation algorithm used for obtaining actor marginals will provide marginal probabilities $p_{\by_i}$ and pairwise probabilities $p_{\by_i, \by_j}$ \cite{nowozin_tutorial} for all actor samples, which directly gives us the conditional actor marginal probabilities $p_{\by_i|\by_0}$ with one LBP pass. However, $p_{\by_i, \by_j|\by_0}$ is not readily provided by the algorithm, requiring us to run LBP for every value of $\by_0$ to obtain these conditional pairwise marginals. 

We acknowledge that the actor-actor interaction term can capture situations that the actor-specific term does not, specifically where the ego-agent's actions led to dangerous interactions between two other actors (e.g. the SDV causes a neighboring car to swerve and narrowly collide with another car). We can potentially approximate the term by only considering neighboring actors to the SDV -- we leave this for future work.

\subsection{Interpolation} \label{sec:interpolate} 
We observe an additional level of flexibility in this model: being able to
interpolate between our reactive/non-reactive objectives, which have thus far
been presented as distinct. A potential advantage of interpolation is the ability to customize between conservative, non-reactive driving to efficient navigation depending on the user's preference. 
%\wenyuan{I think here it's better if we say one
%sentence of potential benefits of interpolation. Maybe something like
%customizing from conservative driving to efficient navigation based on
%users' preference. But it's fine if @jerry doesn't wanna add.} 
Recall that our reactive objective is defined as $f  =   \Exp_{\cY_r \sim p(\cY_r | \by_0, \cX; \bw)}[C(\cY, \cX; \bw)]$, which can be simplified into 
\begin{align}
f =  C_{\text{traj}}^{\by_0} +  \sum_{i,\by_i}p_{\by_i|\by_0}C_{\text{inter}}^{\by_0, \by_i} +  \sum_{i,\by_i}p_{\by_i|\by_0}C_{\text{traj}}^{\by_i}
\end{align}
(see Sec. III-B, III-C). Similarly, our non-reactive baseline objective is defined as $f_{\text{nonreactive}} =  \Exp_{\cY_r \sim p(\cY_r | \cX; \bw)}[C(\cY, \cX; \bw)]$, and can be simplified into 
\begin{align}
f_{\text{nonreactive}} =  C_{\text{traj}}^{\by_0} +  \sum_{i,\by_i}p_{\by_i}C_{\text{inter}}^{\by_0, \by_i}
\end{align}
The key to interpolation between these two objectives lies within our conditional prediction model for a given actor $\by_i$ within the reactive objective: $p(\by_i | \by_0, \cX; \bw)$, currently conditioned on a single ego-agent plan. We can modify the conditioning to be on a set: $S^{\by_0}$ with $k, 1 \leq k \leq K$ elements, which are the top-$k$ candidate trajectories closest to $\by_0$ by L2 distance. Then, we define $p(\by_i | S^{\by_0}, \cX; \bw) = \frac{1}{Z} \sum_{\bar{\by}_0 \in S^{\by_0}} {p(\by_i, \bar{\by}_0, \cX; \bw)}$, where $Z$ is a normalizing constant. Intuitively, conditioning actor predictions on this set implies that actors do not know the exact plan that the SDV has, but may have a rough idea about the general intent. When $|S^{\by_0}|$ is 1, we obtain our reactive model. When $|S^{\by_0}|$ is $K$, it is straightforward to see that $Z=1$, and hence we obtain our actor marginals $p(\by_i | \cX; \bw)$ used in the non-reactive model. Moreover, when $|S^{\by_0}| = K$ the actor-specific cost term $\sum_{i,\by_i}p_{\by_i | S^{\by_0}}C_{\text{traj}}^{\by_i} = \sum_{i,\by_i}p_{\by_i}C_{\text{traj}}^{\by_i}$ no longer depends on the candidate SDV trajectory $\by_0$; hence we can remove it from the planing objective, which results in the non-reactive objective. 

Please see Sec. \ref{sec:exp_interpolate} for experimental results in our simulation scenarios at different conditioning set sizes.

\subsection{Non-Reactive Objective and Marginal Likelihood} \label{sec:nonreact_marg} 
%Recall from the main paper that we can also define non-reactive objective under our structured model $f_{\text{nonreactive}} =  \Exp_{\cY_r \sim p(\cY_r | \cX; \bw)}[C(\cY, \cX; \bw)]$. 
%From here, "
Additionally, it is fairly straightforward to show that the non-reactive objective is closely related to maximizing the marginal likelihood of the ego-agent. Let the marginal likelihood of the ego-agent be denoted as $p(\by_0 | \cX; \bw)$.

{\small
	\begin{align}
		\ln p(\by_0 | \cX; \bw) &= \ln \Exp_{\cY_r \sim p(\cY_r | \cX; \bw)}[p(\by_0 | \cY_r, \cX; \bw)] \\
		&>= \Exp_{\cY_r \sim p(\cY_r | \cX; \bw)}[\ln p(\by_0 | \cY_r, \cX; \bw)] \label{eq:jensens}
%		&= \Exp_{\cY_r \sim p(\cY_r | \cX; \bw)}[\ln p(\cY, \cX; \bw)] \\ 
%		&= \Exp_{\cY_r \sim p(\cY_r | \cX; \bw)}[C(\cY, \cX; \bw)]
%	f_{\text{nonreactive}} =  \Exp_{\cY_r \sim p(\cY_r | \cX; \bw)}[C(\cY, \cX; \bw)] 
%	\label{eq:nonreact_marg1}
	\end{align}
}

The lower bound in \ref{eq:jensens} is obtained through Jensen's inequality. Now, note that if we try to maximize this term through $\by_0$, we obtain our non-reactive planning objective (which is a minimization over the joint costs):  
{\small
	\begin{align}
	& \text{argmax}_{\by_0} \Exp_{\cY_r \sim p(\cY_r | \cX; \bw)}[\ln p(\by_0 | \cY_r, \cX; \bw)] \\
			&= \text{argmax}_{\by_0}  \Exp_{\cY_r \sim p(\cY_r | \cX; \bw)}[\ln p(\cY, \cX; \bw)] \\ 
			&= \text{argmin}_{\by_0}  \Exp_{\cY_r \sim p(\cY_r | \cX; \bw)}[C(\cY, \cX; \bw)]
	\label{eq:nonreact_marg2}
	\end{align}
}

This implies that maximizing the marginal likelihood of the SDV trajectory under our model can be considered as a non-reactive planner.

\section{Effects of Varying Weights on Planning Costs}
\begin{figure}
	\vspace{-5mm}
	\includegraphics[width=0.48\linewidth]{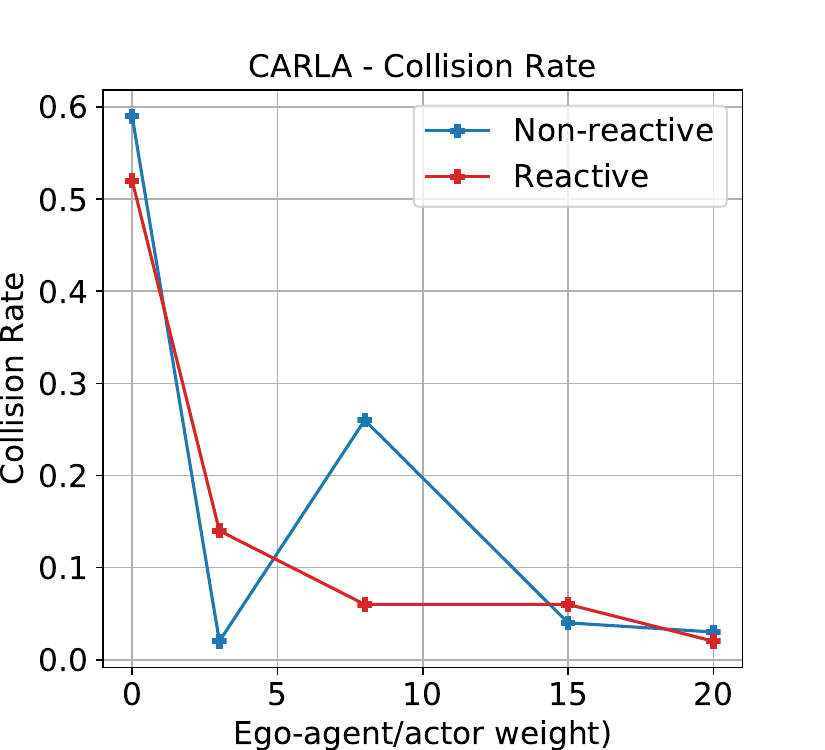}
	\includegraphics[width=0.48\linewidth]{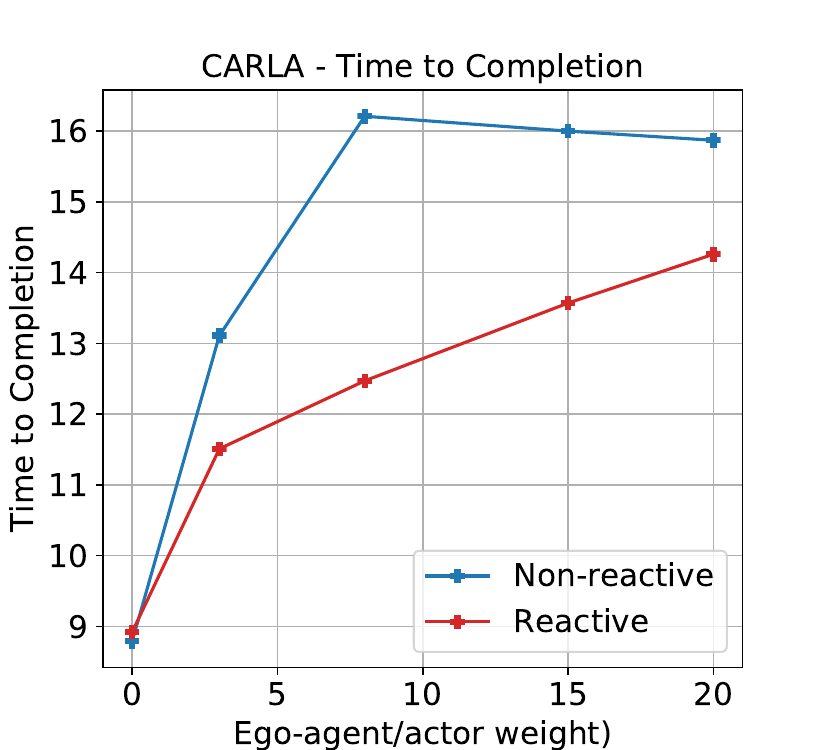} \\
	\includegraphics[width=0.48\linewidth]{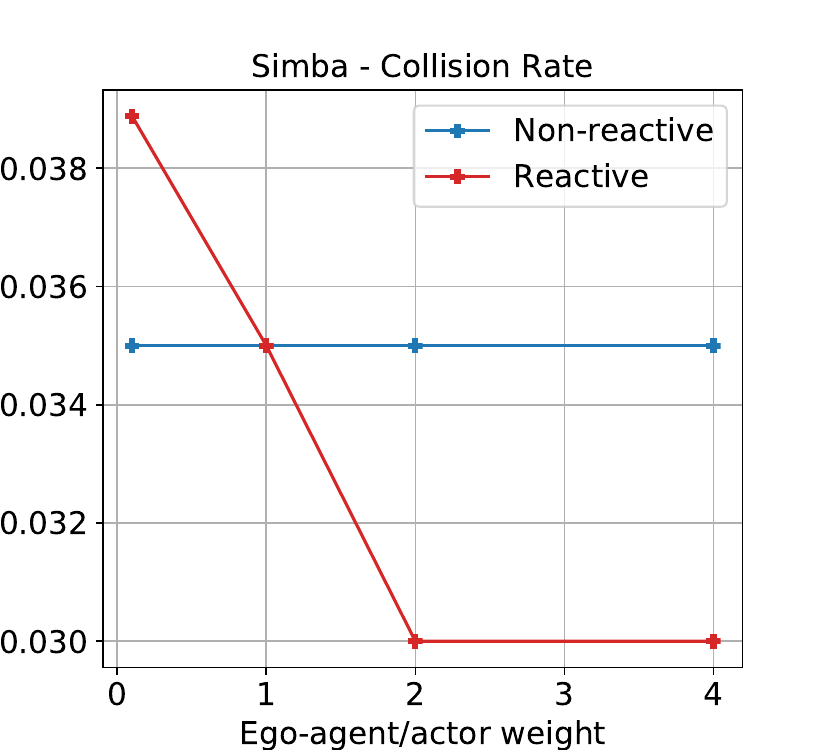}
	\includegraphics[width=0.48\linewidth]{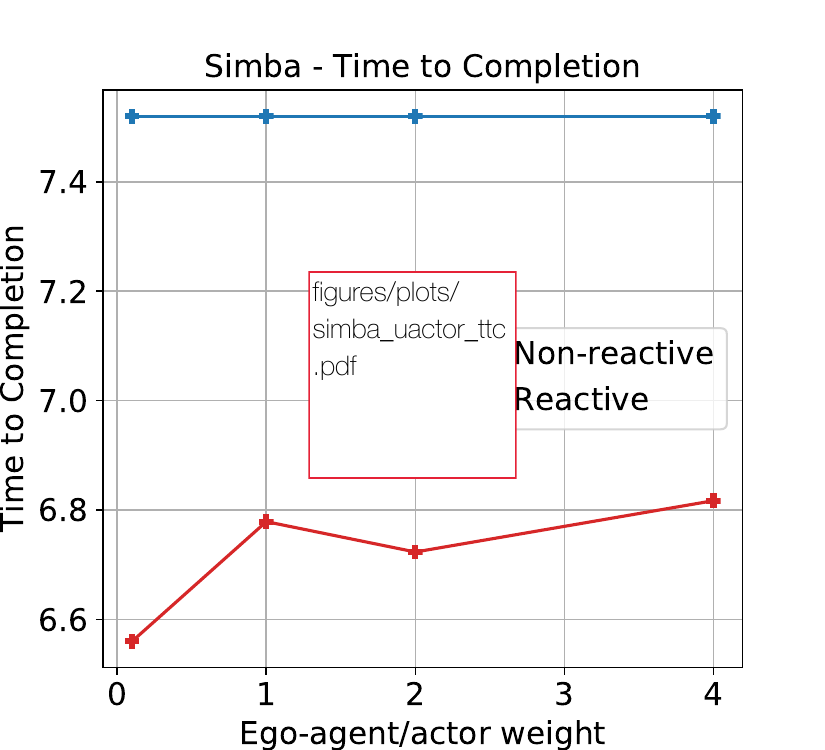} \\
	\caption{Analyzing impact of varying $\lambda_b$ (ego-agent/actor weight) in CARLA (top) and Simba (bottom).}
	\label{fig:vary_hyp}
\end{figure}

%(TODO: move this and interpolation results to supplementary )
In practice when implementing our reactive objective (Sec. III-B of the main paper), we define weights on the SDV/actor interaction  energy ($\lambda_b$) and the actor-specific energy $\lambda_c$ to more flexibly control for safety during planning. These weight values for both our reactive planner and the non-reactive baseline are determined from the scenario validation set. To provide further insight into the planning costs, we analyze the impact of varying $\lambda_b, \lambda_c$ on our closed-loop evaluation metrics. We observe that when $\lambda_b$ decreases, collision rates for both non-reactive/reactive models go up while time to completion somewhat trends down (Fig. \ref{fig:vary_hyp}, top). This is reasonable given that $\lambda_b$ directly controls the weight on the pairwise energy, which includes collision. Additionally, when varying $\lambda_c$ in Simba, we observe that while the variation in TTC is somewhat negligible for the reactive model, collision rate does trend upward while $\lambda_c$ is decreased - implying that considering the actor unary energy does have some weight in maintaining safer behavior. Of course, we also emphasize that $\lambda_c$ makes no difference in the non-reactive results, since the actor unary term can be cancelled in the non-reactive objective (see Sec. III-B in the main paper).
%$\lambda_b$ represents the weight on the interaction energy between the SDV and other actors, and $\lambda_c$ represents the weight on the actor-specific cost of other actor trajectories. 
%\wenyuan{As I mentioned in the very beginning of this section, I personally
%	still confuse about the take home message you wanna deliver here. Although not a
%	proper comparison, this somewhat give me the feeling that in a neural network
%	paper the author talk about the effect of learning rate (also a hyperparameter
%	as here) which is not very common. But maybe one thing interesting to discuss is
%	how to select/tradeoff $\lambda$ or even $S$ in the next section. Here you
%	present many observations but no conclusion is drew. I personally think having
%	something like a conclusion is better, but it's up to you.}
\section{Interpolation Results between Non-Reactive / Reactive} \label{sec:exp_interpolate}
\begin{table}[]
	\centering
	\scalebox{0.62}{
		\begin{tabular}{c|c|ccccc}
			\toprule
			Simulator & $|S^{\by_0}|$ &  Success (\%) & TTC (s) & Goal Distance (m) & Collision Rate (\%)  & Actor Brake   \\ \midrule
			CARLA & 1 (full reactive) & 72.0 & 12.8 & 2.0  & 5.0  & 41.8 \\
			& $0.2K$ & 52.0 & 14.6 & 3.3 & 1.0  &  37.7 \\
			& $0.4K$ & 42.0 & 15.2  & 3.7  & 4.0  &  37.3 \\
			& $0.6K$ & 46.0 & 14.9 & 4.7 & 5.0 &  32.8 \\
			& $0.8K$ & 52.0 & 14.5 & 4.3 & 5.0 &  36.7 \\
			& $K$ (non-reactive) & 45.0 & 16.0 & 4.4	& 5.0 &  37.1 \\
			\midrule
			Simba & 1 (full reactive)  & 82.0 & 6.8 & 4.3 & 3.5\%  & - \\
			& $0.2K$ & 73.5 & 7.5 & 5.4 & 3.5\%  &  - \\
			& $0.4K$ & 76.5 & 7.4 & 5.4 & 2.5\%  &  - \\
			& $0.6K$ & 70.5 & 7.5 & 5.2 & 3.5\%  &  - \\
			& $0.8K$ & 68.0 & 7.6 & 5.2 & 3.5\% &  - \\
			& $K$ (non-reactive) & 70.0 & 7.5 & 5.2 & 3.5\%  &  - \\
			\bottomrule
	\end{tabular}}
	\caption{Interpolating between a reactive and non-reactive model by varying the size of conditioning set $S^{y_e}$, in CARLA and Simba.}
	\label{tab:interpolate}
\end{table} 

Sec. \ref{sec:interpolate} in this document showed that we can flexibly interpolate between reactive and non-reactive objectives by increasing the size of the conditioning set $S^{\by_0}$. We demonstrate this interpolation, averaged across CARLA/Simba scenarios, in Tab. \ref{tab:interpolate}. The extremes $|S^{\by_0}|=1, |S^{\by_0}|=K$ demonstrate a tradeoff between full reactive and full-nonreactive behavior. The metrics tend to be more strongly pulled towards the non-reactive side as set size increases -- success rates trend downward and time to completion rates trend upwards, consistent with the difference in performance between the full reactive/non-reactive objectives in Tab. I in the main paper.  We additionally observe that collision rates are roughly similar across the different conditioning set sizes. While the results are not conclusive, we note that they hint at a setting where an interpolated planning objective can achieve high success rates while planning more safely compared to either extreme. 

\section{PRECOG Details} 
Here, we provide more details regarding our PRECOG implementation
\cite{rhinehart_precog}. PRECOG is a planning objective based on a conditional
forecasting model called Estimating Social-forecast Probability (ESP). We first
implement ESP and verify that it reproduces prediction metrics provided by the
authors in CARLA (also see Table III in the main paper). We then attach the
PRECOG planning objective on top. 

\subsection{ESP Architecture} 
The ESP architecture largely follows the details specified in the original
paper, with slight modifications similar to the insights discovered in
\cite{casas_ilvm}. First, we use the same whisker featurization scheme as
specified in the paper, but due to memory limitations in UrbanCity we sample
from a set of three radii $[1,2,4]$ as opposed to the original 6. Our past
trajectory encoder is a GRU with hidden state 128 that sequentially runs across
the past time dimension and takes in the vehicle-relative coordinates,
width/height, and heading as inputs. Moreover, given that our scenes can have a
variable number of actors as opposed to constant number in the original paper, we use $k$-nearest neighbors with $k=4$ to select the nearest neighbor features at every future timestep. Finally, we found that in the autoregressive model setting, training using direct teacher forcing \cite{lamb_teacherforcing} by conditioning the next state on the ground-truth current state caused a large mismatch between training and inference. Instead, we add white noise of 0.2m to the conditioning ground-truth states during training to better reflect error during inference. 

\subsection{Implementation of PRECOG objective} 

The PRECOG planning objective is given by: 
\begin{align}
\mathbf{z}^{r*} = \text{argmax}_{\mathbf{z}^r} \mathbb{E}_{\mathbf{Z}^h}[\log q(f(\mathbf{Z}) | \phi) + \log p(\mathcal{G} | f(\mathbf{Z}, \phi)]
\end{align}

where the second term represents the goal likelihood, and the first term represents the "multi-agent" prior, which is a joint density term that can be readily evaluated by the model. In order to plan with the PRECOG objective, one must optimize the ego-agent latent $\mathbf{z}^r$ over an expectation of latents sampled from other actors $\mathbf{Z}^h$. 

The joint density term can be evaluated by computing the log-likelihood according to the decoded Gaussian at each timestep $\log q(f(\mathbf{Z})) = \log q(S) = \sum_{t=1}^{T} q(\mathbf{S}_t | \mathbf{S}_{1:t-1}) = \sum_{t=1}^{T} \mathcal{N}(\mathbf{S}_t; \mathbf{{\mu}_t}, \Sigma_t)$. Meanwhile, the authors use an example of a goal state penalizing L2 distance as an example of goal likelihood (assuming a Gaussian distribution), which admits a straightforward translation into our definition of a goal energy, for both goal states and goal lanes. Hence, we use the same goal energy definition given in Sec. 4-A of the main paper to compute the goal likelihood. We weight the prior likelihood and goal likelihood with two hyperparameters $\lambda_1, \lambda_2$, which are determined from the validation sets.

In practice, we implement the PRECOG objective as follows: we sample 100
ego-agent latents $\mathbf{z}^r$, effectively using random shooting
%\wenyuan{this is the same as random sampling?} 
% \jerry{yep}
rather than gradient descent. In
discussion with the authors they confirmed that results should be similar. Then
for each ego-agent latent, we sample 15 joint actor latent samples
$\mathbf{Z}^h$ as a Monte-Carlo approximation to the expectation. We then
evaluate the goal/prior likelihood costs for each candidate ego-agent latent and
select the ego-agent latent with the smallest cost. Evaluation of the planning
costs for all candidate samples can be efficiently done in a batch manner using
one forward GPU pass. Note that in selecting the optimal ego-agent latent
$\mathbf{z}^r$, there is an intricacy that since PRECOG is a joint
autoregressive model, the ego-agent latent does not correspond to a fixed
ego-agent trajectory, as the final trajectory will depend on the other actor
latents. We avoid this challenge in simulation execution by replanning
sufficiently often (0.3s)
%\wenyuan{why different numbers here?}, 
and also observing that the other actor latents do not generally perturb the ego-agent trajectory too much. 

%\wenyuan{You haven't talk about architecture. Are you using the same arch in
%PRECOG paper?}

\subsection{Discussion of Results} 
As indicated in Table I of the main paper, the PRECOG model underperforms both our non-reactive and reactive objectives based on the energy-based model. We qualitatively analyzed some of the simulation scenarios and offer some hypotheses for the results. The first is that since PRECOG does not explicitly define a collision prior, it's possible that the model does not try to avoid collision in all cases, especially on test scenarios that are out-of-distribution from the training data (especially in CARLA, where the traffic density in simulation is higher than in training). The second is that sampling in latent space does not guarantee a diverse range of trajectories for the ego-agent. In fact, we notice that in some turning scenarios where we set the goal state to be the result of a turn, the ego-agent still goes in a straight line, even when the prior likelihood weight goes to 0 and the number of ego-agent samples is high (tried setting to 1000). We hypothesize that this is partially due to test distribution shift. Nevertheless, we find learned autoregressive models promising to keep in mind for the future. 

We showcase a qualitative example comparing PRECOG with our model in the next section, in Fig. \ref{fig:supp_qual_simba2}. 

\section{Additional Qualitative Results}

We present a few additional qualitative results to better visually highlight the difference between our reactive and non-reactive model in various interactive scenarios. For each demonstration, we provide snapshots of the simulation at different timesteps. As with the results in the main paper, the ego-agent and planned trajectory are in \textcolor{green}{green}, the actors are represented by an assortment of other colors, and the goal position or lane is in \textcolor{cyan}{cyan}. Results are best viewed by zooming in with a digital reader. 

\begin{figure*}[t!]
	\centering
	\def\imw{0.24\textwidth}
	\setlength{\tabcolsep}{0.2pt}
	\begin{tabular}{cccc}
	\figcap{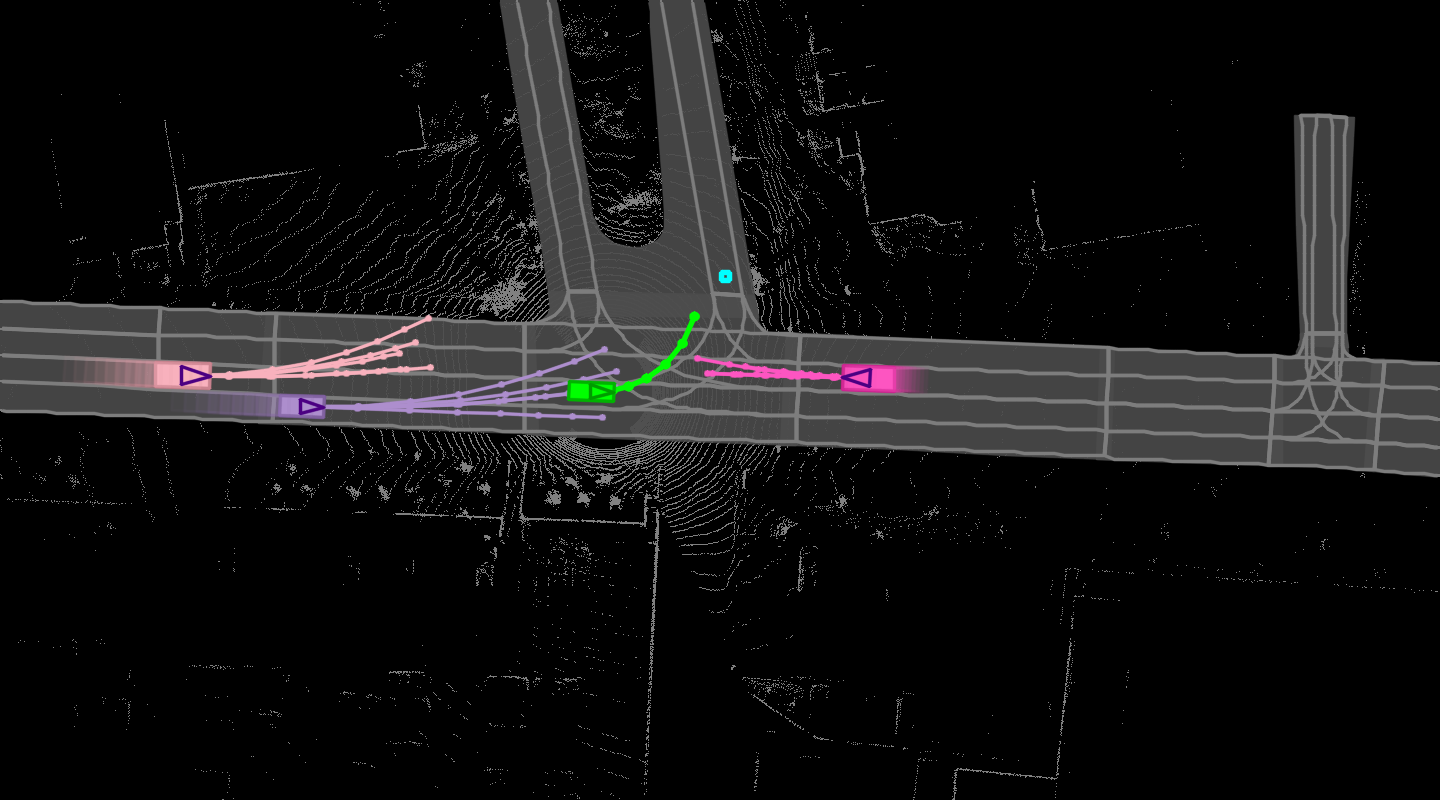}{400 250 400 250}{Reactive, t=0s} &
	\figcap{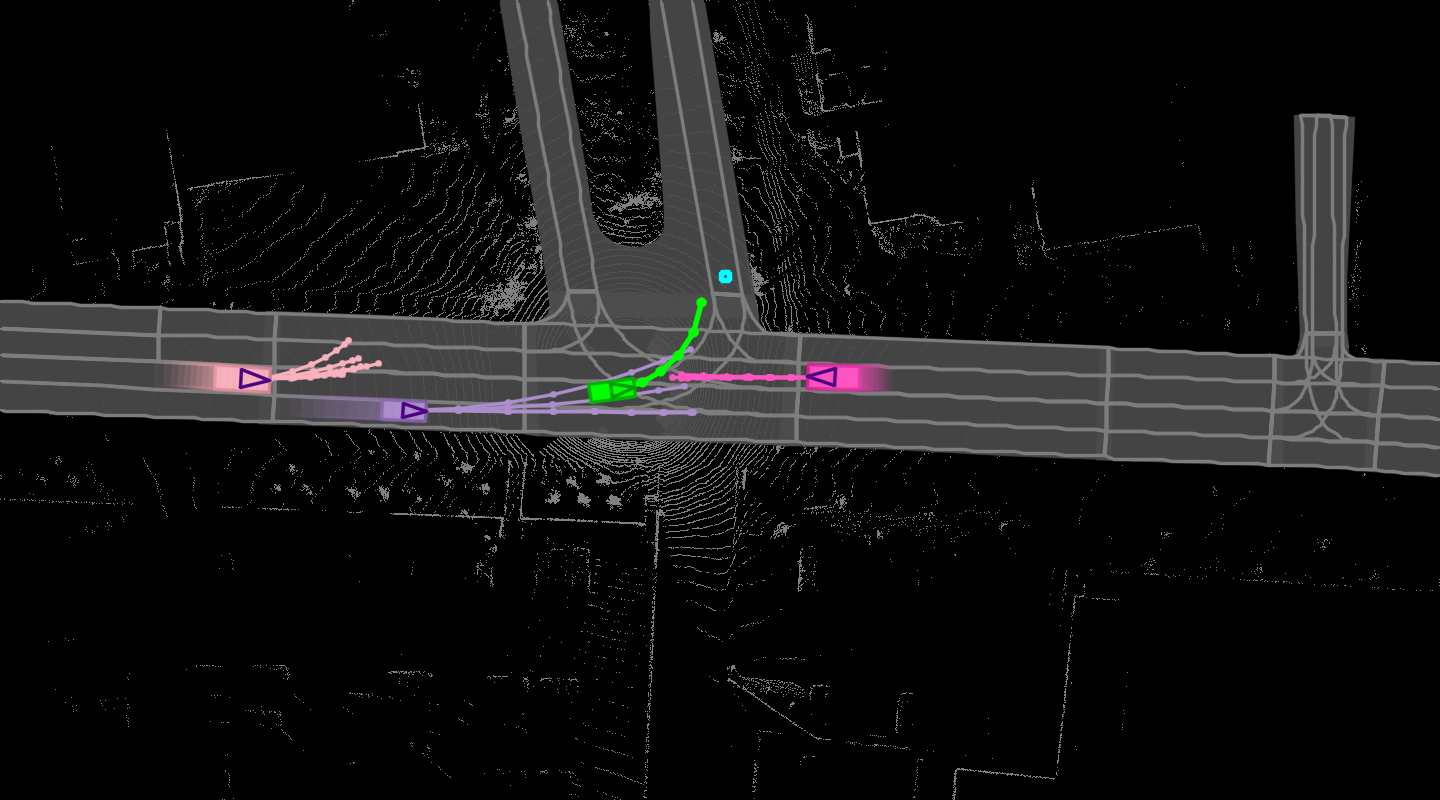}{400 250 400 250}{Reactive, t=1s} &
	\figcap{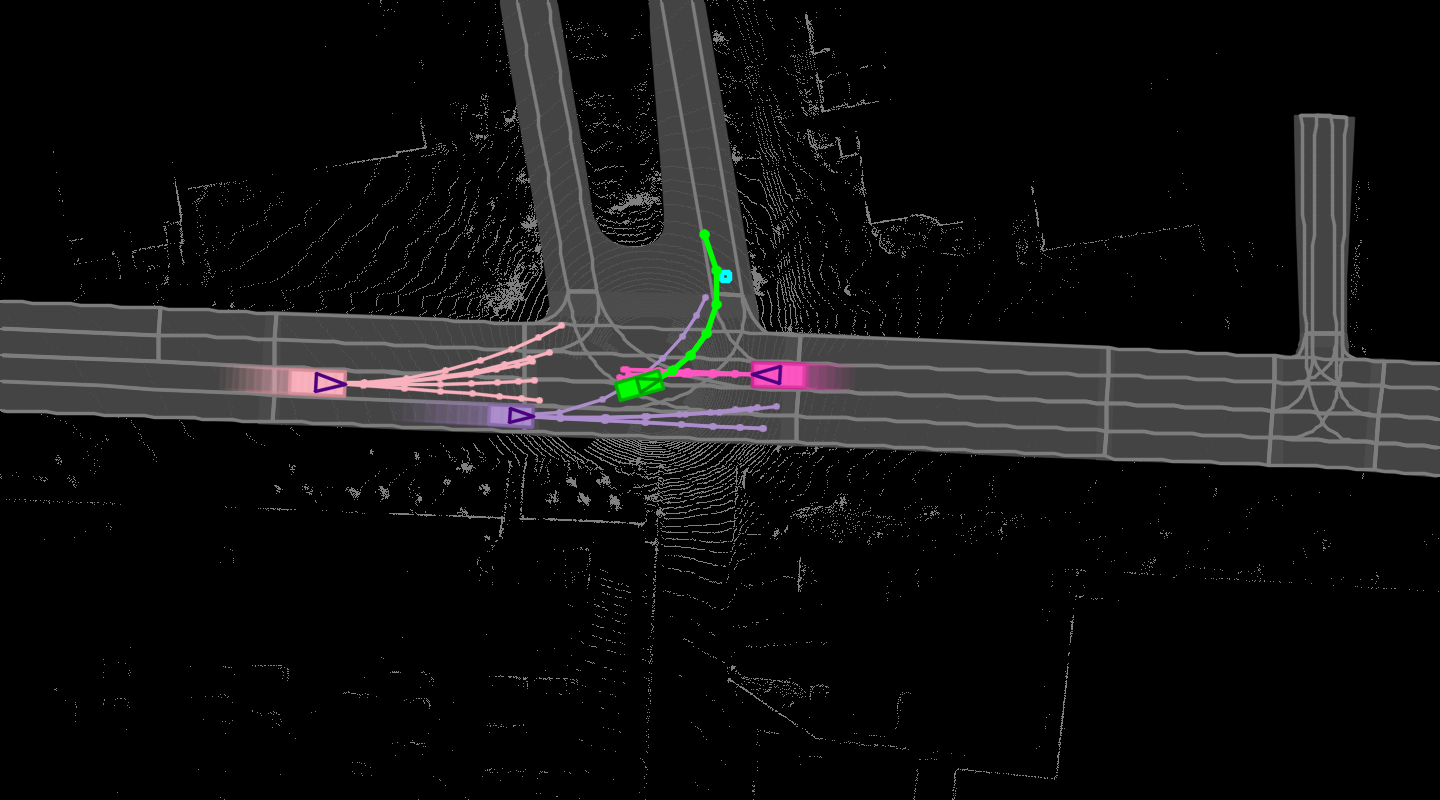}{400 250 400 250}{Reactive, t=2s} &
	\figcap{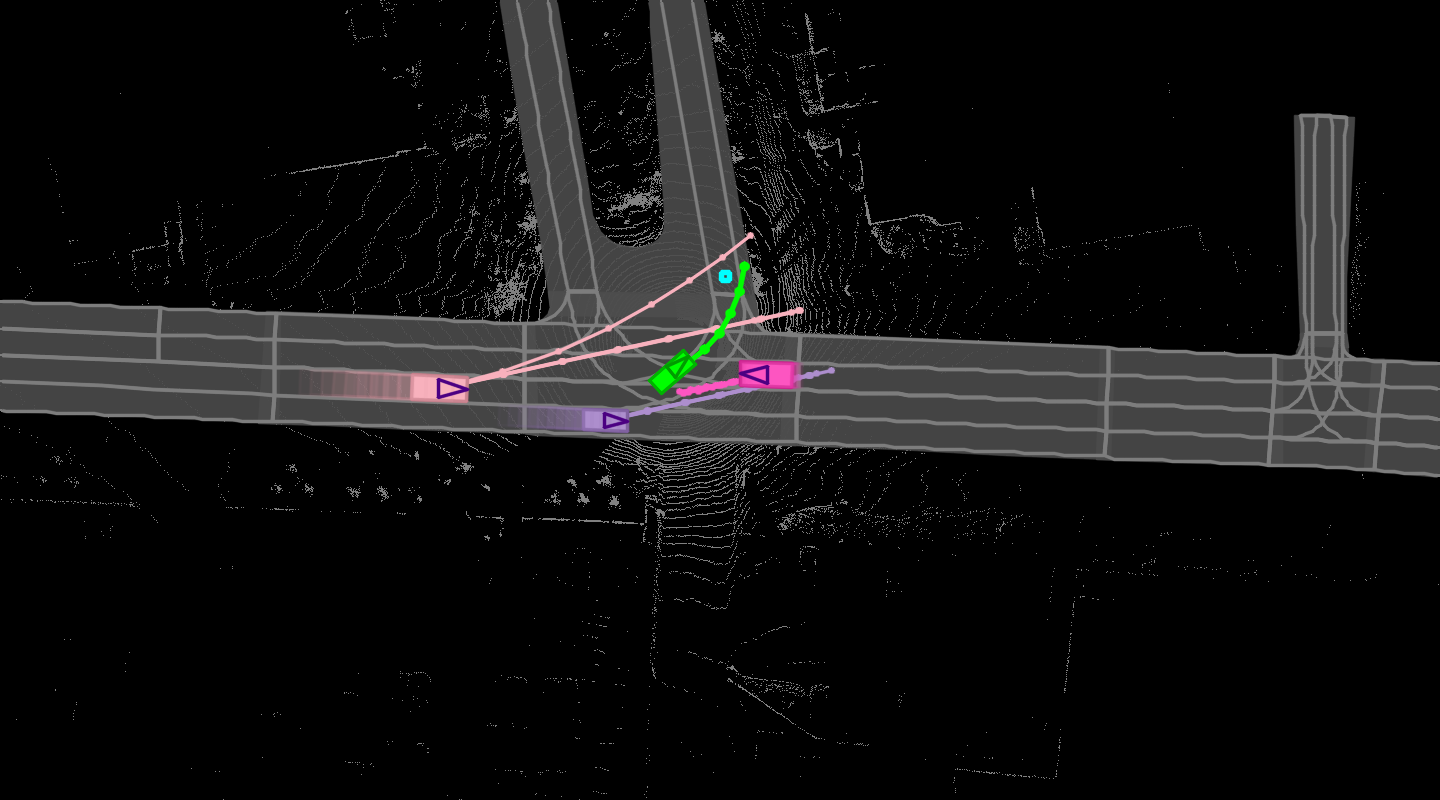}{400 250 400 250}{Reactive, t=3s} \\
	\figcap{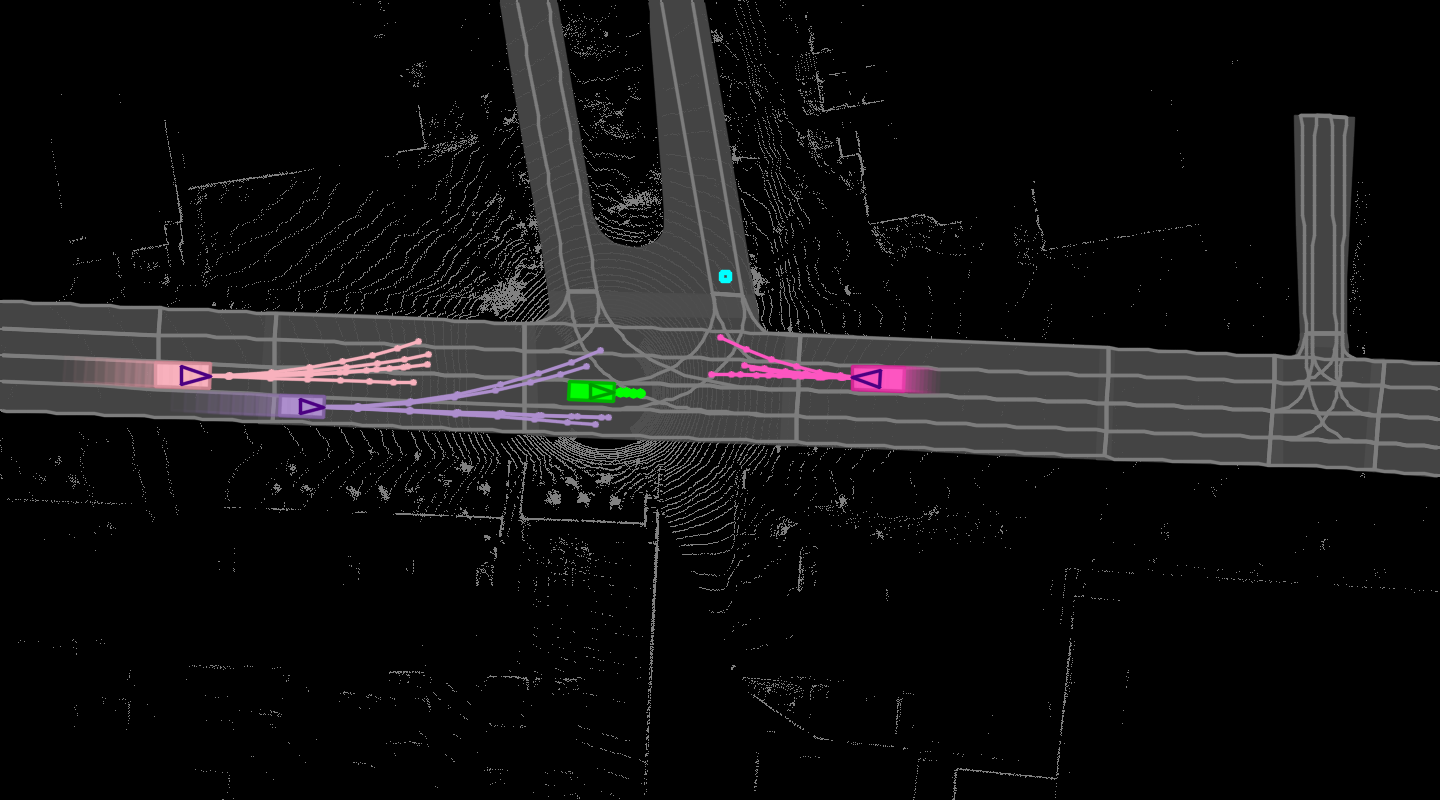}{400 250 400 250}{Non-Reactive, t=0s} &
	\figcap{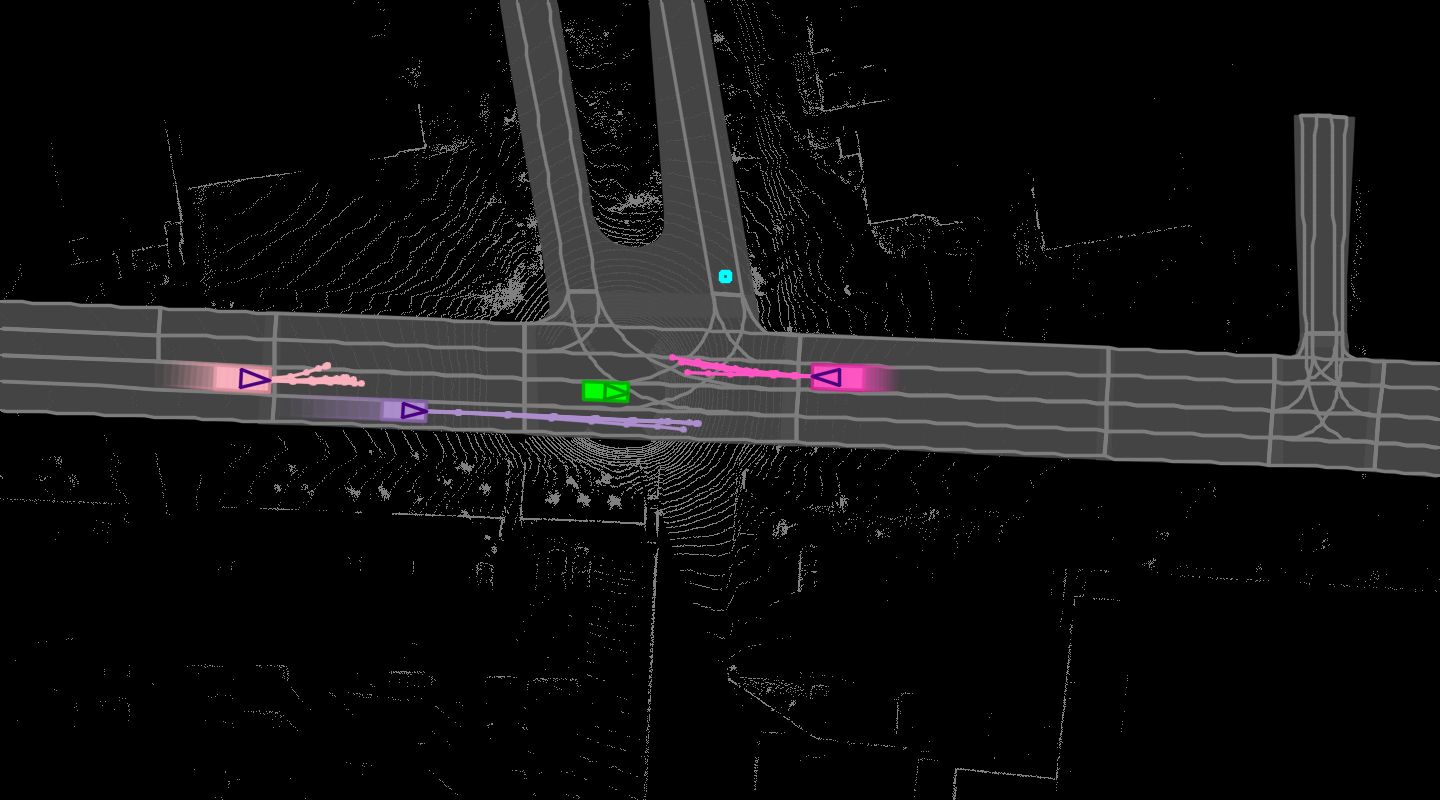}{400 250 400 250}{Non-Reactive, t=1s} &
	\figcap{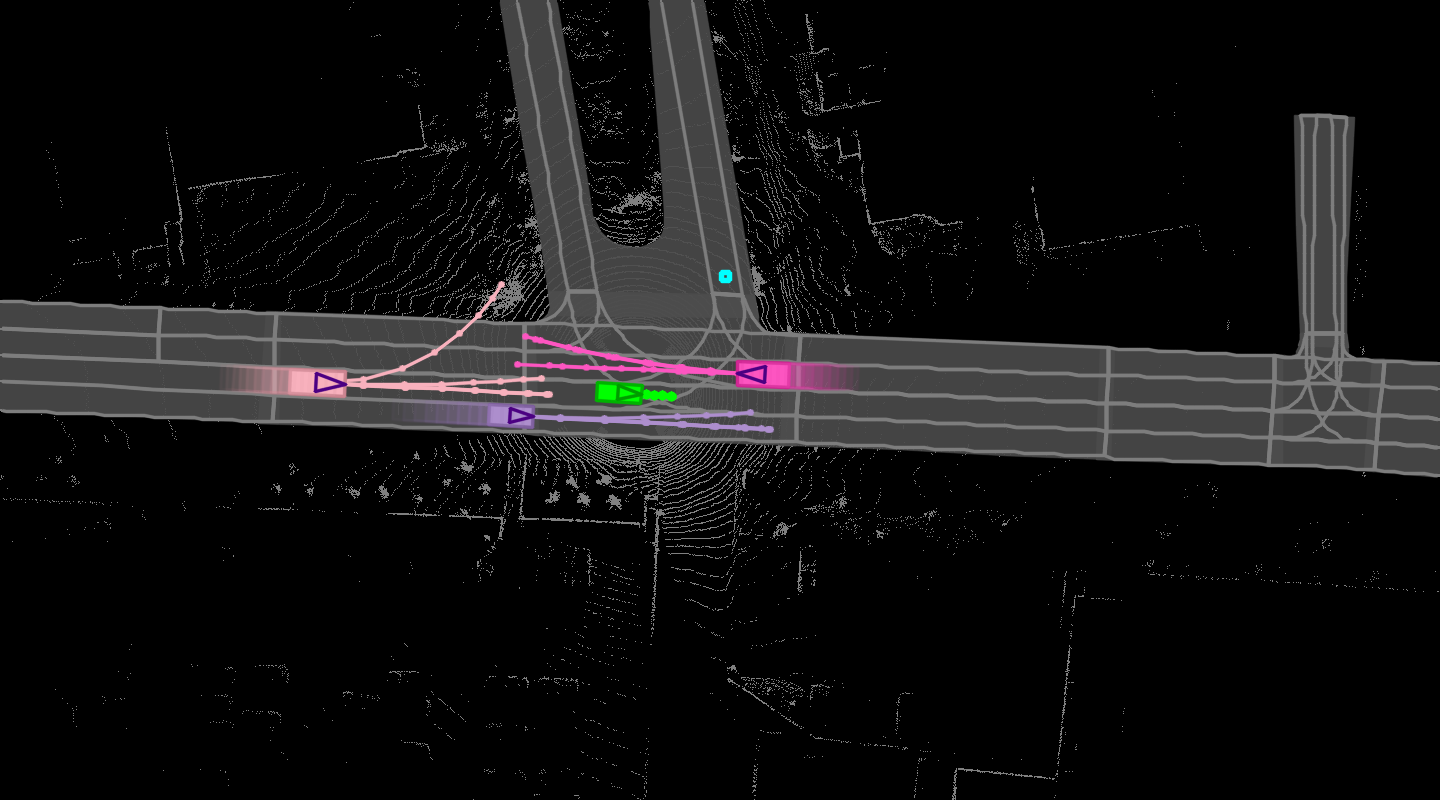}{400 250 400 250}{Non-Reactive, t=2s} &
	\figcap{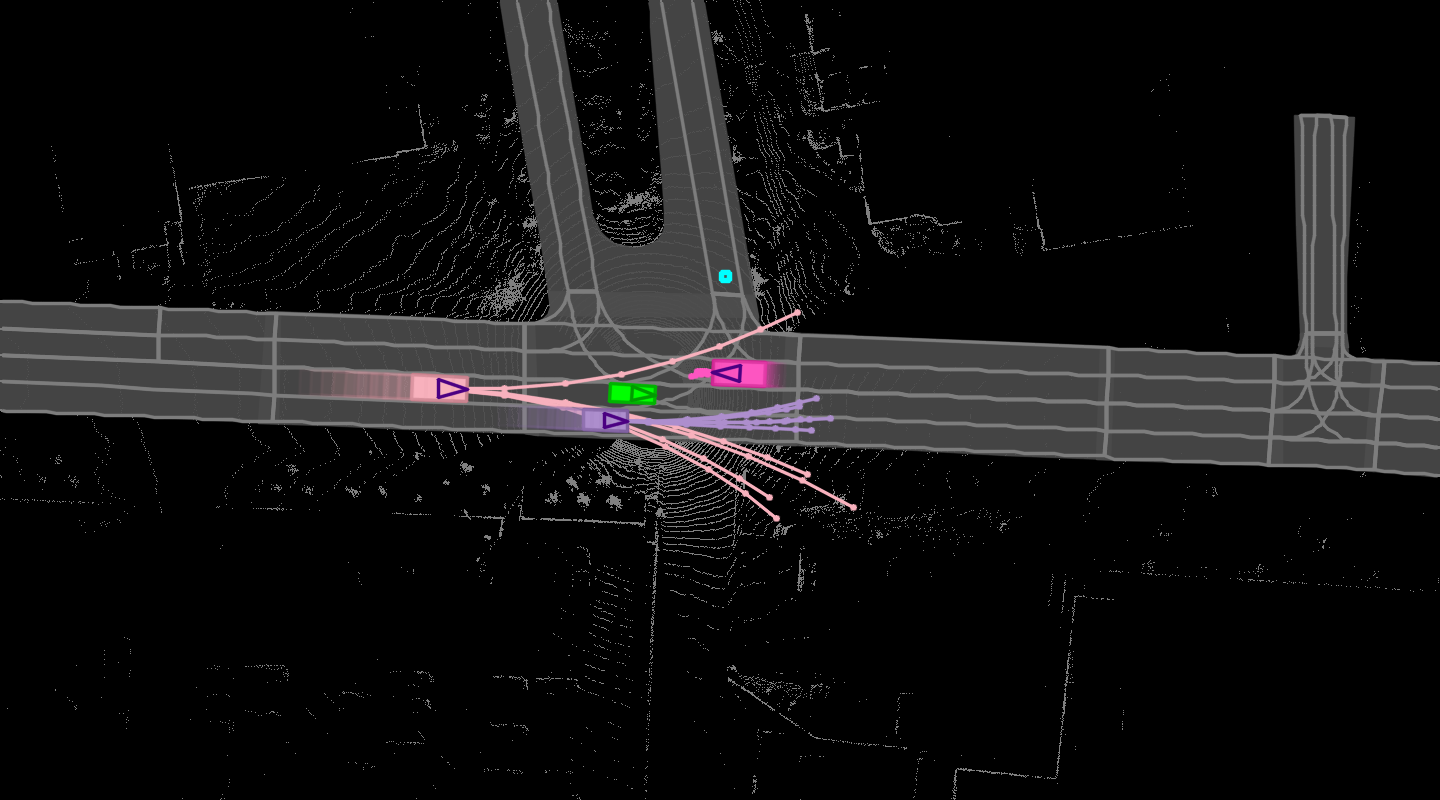}{400 250 400 250}{Non-Reactive, t=3s} \\
	\end{tabular}
	\caption{Visualization of a Simba turn for non-reactive (bottom) and reactive (top) models at 3 different time steps: 0s, 1s, 2s, 3s (left-to-right). 
		%	\caption{Visualization of a CARLA lane merge for non-reactive (bottom) and reactive (top) models at 3 different time steps: 1s (left), 2s (middle), 3s (right).  AV is green, planned trajectory orange. Other actors in blue/purple, with estimated heading in red. Goal lane in cyan.  
		%\wenyuan{I think it might be
		%better to overlay the lane boundary on the lidar viz, it's a bit hard to
		%understand where is the road and what would be a reasonable behavior} 
	}
	\label{fig:supp_qual_simba1}
\end{figure*}

In Fig. \ref{fig:supp_qual_simba1} we demonstrate the ego-agent performing an unprotected left turn in front of an oncoming vehicle in Simba. We first emphasize that since this scenario was initialized from a real driving sequence, the actual ``expert'' trajectory also performed the unprotected left turn against the same oncoming vehicle, implying that such an action is not an unsafe maneuver depending on the oncoming vehicle's position/velocity. The visualizations of our models show that the reactive model is successfully able to perform the left turn, while the non-reactive model surprisingly gets stuck in place, even as the oncoming vehicle slows down. We speculate that this may be due to the model choosing to stay still rather than violate the safety distance of the other actor. 

\begin{figure*}
	\centering
	\def\imw{0.24\textwidth}
	\setlength{\tabcolsep}{0.2pt}
	\begin{tabular}{cccc}
	\figcap{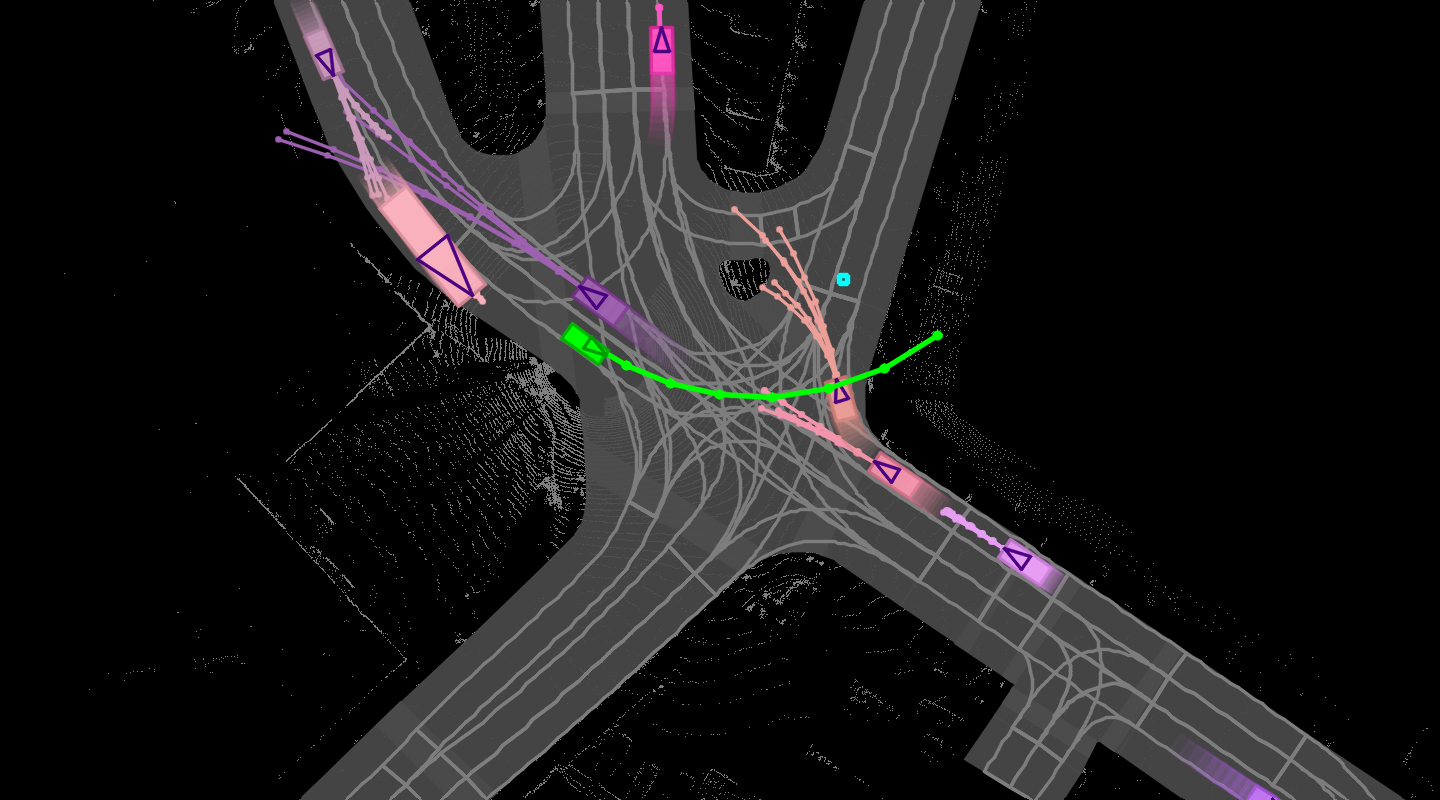}{400 250 400 250}{Reactive, t=0s} &
	\figcap{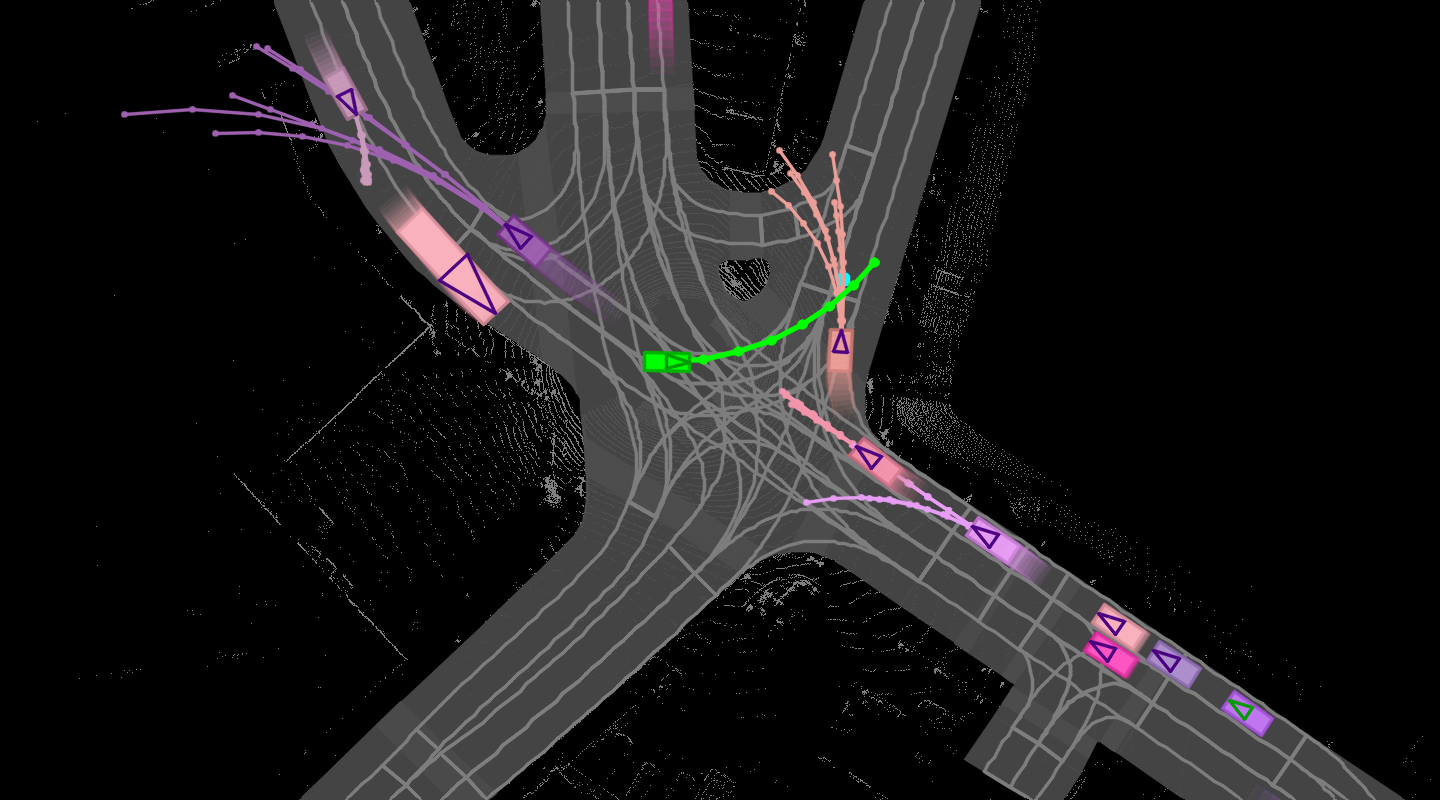}{400 250 400 250}{Reactive, t=1s} &
	\figcap{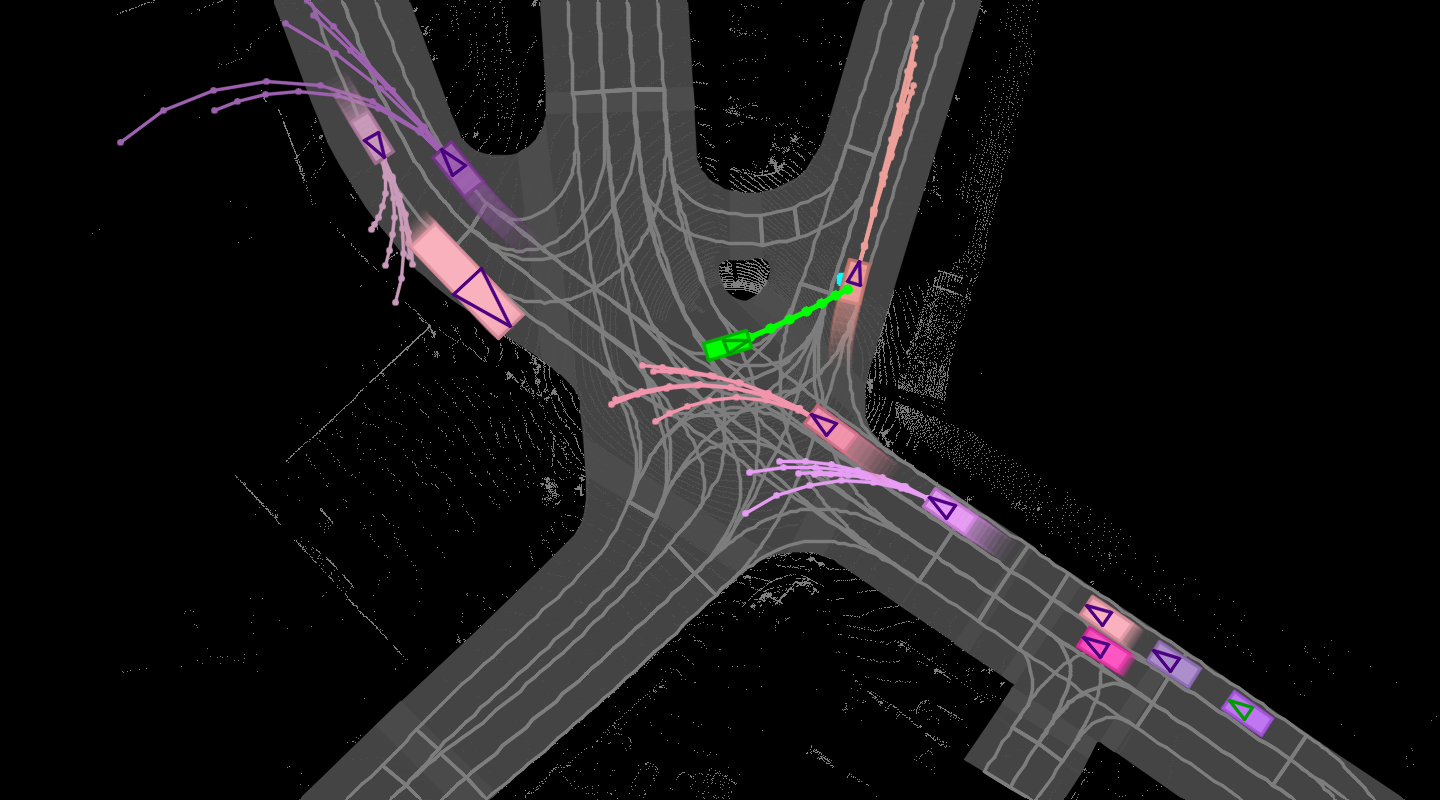}{400 250 400 250}{Reactive, t=2s} &
	\figcap{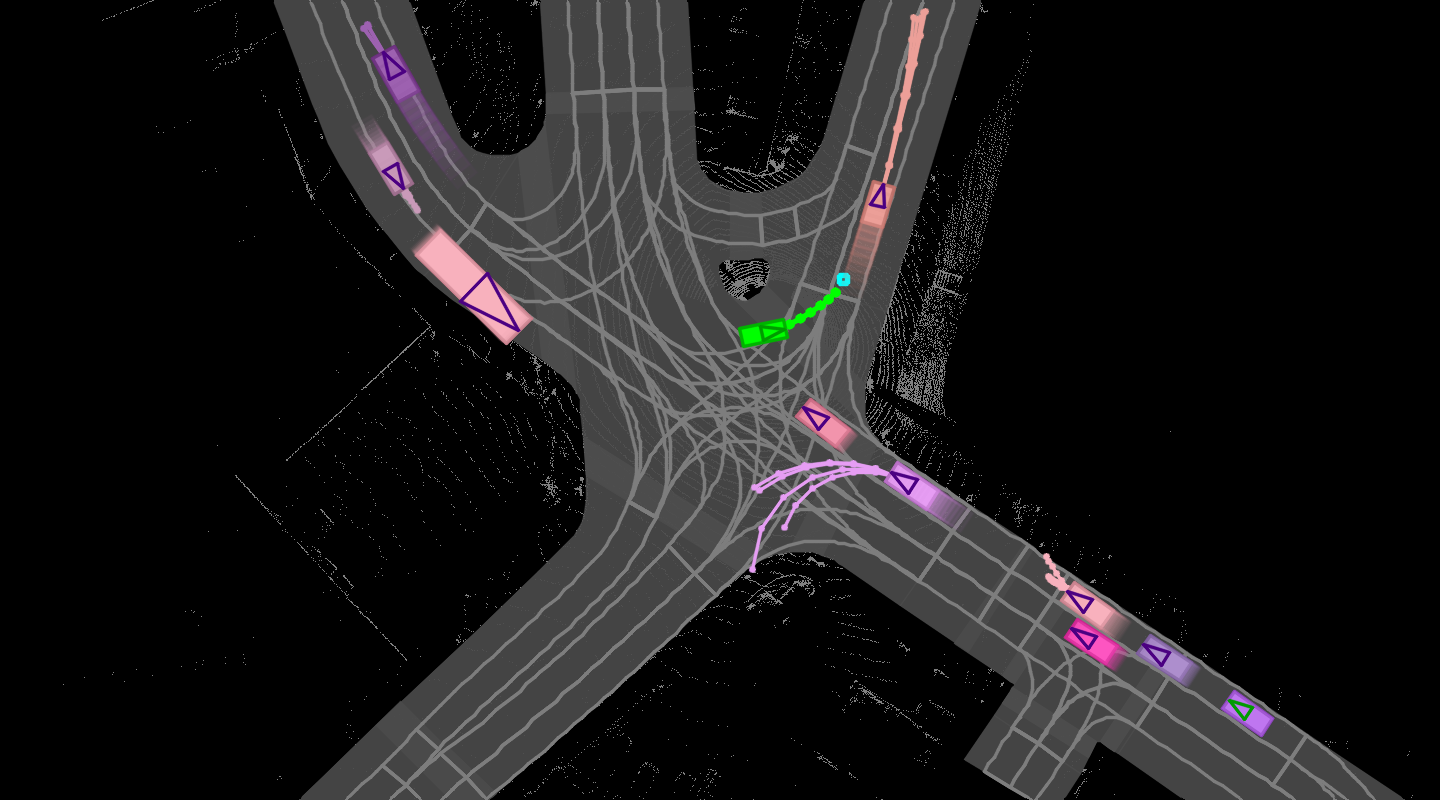}{400 250 400 250}{Reactive, t=3s} \\
	\figcap{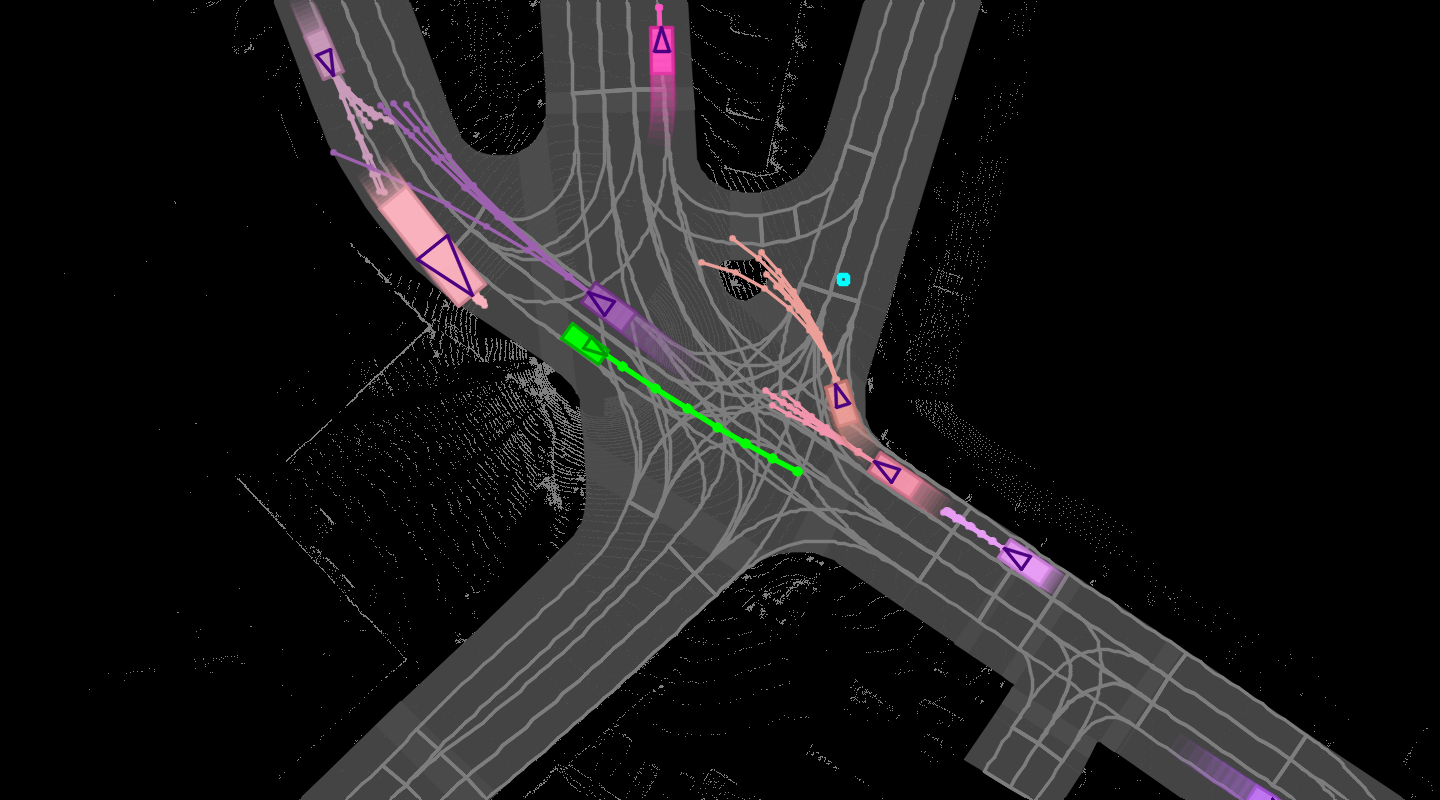}{400 250 400 250}{Non-Reactive, t=0s} &
	\figcap{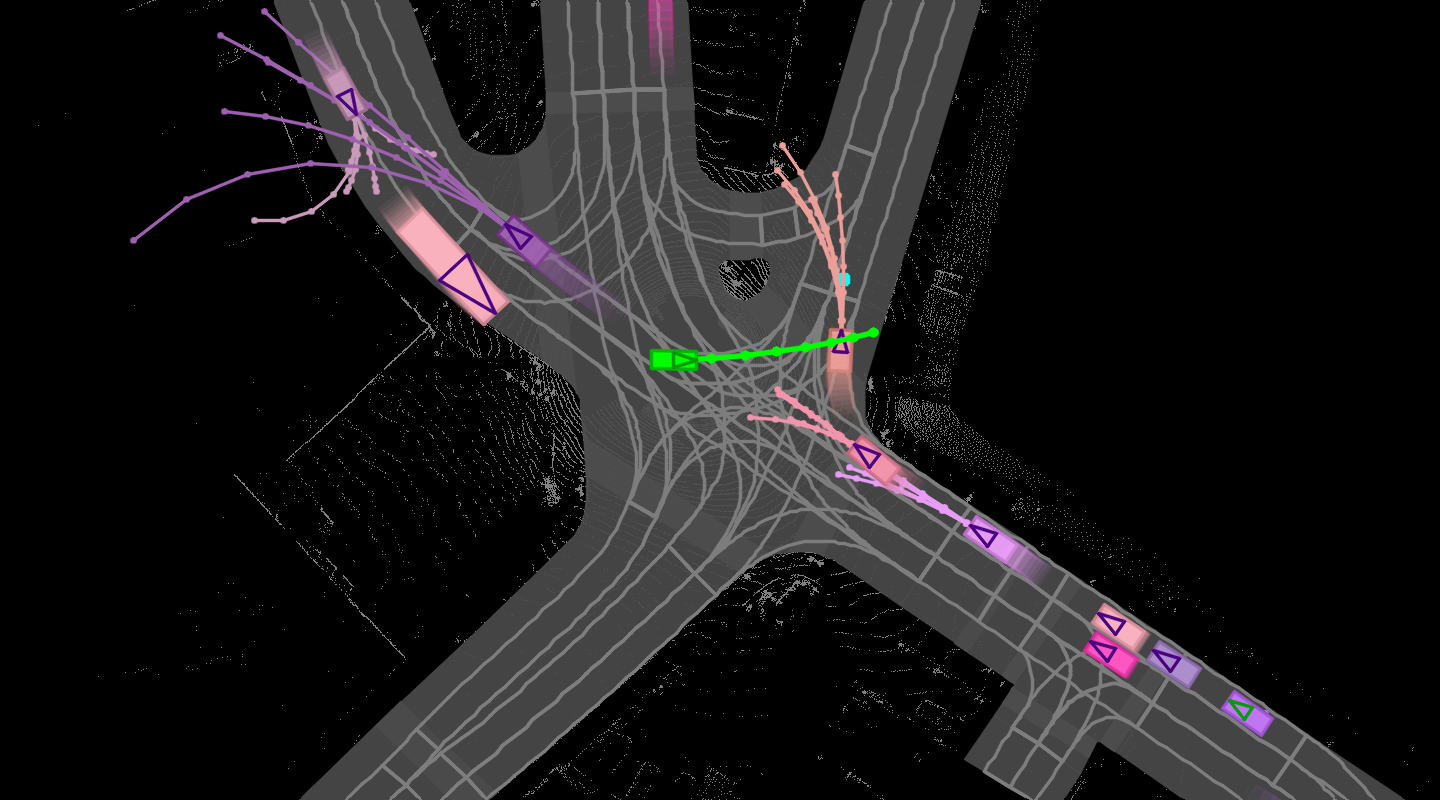}{400 250 400 250}{Non-Reactive, t=1s} &
	\figcap{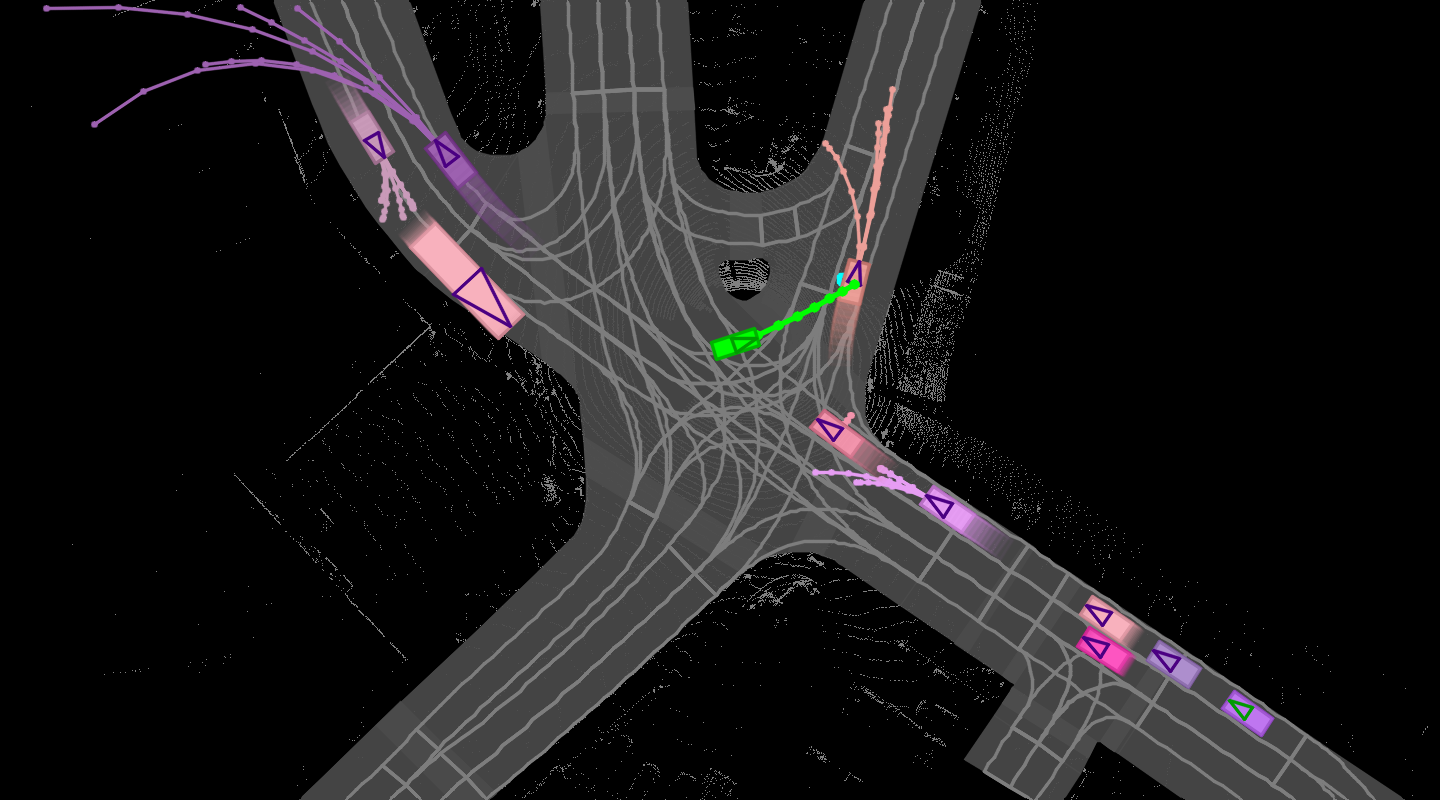}{400 250 400 250}{Non-Reactive, t=2s} &
	\figcap{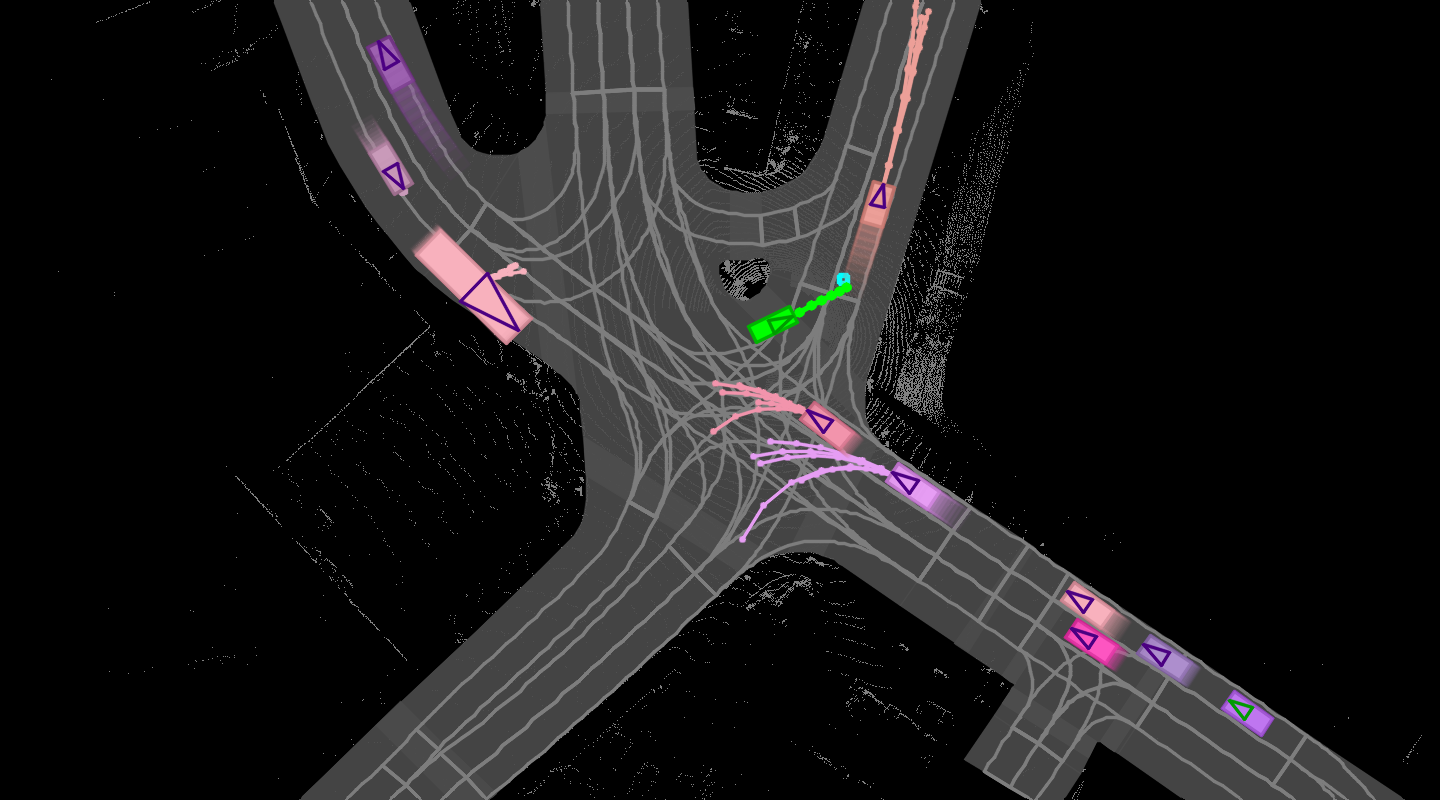}{400 250 400 250}{Non-Reactive, t=3s} \\ 
	\figcap{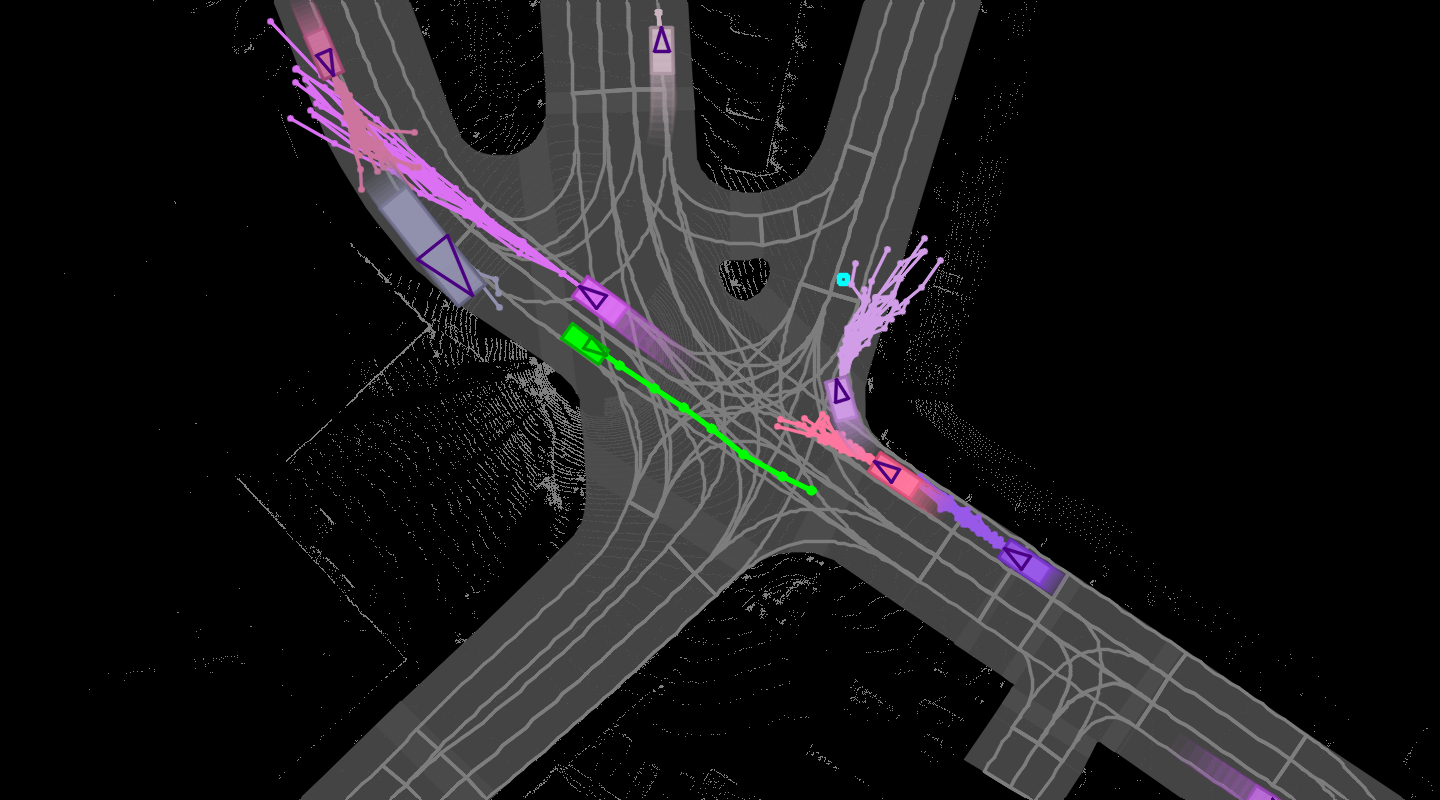}{400 250 400 250}{PRECOG, t=0s} &
	\figcap{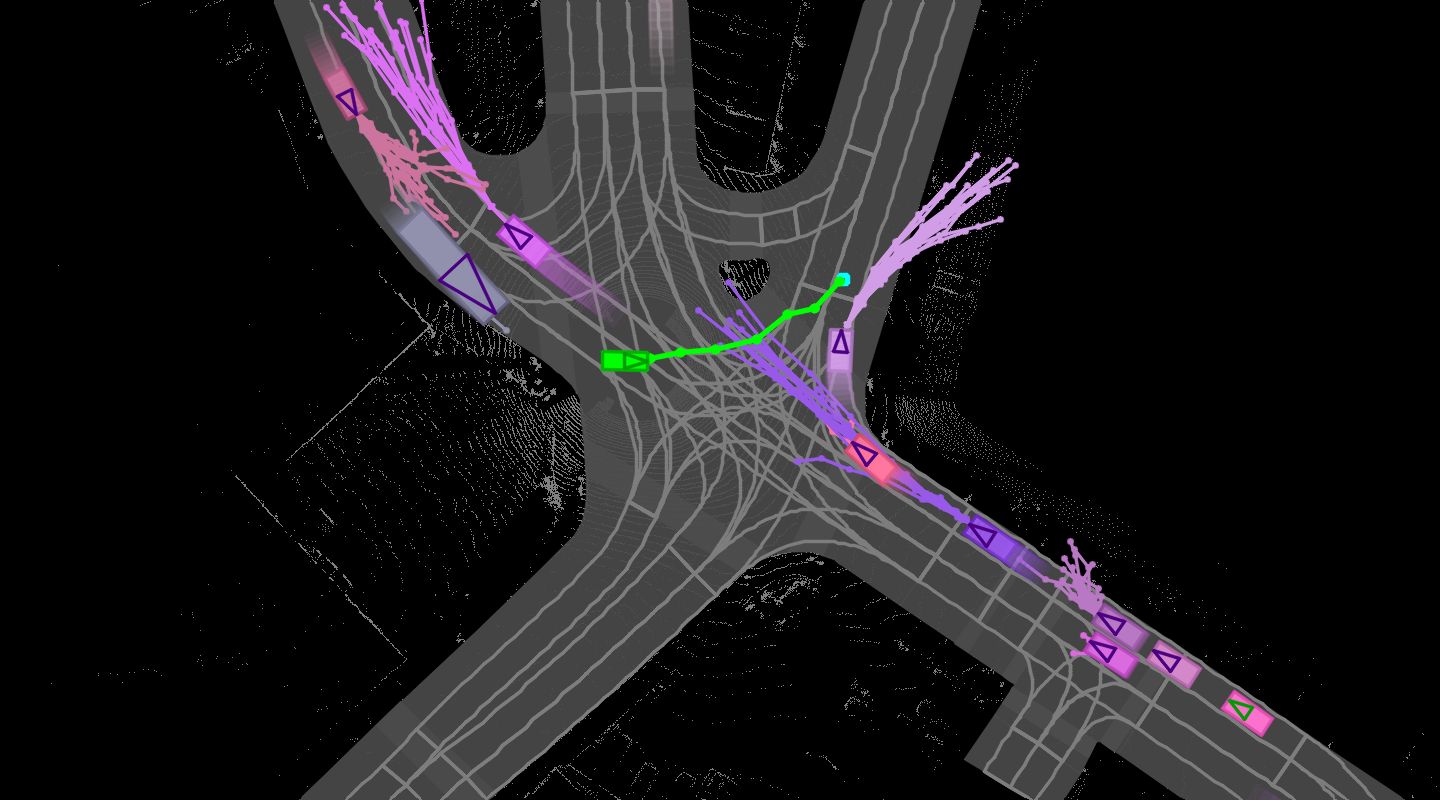}{400 250 400 250}{PRECOG, t=1s} &
	\figcap{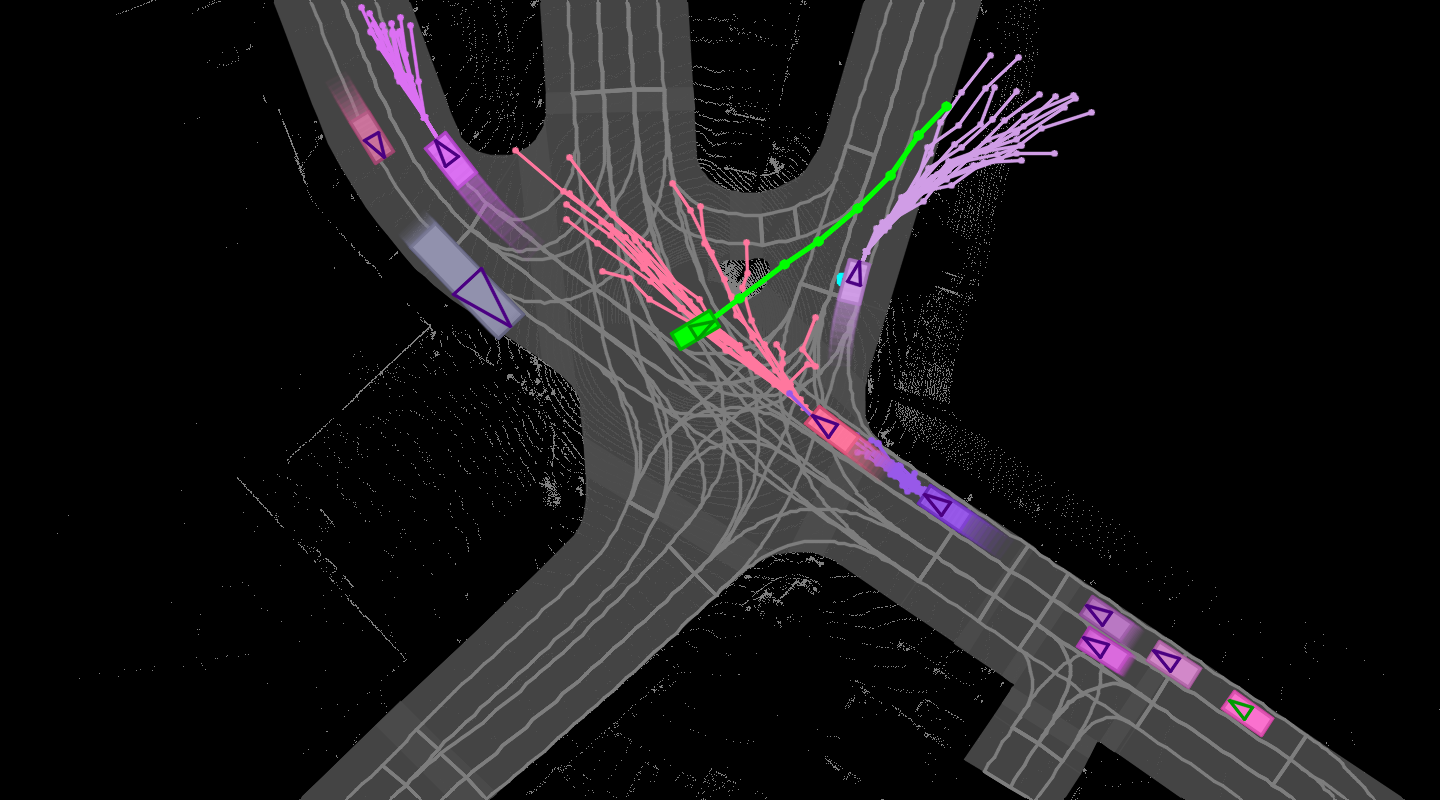}{400 250 400 250}{PRECOG, t=2s} &
	\figcap{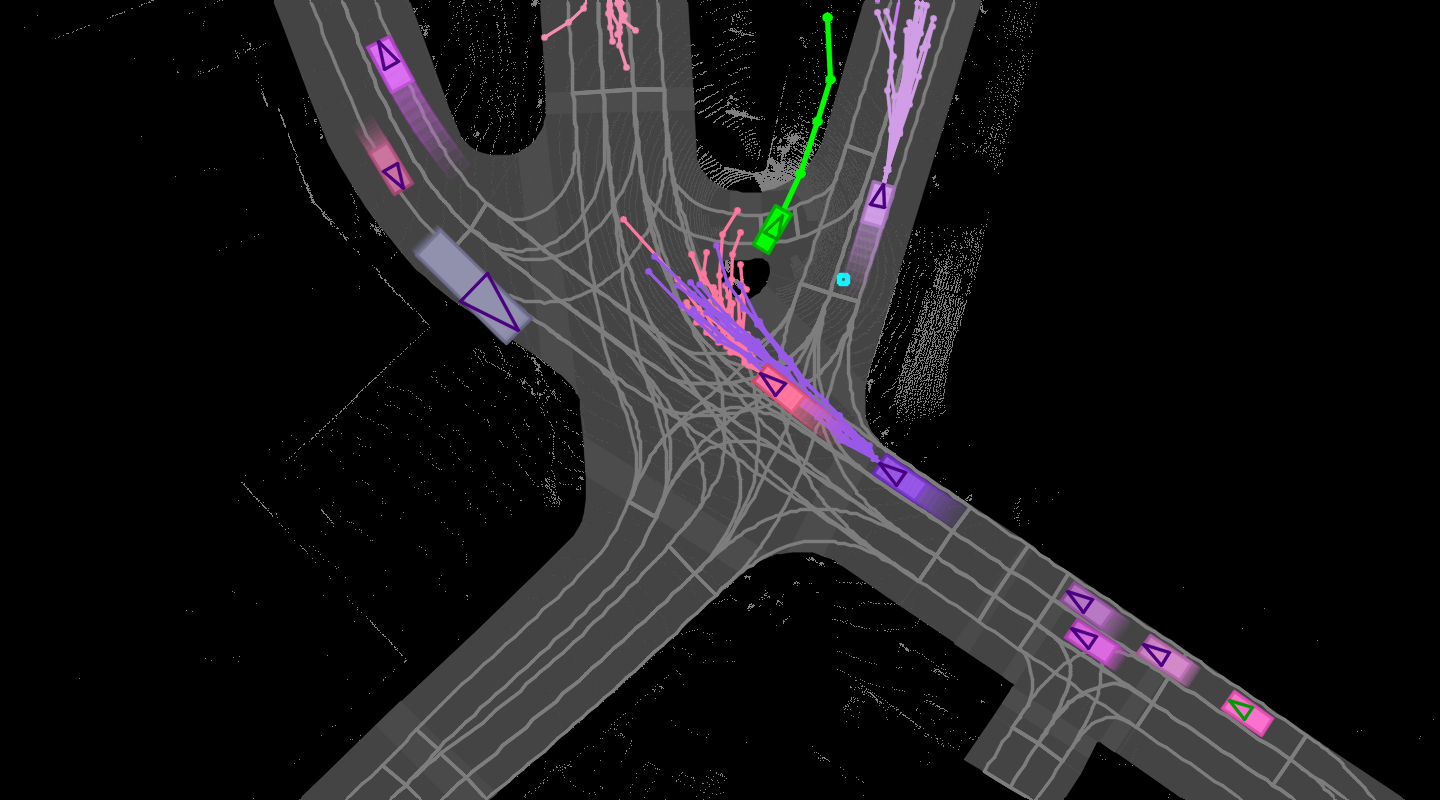}{400 250 400 250}{PRECOG, t=3s} \\
\end{tabular}
	\caption{Visualization of a Simba turn scenario for reactive (top), non-reactive (middle), and PRECOG (bottom) models at 3 different time steps: 0s, 1s, 2s, 3s (left-to-right).
	}
	\label{fig:supp_qual_simba2}
\end{figure*}

Fig. \ref{fig:supp_qual_simba2} showcases a comparison between our reactive/non-reactive models and PRECOG in performing a left turn at a busy intersection. We note that both the reactive/non-reactive models are able to reach the goal state, though admittedly they violate lane boundaries in doing so (lane following is not explicitly encoded as an energy in our model). Interestingly, the PRECOG model plans a trajectory to the goal at $t=1$ but is not able to complete it at later timestamps, either implying that the latent samples for the ego-agent do not capture such a behavior or that the prior likelihood cost is too high to go any further. It's possible that the model can be tuned further, both in terms of the data, the training scheme, as well as the PRECOG evaluation procedure, so we mostly present these as initial results for future investigation. 

\begin{figure*}
	\centering
	\def\imw{0.24\textwidth}
	\setlength{\tabcolsep}{0.2pt}
	\begin{tabular}{cccc}
	\figcap{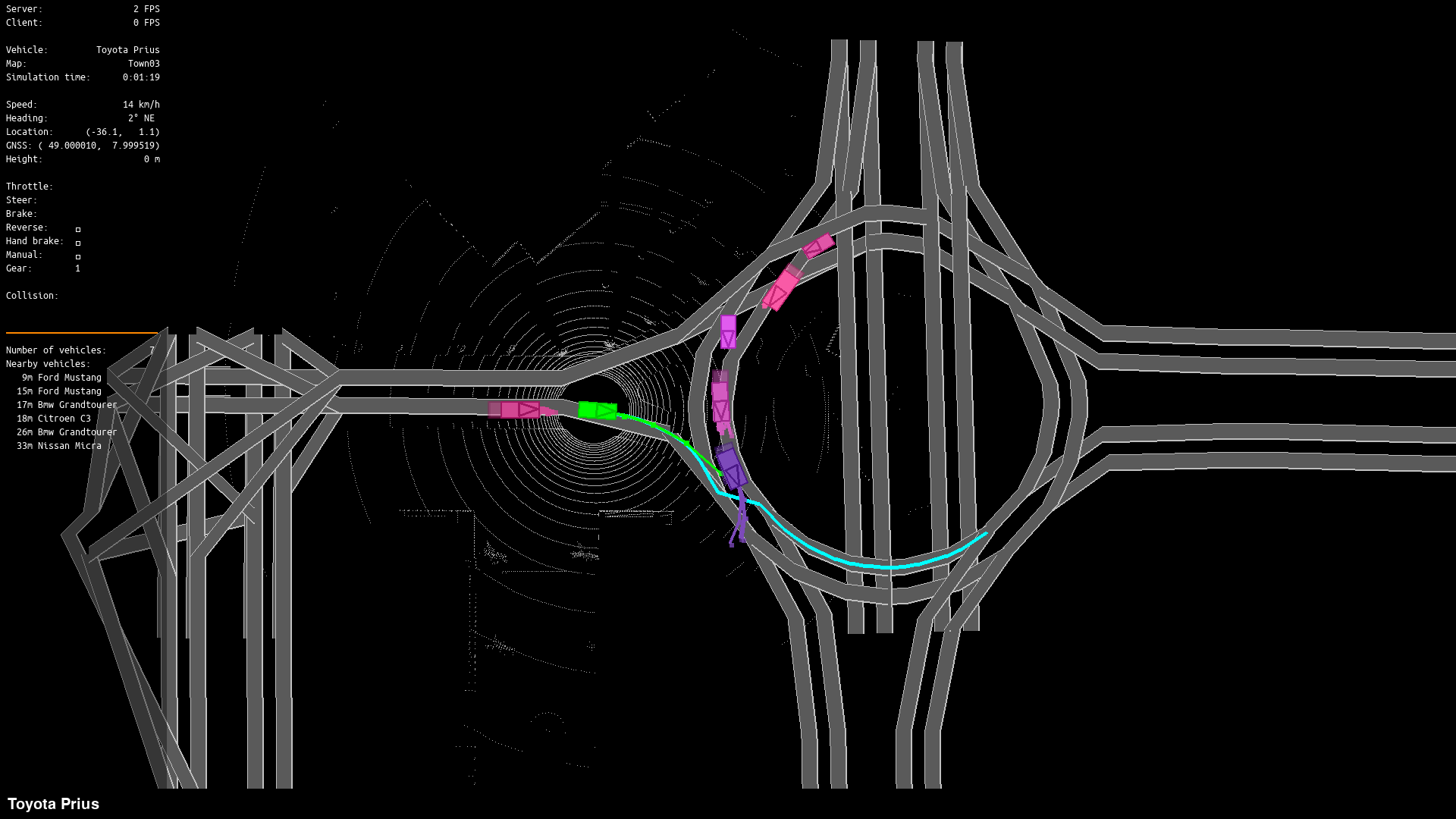}{650 340 750 460}{Reactive, t=0s} &
	\figcap{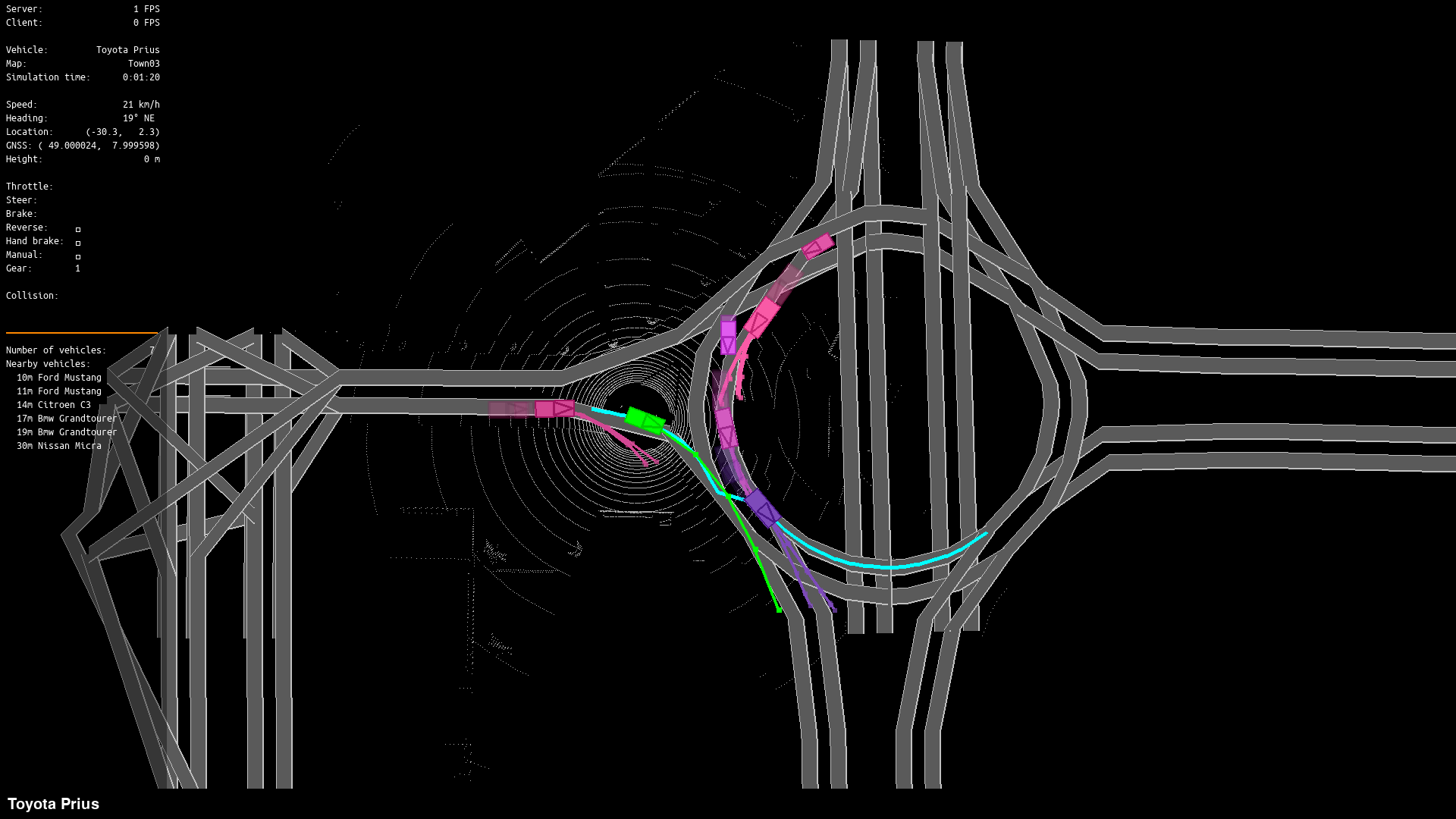}{650 340 750 460}{Reactive, t=1s} &
	\figcap{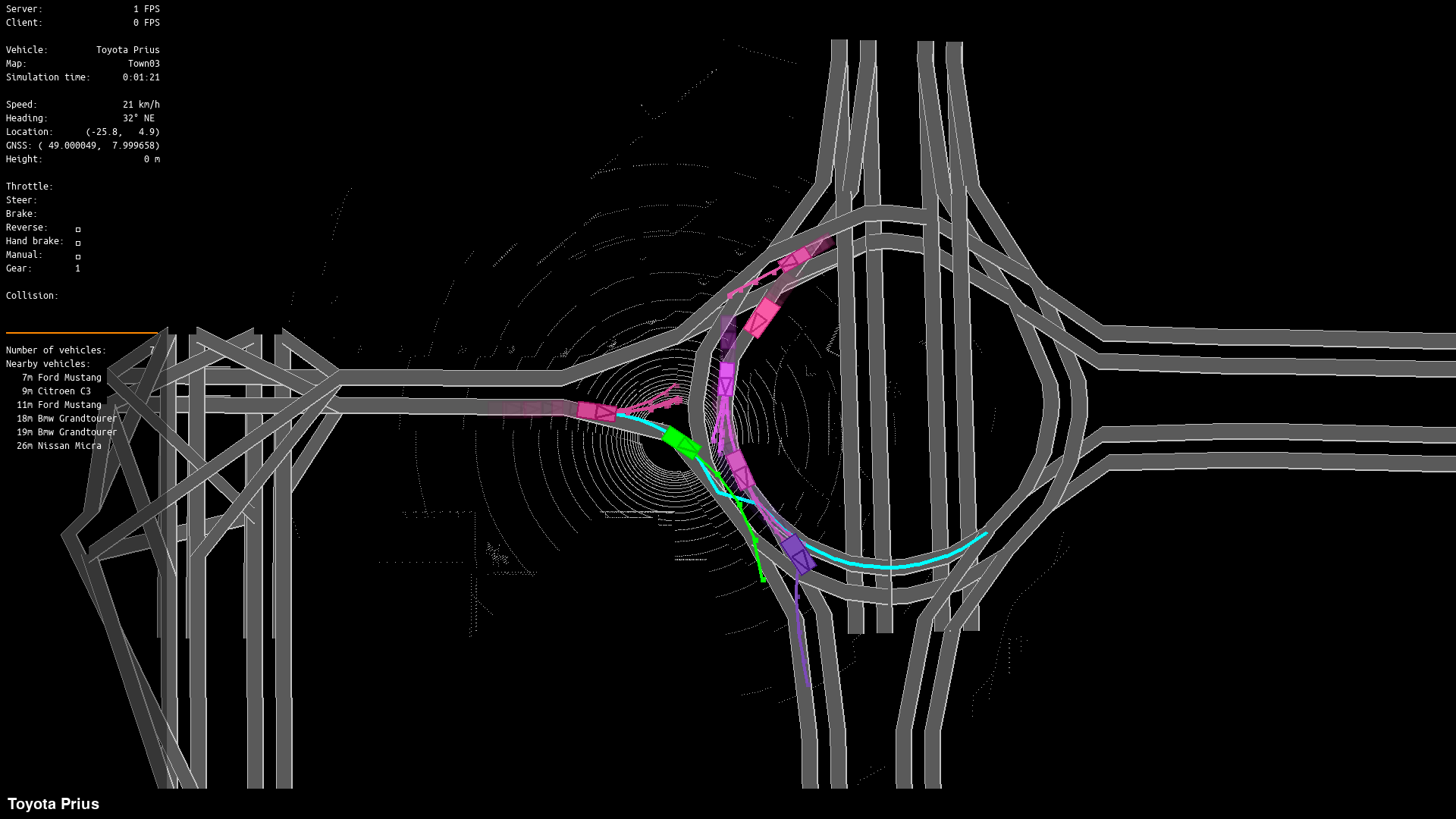}{650 340 750 460}{Reactive, t=2s} &
	\figcap{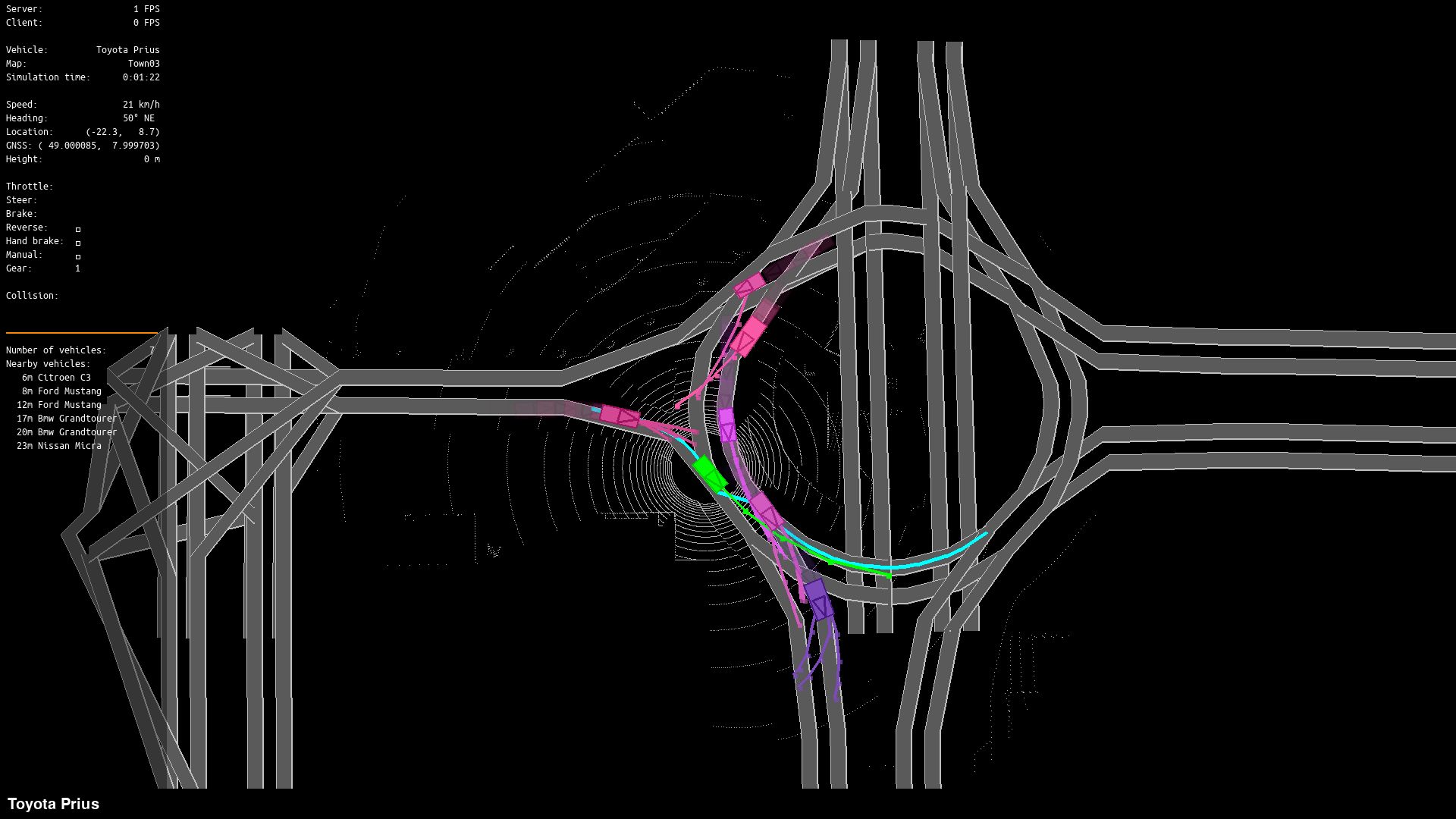}{650 340 750 460}{Reactive, t=3s} \\
	\figcap{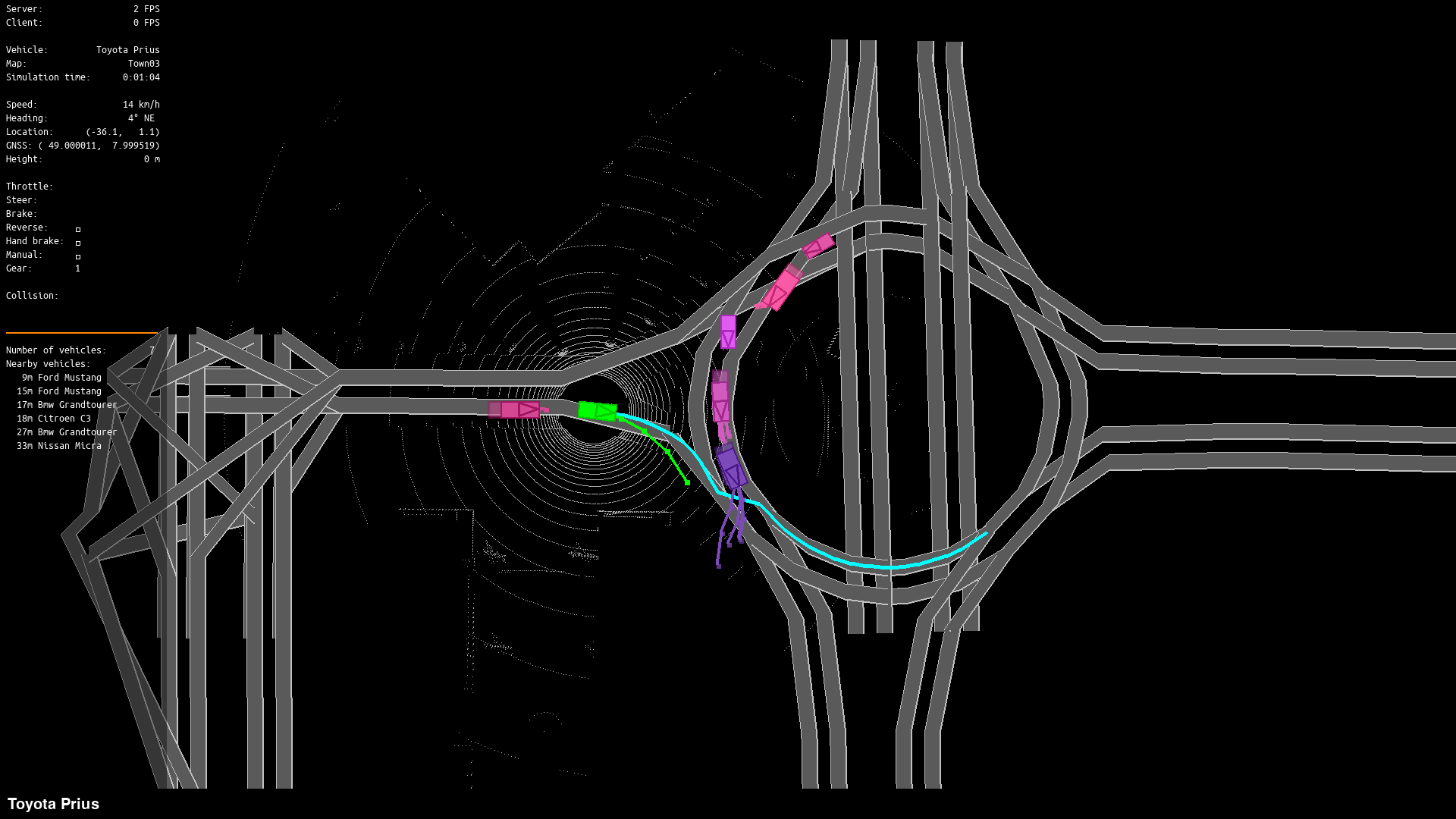}{650 340 750 460}{Non-Reactive, t=0s} &
	\figcap{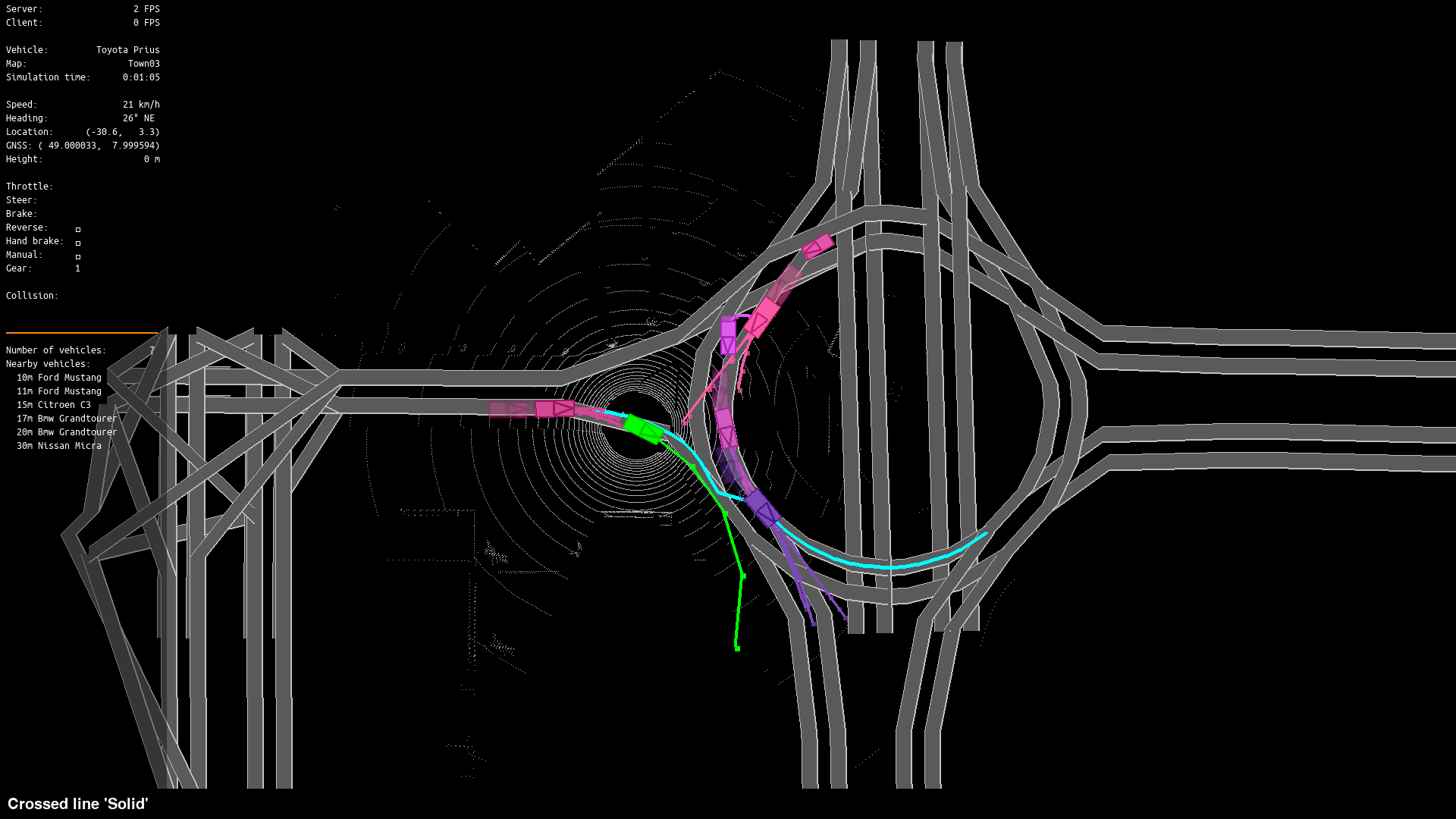}{650 340 750 460}{Non-Reactive, t=1s} &
	\figcap{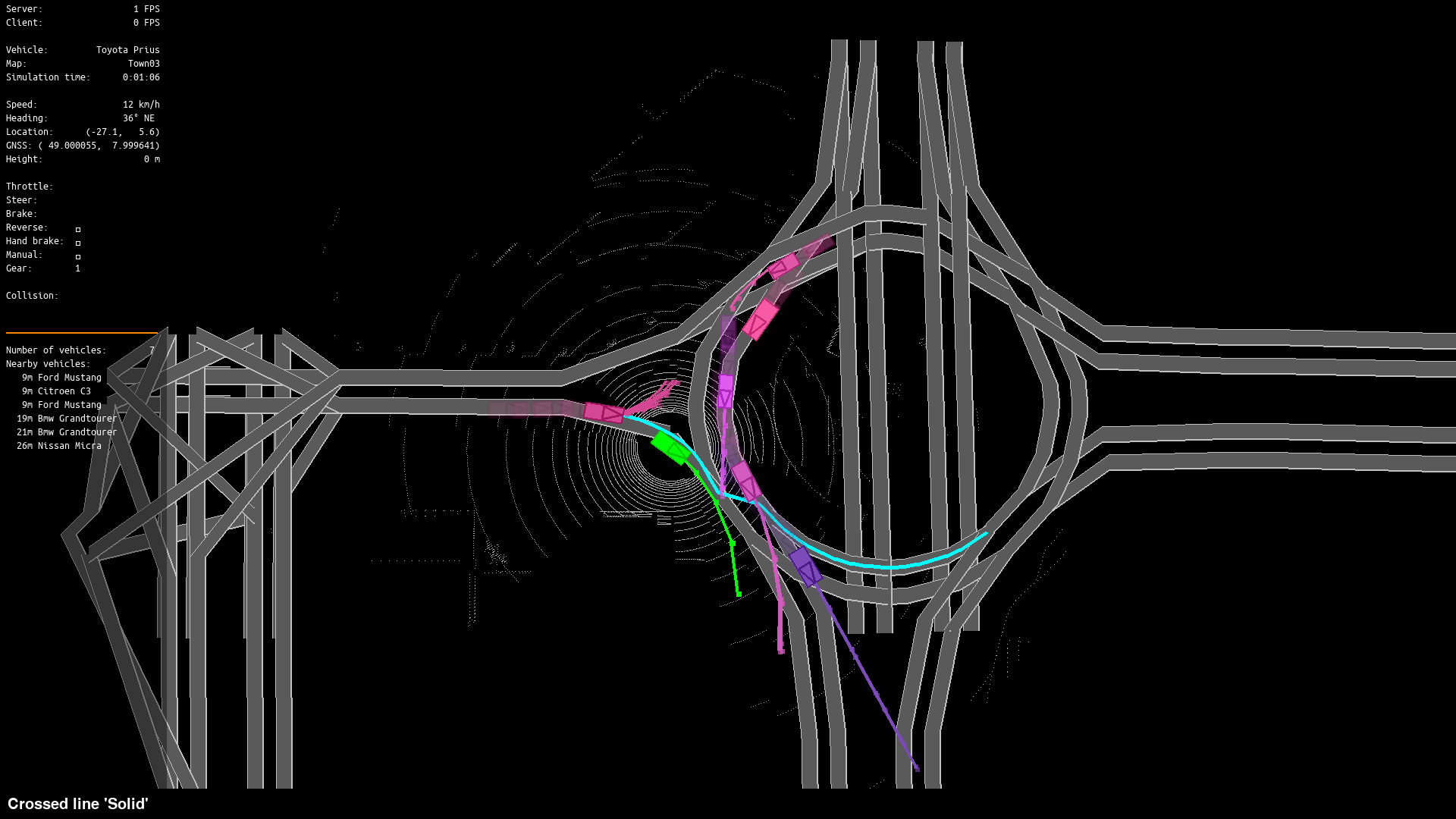}{650 340 750 460}{Non-Reactive, t=2s} &
	\figcap{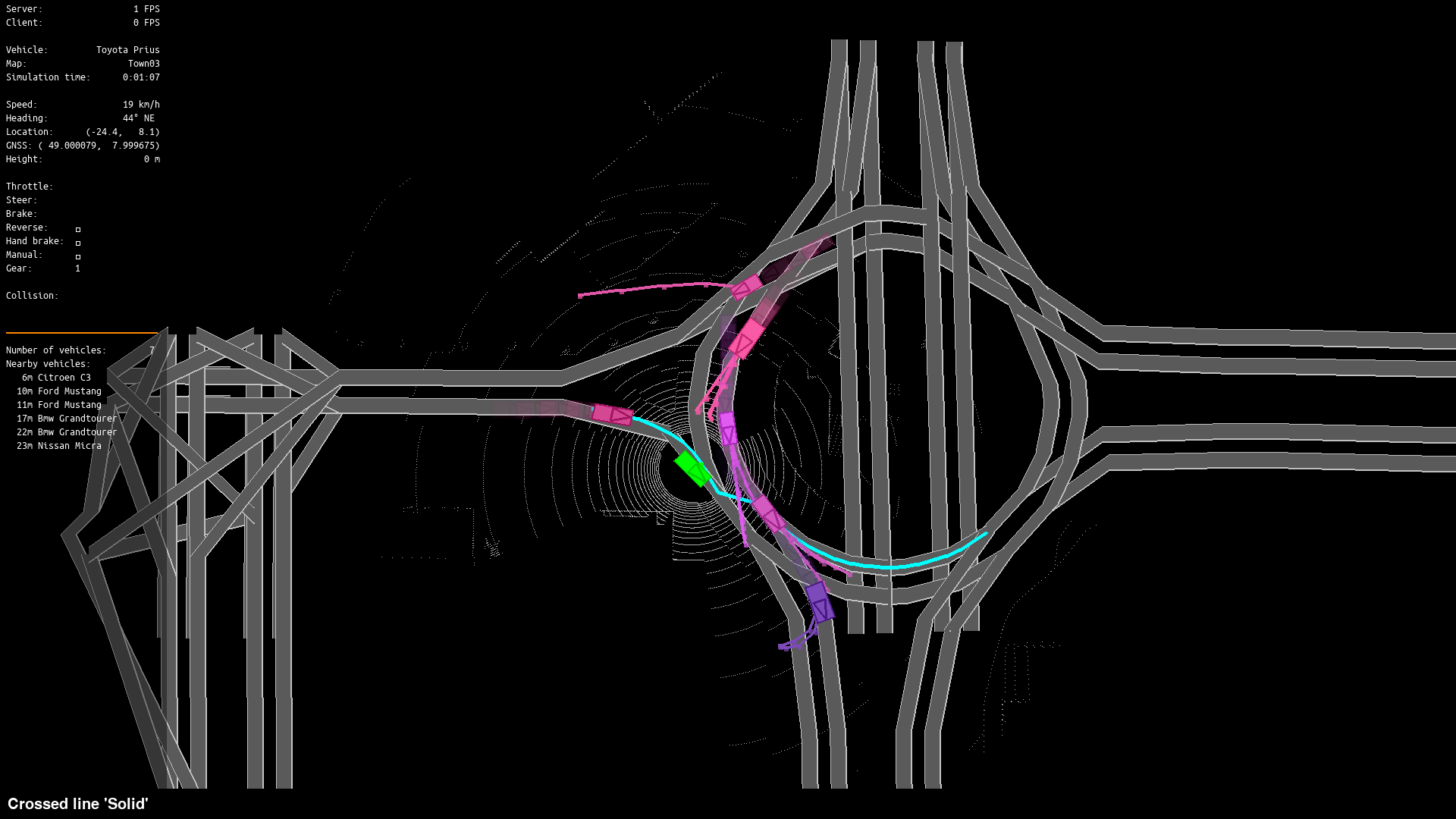}{650 340 750 460}{Non-Reactive, t=3s} \\
	\end{tabular}
	\caption{Visualization of a CARLA roundabout turn for non-reactive (bottom) and reactive (top) models at 3 different time steps: 0s, 1s, 2s, 3s (left-to-right).
	}
	\label{fig:supp_qual_carla2}
\end{figure*}

In Fig. \ref{fig:supp_qual_carla1}, \ref{fig:supp_qual_carla2} we demonstrate a lane merge scenario and a roundabout turn scenario in CARLA. We note that these are complex scenarios involving multiple actors in the lane that the ego-agent is supposed to merge into. In Fig. \ref{fig:supp_qual_carla1}, the visualizations show that the reactive agent is able to spot a gap at $t=2$, and merge in at $t=3$. Meanwhile, the non-reactive agent keeps going straight until $t=3$, and even then it wavers between merging and going straight. Fig. \ref{fig:supp_qual_carla2} demonstrates the ego-agent merging into a roundabout turn with multiple actors. While both models reach similar states initially, towards the end the reactive model reasons that it can still merge inwards, while the non-reactive model is stuck waiting for all the actors to pass.

\begin{figure*}
	\centering
	\def\imw{0.24\textwidth}
	\setlength{\tabcolsep}{0.2pt}
	\begin{tabular}{cccc}
	\figcap{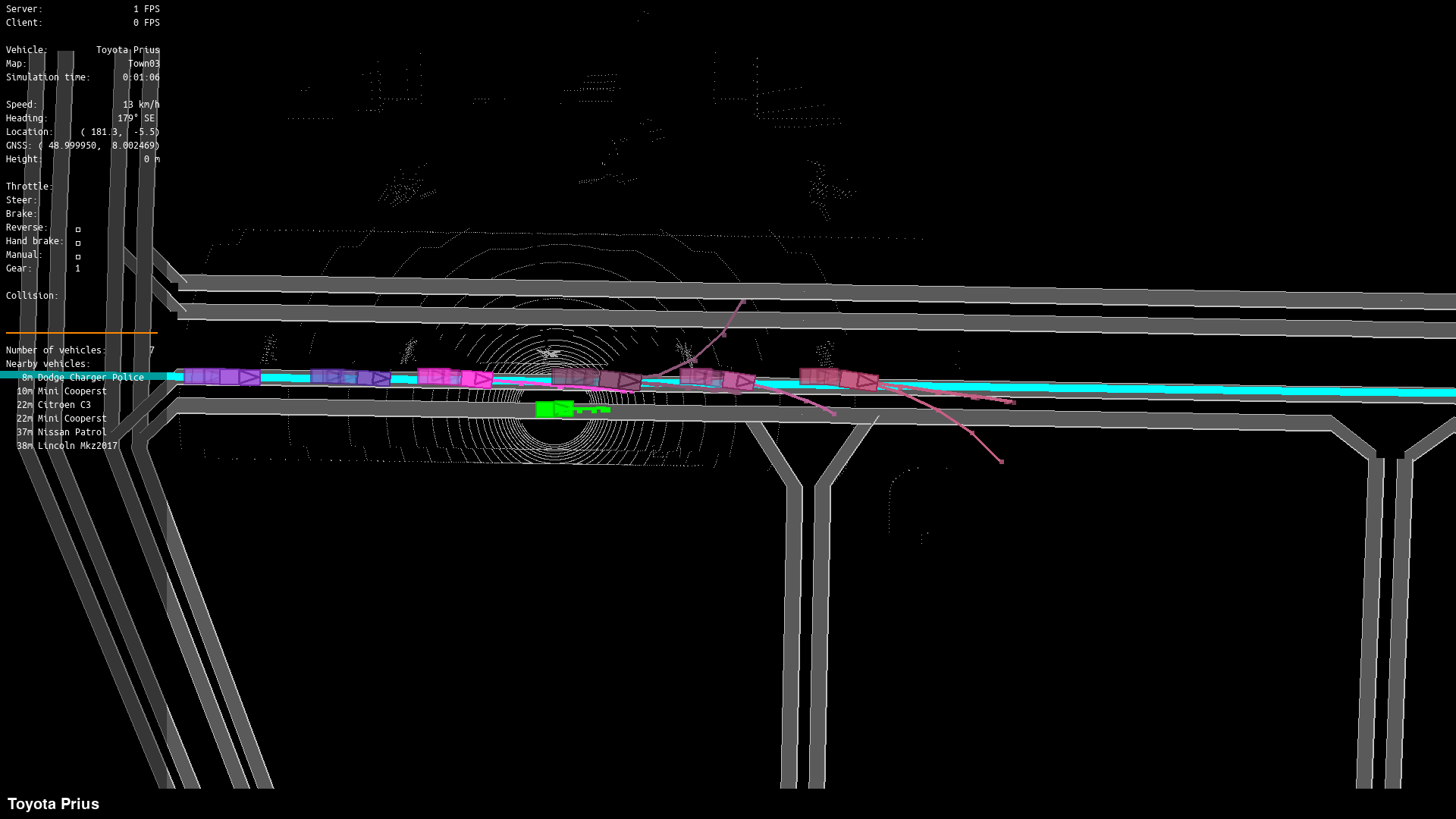}{600 400 800 400}{Reactive, t=1s} & 
	\figcap{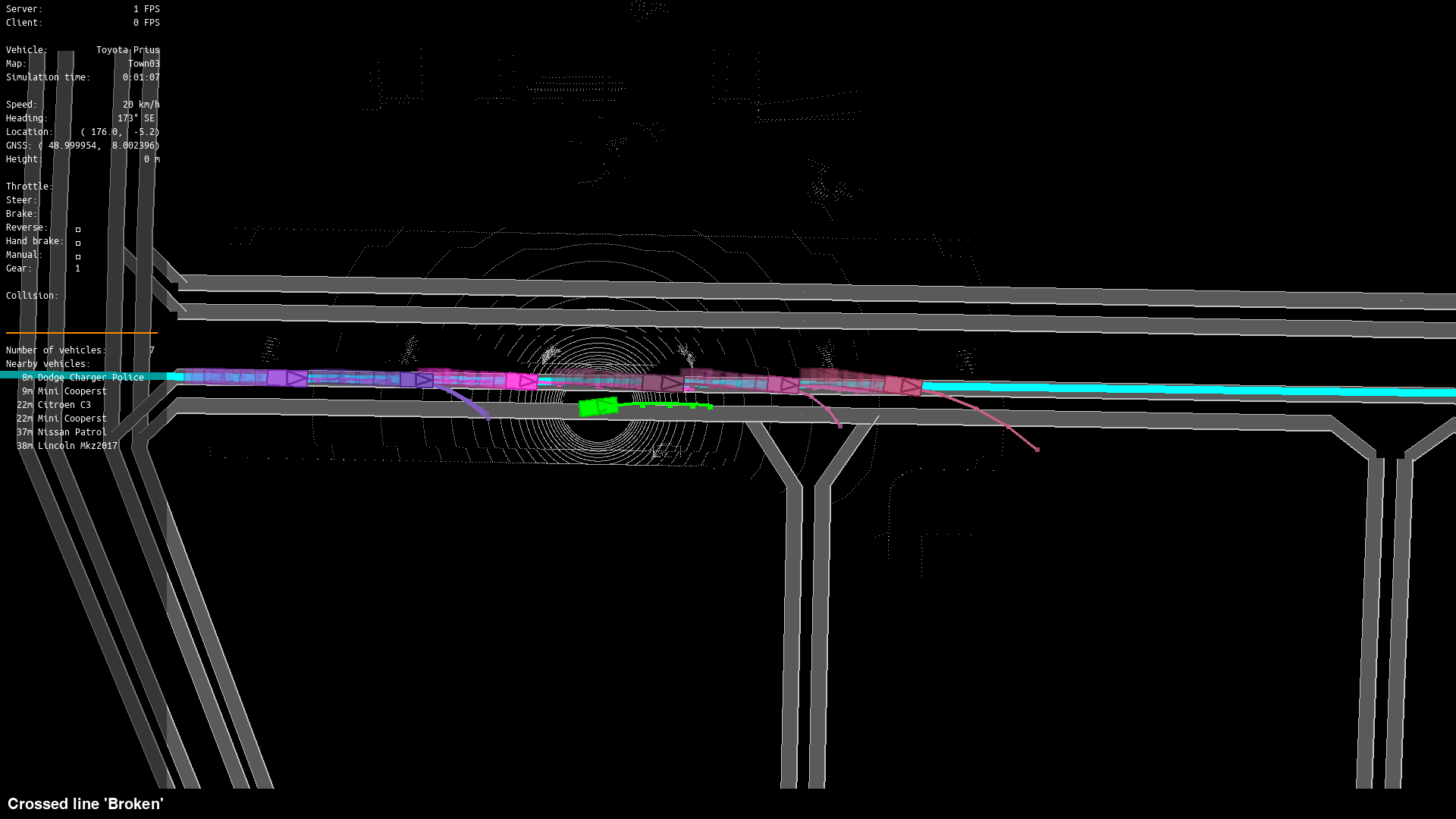}{600 400 800 400}{Reactive, t=2s} &
	\figcap{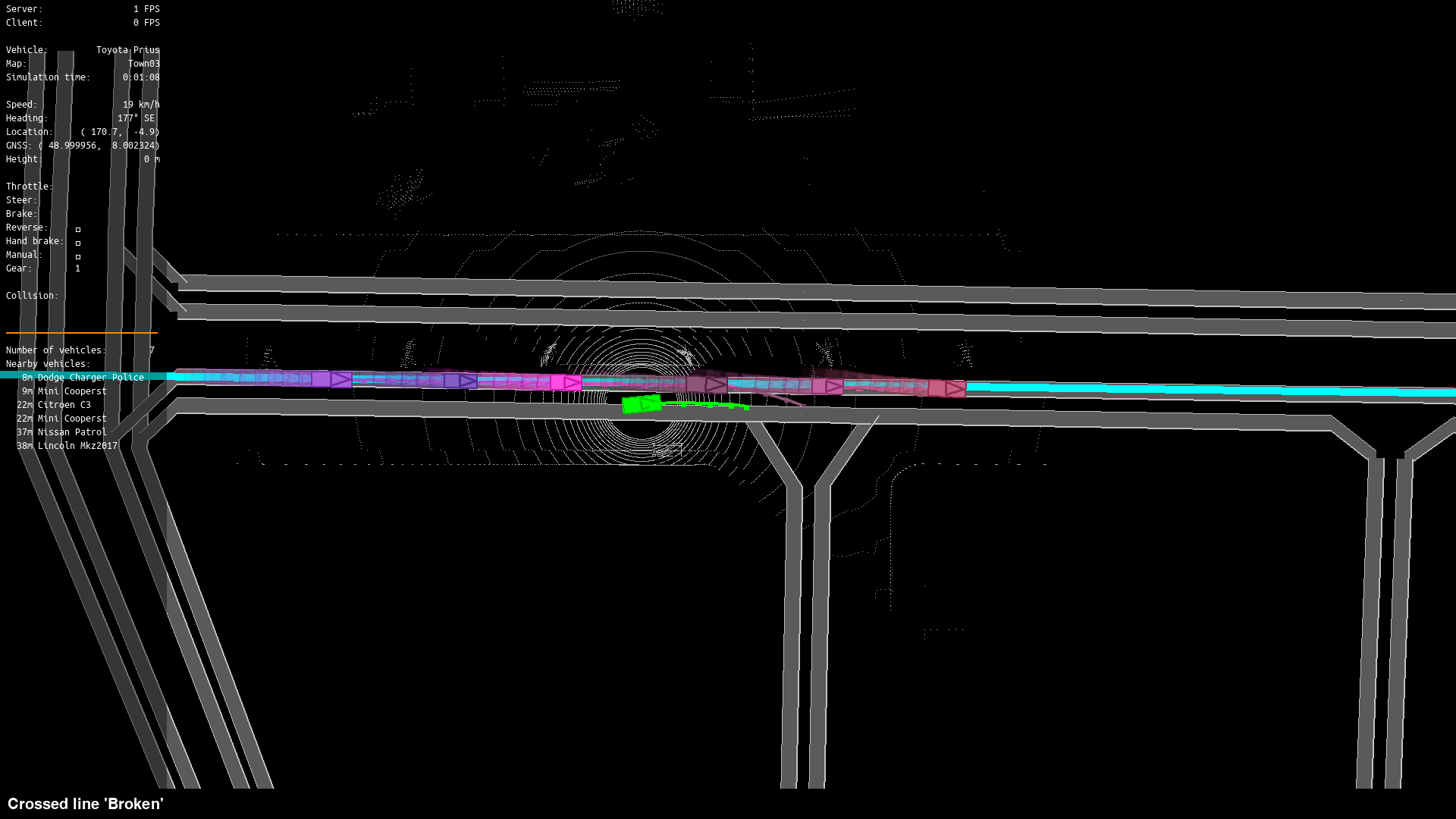}{600 400 800 400}{Reactive, t=3s} &
	\figcap{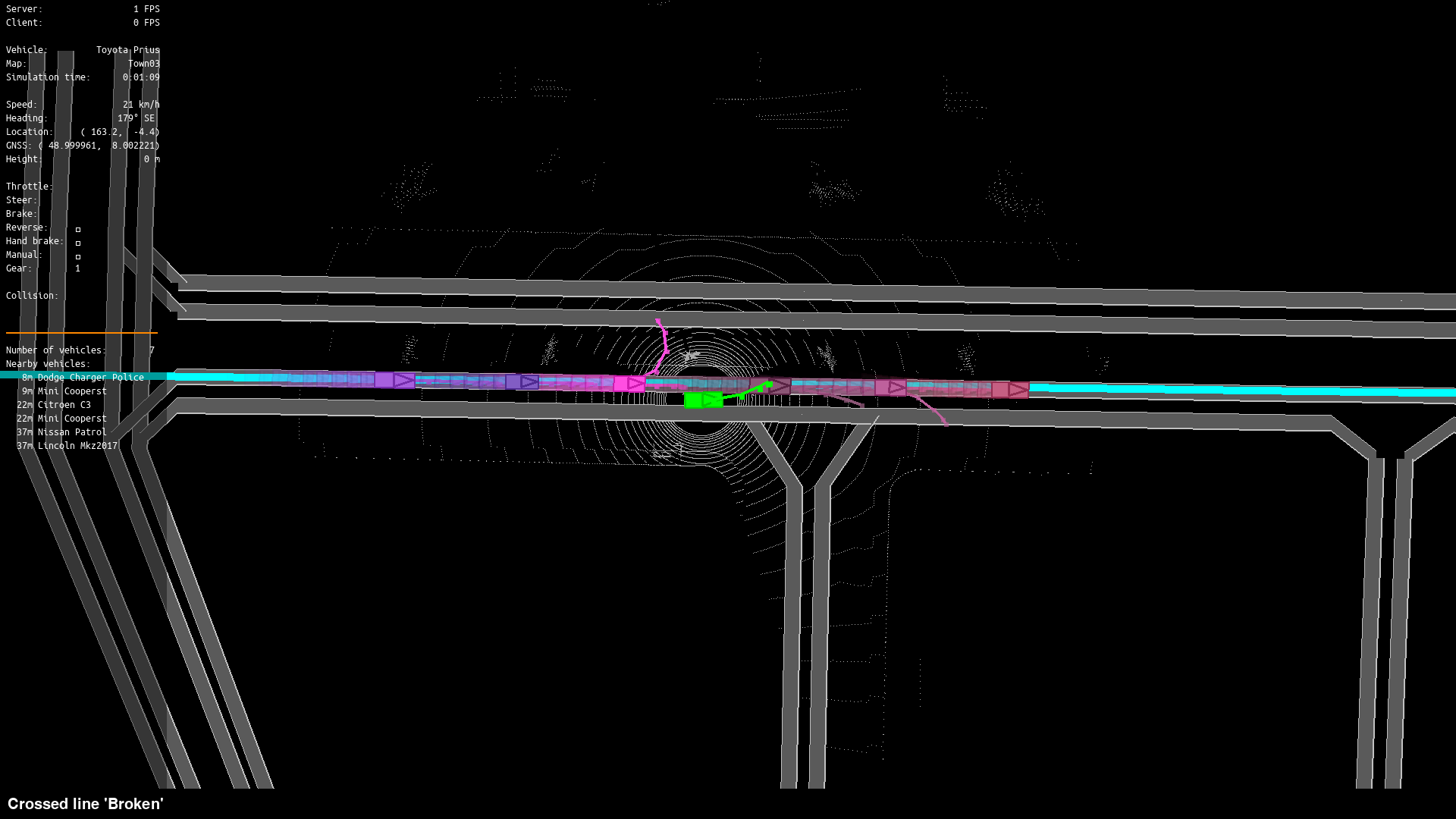}{600 400 800 400}{Reactive, t=4s} \\
	\figcap{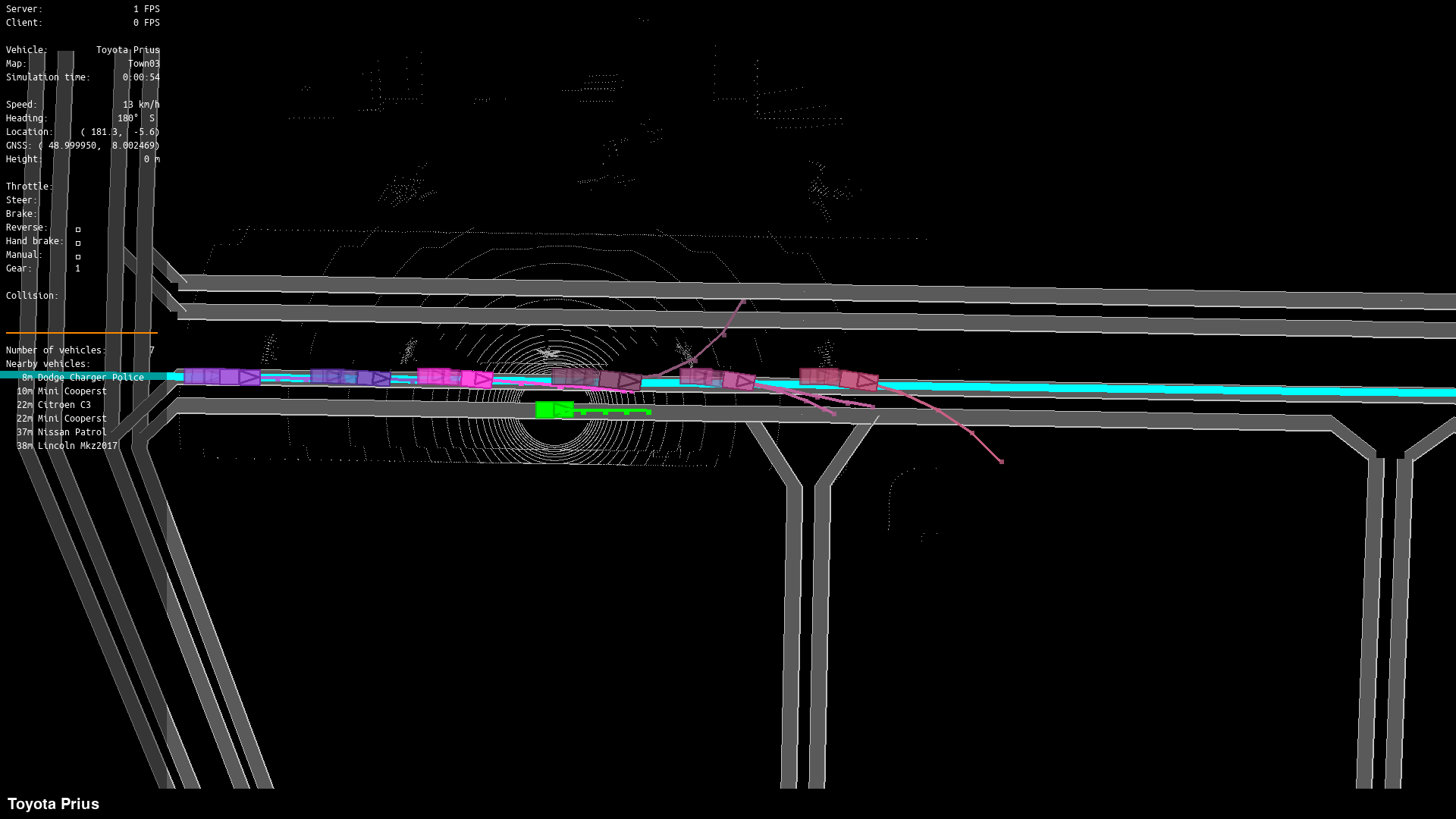}{600 400 800 400}{Non-Reactive, t=1s} &
	\figcap{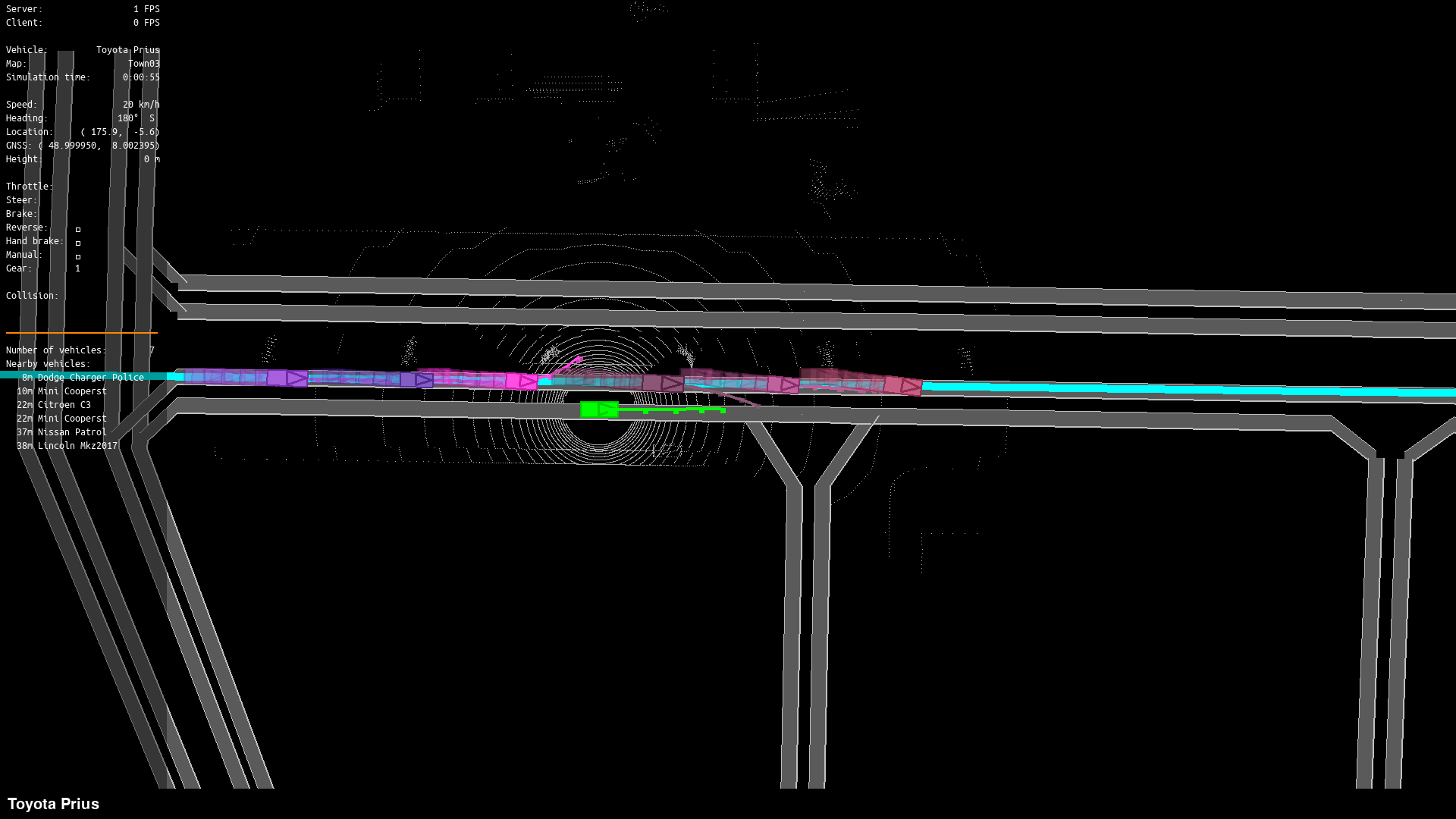}{600 400 800 400}{Non-Reactive, t=2s} &
	\figcap{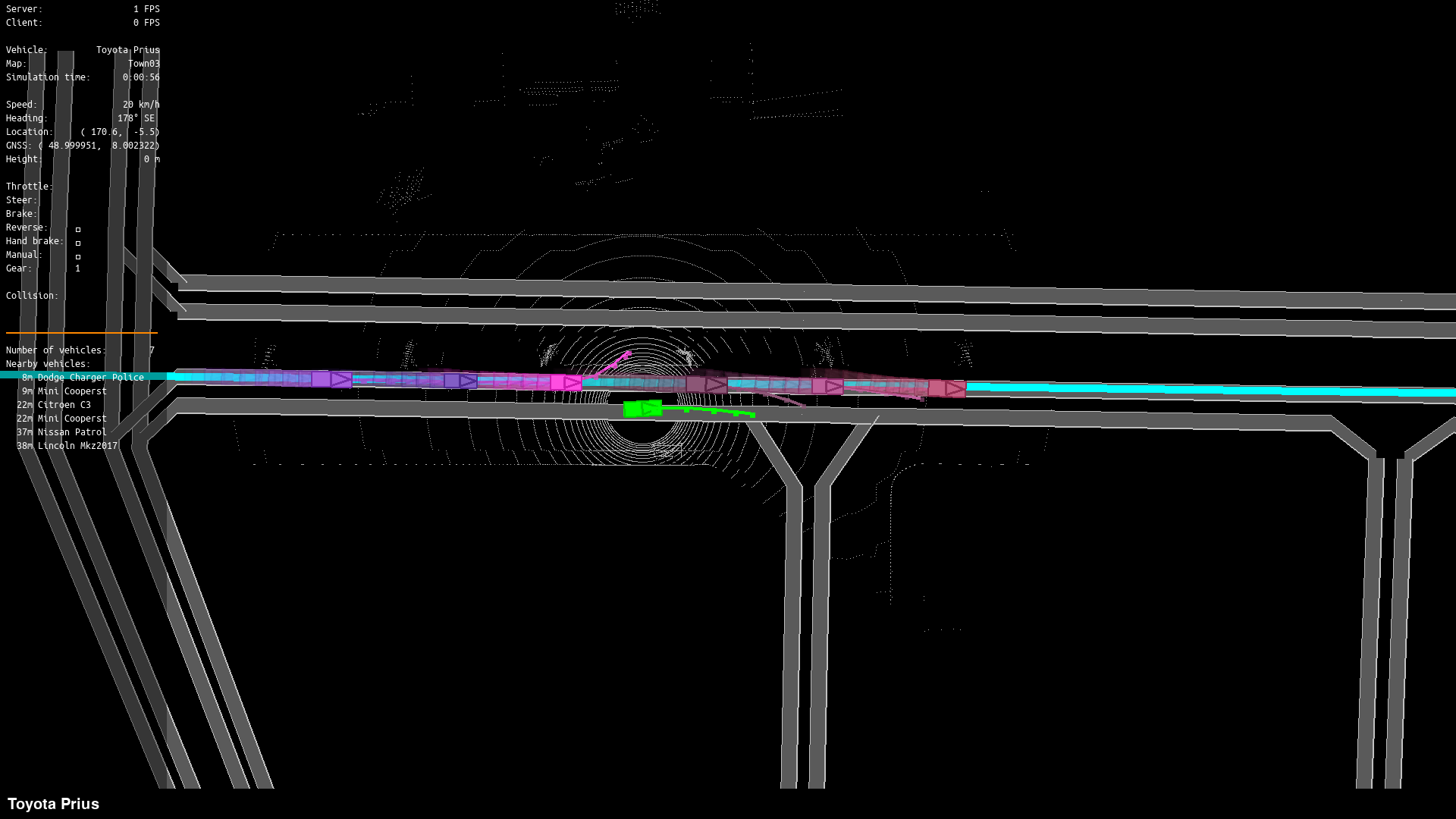}{600 400 800 400}{Non-Reactive, t=3s} &
	\figcap{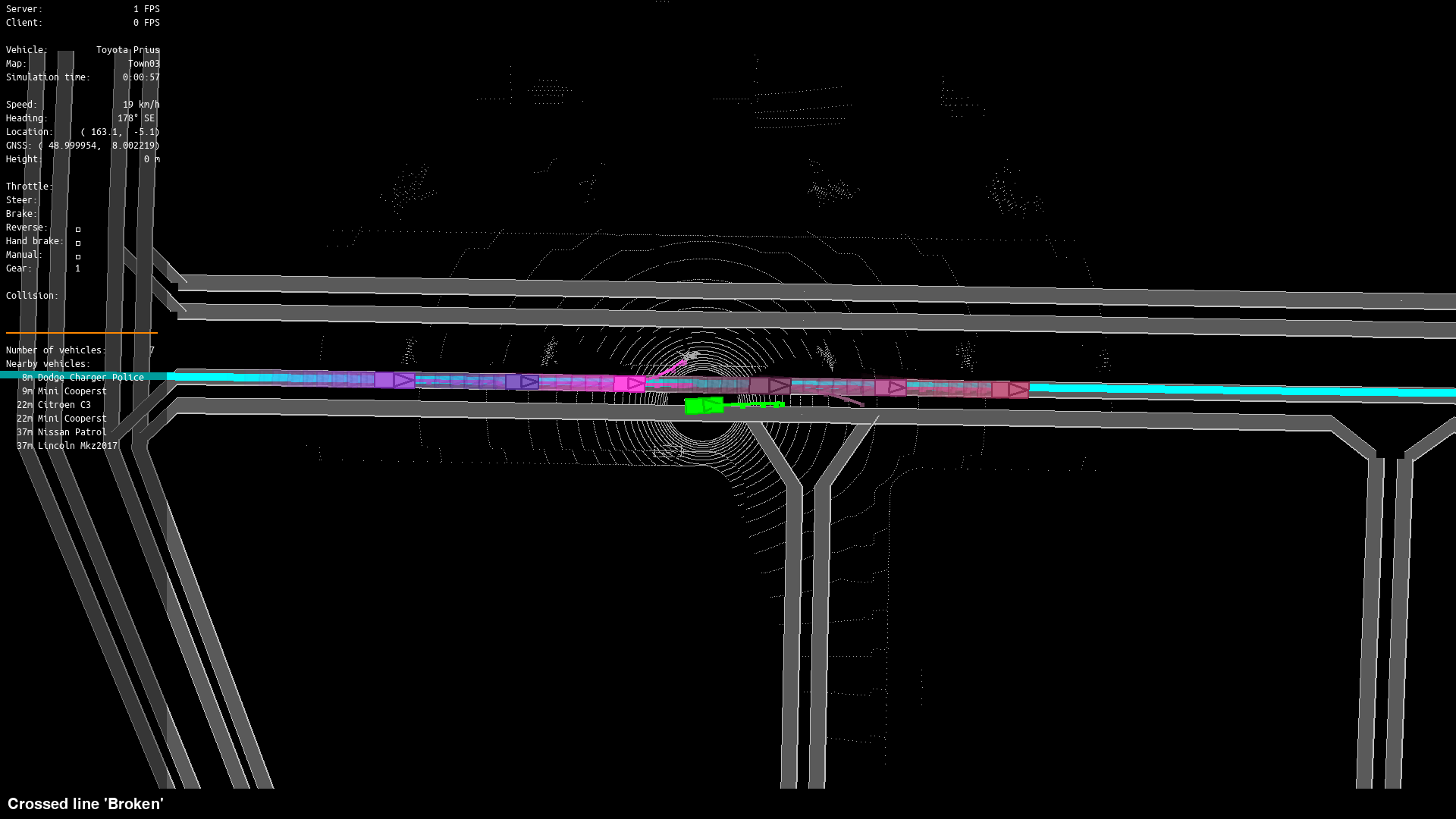}{600 400 800 400}{Non-Reactive, t=4s}  \\
	\end{tabular}
	\caption{Visualization of a CARLA lane merge for non-reactive (bottom) and reactive (top) models at 3 different time steps: 1s, 2s, 3s, 4s (left-to-right). 
	}
	\label{fig:supp_qual_carla1}
\end{figure*}

\clearpage

\end{document}